\documentclass{article}
\PassOptionsToPackage{round}{natbib}

\usepackage[main, final]{neurips_2026}

\usepackage{xcolor}         
\usepackage[utf8]{inputenc} 
\usepackage[T1]{fontenc}    
\definecolor{linkblue}{HTML}{001473}
\usepackage{hyperref}       
\hypersetup{
    colorlinks=true,
    linkcolor=linkblue,
    citecolor=linkblue,
    urlcolor=linkblue
}
\usepackage{url}            
\usepackage{booktabs}       
\usepackage{amsfonts}       
\usepackage{nicefrac}       
\usepackage{microtype}      

\usepackage[nolist,nohyperlinks]{acronym}
\usepackage{amsmath}
\usepackage{bm}
\usepackage{todonotes}
\usepackage{graphicx}
\usepackage[capitalize, nameinlink]{cleveref}
\usepackage{caption}
\usepackage{subcaption}
\usepackage{tikz}
\usetikzlibrary{decorations.pathreplacing, arrows.meta, calc}
\usepackage{multirow}
\usepackage[inline]{enumitem}
\usepackage{wrapfig}
\usepackage{graphicx}
\usepackage{float}

\begin{acronym}[LONGEST]
  \acro{SciML}{Scientific Machine Learning}
  \acro{PDE}{Partial Differential Equation}
  \acro{CFD}{Computational Fluid Dynamics}
  \acro{FDM}{Finite Difference Method}
  \acro{FVM}{Finite Volume Method}
  \acro{FEM}{Finite Element Method}
  \acro{CG}{Continuous Galerkin}
  \acro{RANS}{Reynolds-Averaged Navier-Stokes}
  \acro{CNN}{Convolutional Neural Network}
  \acro{FNO}{Fourier Neural Operator}
  \acro{HF}{High Fidelity}
  \acro{MF}{Multi Fidelity}
  \acro{WLS}{Weighted Least Squares}

\end{acronym}

\title{Compute-optimal data scaling for neural surrogates via multi-fidelity training}

\author{
  \textbf{Paul Setinek}\thanks{Equal contribution. Correspondence to setinek@ml.jku.at}$\:\:^1$\hspace{16px}%
  \textbf{Fabian Paischer}$^{1,2}$\hspace{16px}%
  \textbf{Nils Thuerey}$^{3,4}$\vspace{6px}\\%
  \textbf{Johannes Brandstetter}$^{1,2}$\hspace{16px}%
  \textbf{Felix Koehler}\footnotemark[1]$\:\:^{3,4}$\vspace{8px}\\%
  {$^1$~Institute for Machine Learning, JKU Linz}\hspace{12px}{$^2$~Mistral AI}\vspace{6px}\\%
  {$^3$~Technical University of Munich}\hspace{12px}{$^4$~Munich Center for Machine Learning (MCML)}\vspace{8px}\\%
  \href{https://github.com/psetinek/multi-fidelity-neural-surrogates}{\raisebox{-0.3em}{\includegraphics[width=1.15em]{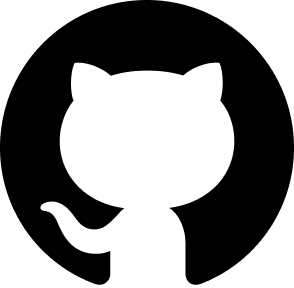}} \texttt{Code}}\hspace{1em}
\href{https://huggingface.co/datasets/psetinek/multi-fidelity-airfoil}{\raisebox{-0.3em}{\includegraphics[width=1.3em]{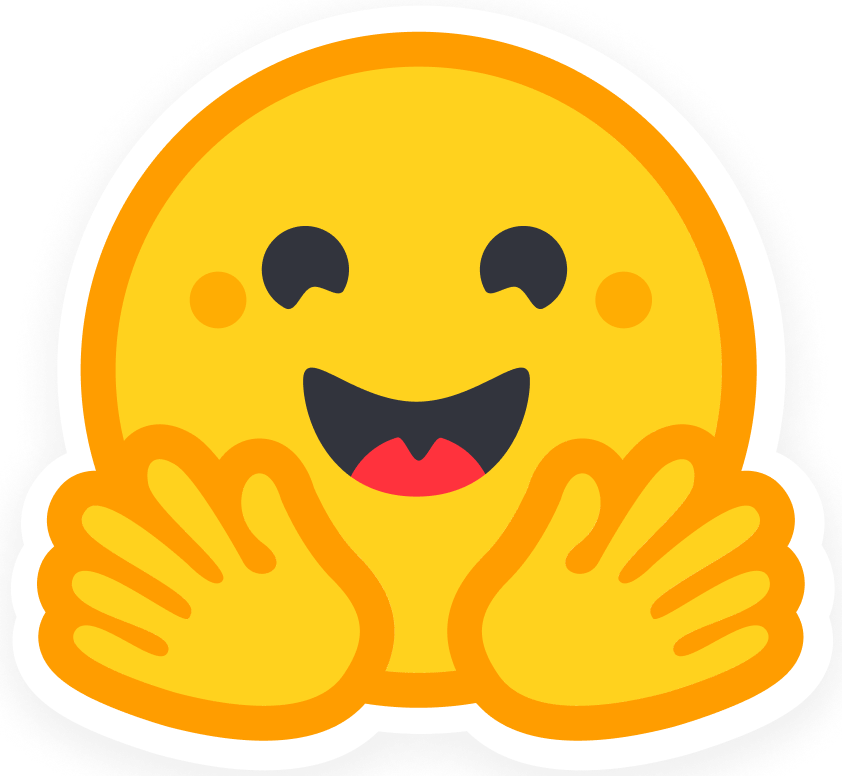}} \texttt{Data}}
\vspace{-12px}
}
\begin{document}

\maketitle

\begin{abstract}
Neural surrogates for \acp{PDE} are usually trained on synthetic data from numerical simulators.
As models grow and scaling laws demand larger datasets, the cost of data generation often outweighs the training cost and becomes the bottleneck in the scaling quest.
Unlike data in other domains, however, each PDE sample has a tunable cost that depends on its numerical fidelity, with the resulting cost-accuracy trade-off governed by known convergence rates.
We exploit this property by formalizing neural surrogate training as a function of both data quantity and fidelity under a fixed data generation compute budget, treating fidelity as a fine-grained discrete spectrum rather than a binary distinction.
Building on this, we show that multi-fidelity training enables Pareto-optimal error scaling against data generation compute.
By using simulations at lower fidelities as an additional training signal, training on the optimal fidelity mixture matches the accuracy of high-fidelity-only training, with up to $6.3\times$ less data generation compute across our settings.
We validate these findings across six \ac{PDE} problems spanning structured grids and unstructured meshes, two distinct error sources (discretization and iterative-solver error), and four surrogate architectures, totaling over 7,000 training runs.
\end{abstract}

\section{Introduction}
\label{sec:introduction}
Modern machine learning owes a large portion of its success to scale.
This holds across various domains, like computer vision, time series, and especially natural language processing, where models reach hundreds of billions of parameters \citep{grattafiori2024llama3herdmodels, mistral2024large2}.
Following the seminal works on neural scaling laws \citep{kaplan2020scaling,hoffman2022chinchilla}, this growth has been enabled by jointly scaling up model size, training compute and datasets.

The \ac{SciML} community has started to adopt this paradigm, leading to notable breakthroughs in weather and climate forecasting \citep{keisler2022forecasting_global_weather, pathak2022fourcastnet, nguyen2023climax, sanchez2025gencast, bodnar2024aurora}, material design \citep{merchant2023scaling_dl_for_materials, zeni2025generative_materials, yang2024matter_sim} or protein folding \citep{jumper2021alphafold, abramson2024alphafold3}.
These advances have partly been enabled by large curated public datasets, such as ERA5 in weather modeling \citep{hersbach2020era5} or the Protein Data Bank (PDB) \citep{berman2000pdb}.
However, only few scientific domains benefit from such large readily available datasets.

Neural surrogates for \acp{PDE}, which have also been scaled to billions of parameters \citep{paischer2025gyroswin,wang2024orbit,bodnar2024aurora}, are typically trained on synthetic data generated by numerical solvers~\citep{holzschuh2026p3d}.
Therefore, every data sample incurs a non-negligible computational cost, and scaling along the data axis cannot be done by further scraping of existing sources.
\citet{ashton2025fluid} give a prominent example of this, showing that the dominating cost for scaling surrogate models in \ac{CFD} is not the network training, but the data generation.
Furthermore, this cost directly affects the practical viability of such models: the more expensive the dataset, the further the amortization point is pushed back, delaying the stage at which the surrogate becomes cost-effective for classical downstream tasks such as design optimization.

With this in mind, current datasets, regardless of their scale, are typically generated entirely at a single, fixed fidelity, assumed to be high enough for the task at hand.
However, this paradigm neglects a fundamental property of numerical solvers: the computational cost of a sample is not fixed but depends on its fidelity, and the resulting cost-accuracy trade-off follows known convergence rates.
Data quality is thus a controllable, predictable function of compute, unlike in many other domains, where it emerges from curation and filtering~\citep{sorscher2022beyond}. This raises a natural question:

\begin{quote}
    \emph{At the same data generation cost, can using a mixture of samples at different fidelities yield better surrogate accuracy than using high-fidelity data alone?}
\end{quote}

\begin{figure}[t]
    \centering
    \includegraphics[width=0.8\linewidth]{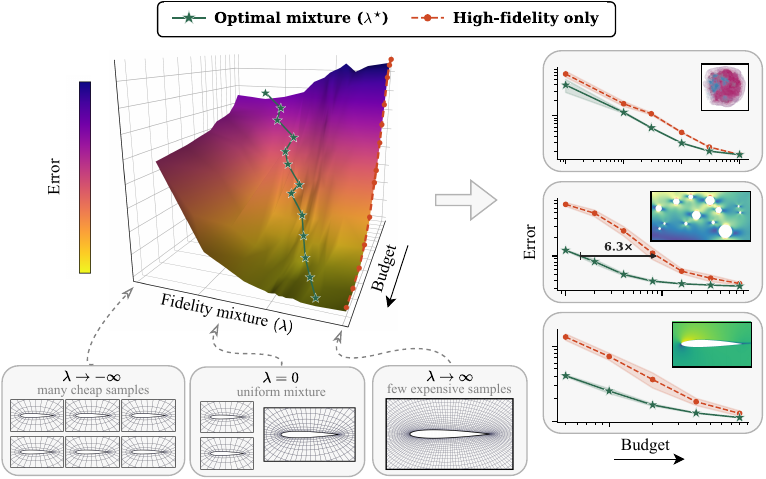}
    \caption{Overview of our multi-fidelity data scaling approach. We split the data axis into generation budget and fidelity mixture, parametrized by $\lambda$. \textbf{Bottom:} example discretization fidelity mixtures. \textbf{Right:} Across diverse settings, multi-fidelity training at optimal mixtures (green) reaches the error of high-fidelity-only training (orange, dashed) with up to $6.3\times$ less data generation compute.
    }
    \label{fig:fig_1}
    \vspace{-12pt}
\end{figure}

We provide empirical evidence that the answer is yes (\cref{fig:fig_1}).
Our main contributions are:

\begin{itemize}[leftmargin=*]
    \item We formalize the neural surrogate training pipeline as a function of both data quantity and fidelity under a fixed data generation compute budget, revealing the inherent trade-off between sample quality and dataset size. We are the first to treat fidelity as a fine-grained discrete spectrum throughout the pipeline, in contrast to the binary formulations prevalent in existing literature.
    \item We show empirically that neural surrogates learn the fidelity of their training data, and that fidelity conditioning enables steering across the fidelity spectrum at inference.
    \item We validate that multi-fidelity training at optimal mixtures enables Pareto-optimal data scaling: across multiple \acp{PDE}, discretization types and neural surrogate architectures, it consistently matches or outperforms single-fidelity training at matched data generation compute. To the best of our knowledge, this is the broadest empirical study of multi-fidelity neural surrogate data scaling,
    comprising over 7,000 training runs and more than 12,000 GPU-hours.
\end{itemize}

\section{Related work}
\label{sec:related_work}

\paragraph{Data scaling and data quality.}
Neural scaling laws for language models established that data must be scaled jointly with model size and training compute to achieve optimal performance~\citep{kaplan2020scaling,hoffman2022chinchilla}.
Subsequent work has increasingly highlighted the importance of data quality rather than sheer quantity~\citep{gunasekar2023textbooks,li2023textbooksareallyouneed2, sachdeva2024dataefficientllms,deepseekai2025deepseekr1}, a finding that has also been acknowledged in time series~\citep{boileau2025towards_interpretable_time_series_fm,ansari2024chronos,auer2025tirex} and vision-language modeling~\citep{sorscher2022beyond,nguyen2022quality,gadre2023datacomp}.
Closest to our work, \citet{goyal2024scaling_laws_data_filtering} show that for vision-language models, the optimal data quality--quantity balance depends on the available compute budget.
However, their analysis targets the budget spent on model training, whereas we focus on the budget for dataset generation, being the primary bottleneck for many AI4Science applications.
In the context of neural surrogates for \acp{PDE}, recent work studies optimal model scaling in the weather domain~\citep{subramanian2026scaling_laws_weather,yu2026scaline_laws_weather_2}.
However, these are based on the openly available ERA5 dataset~\citep{hersbach2020era5} and thus do not account for simulation cost or quality.

\paragraph{Multi-fidelity neural surrogates.}
Recognizing the scarcity of high-fidelity data, prior work has combined inexpensive low-fidelity and scarce high-fidelity simulations to train neural surrogates for field predictions, e.g. via composite DeepONets that learn explicit corrections on top of a low-fidelity network~\citep{lu2022deeponet, howard2023mf_deeponet}.
Others employ transfer learning from low- to high-fidelity data with FNOs~\citep{tang2023mf_fno,carey2025plasma} or U-Nets~\citep{zhang2024windfarm,jiang2023mf_unet}.
While motivated by data scarcity, these approaches neither analyze how surrogate error scales with data generation cost, nor the trade-off between sample fidelity and dataset size at fixed compute.
Closer to our setting, \citet{song2021transfer_learning_mf} and \citet{lyu2023mf_fno_fluids} vary the mixture of training samples at two discretization fidelities under a fixed data generation budget for two-phase porous media flow and heat conduction, respectively.
\citet{setinek2025towards} explore this for a transformer trained on \ac{RANS} simulations using two fidelity levels obtained from different boundary layer assumptions.
Moving beyond binary fidelity levels, \citet{sakarvadia2026false} introduce multi-resolution training to remedy the failure of neural operators to generalize zero-shot across resolutions.
Their low-fidelity samples, however, are created by downsampling high-fidelity simulations, therefore neither mimicking real discretization errors nor accounting for real data cost.
\citet{choudhary2025pregenerating} present a different approach for data-cost-aware training of neural surrogates by mixing simulations of varying problem difficulties (Reynolds number and obstacle count in the domain) that require different compute to generate.

With our work, we unify and extend this fractured landscape: we present an extensive study of error scaling against data generation compute and fidelity mixture, treating fidelity as a fine-grained discrete spectrum spanning multiple error sources (discretization and solver), which naturally mimics real-world pipelines.
We report results across various PDEs, discretization types (regular grids and unstructured meshes), and surrogate architectures including scalable transformers.

\section{Methodology}
\label{sec:methodology}

\subsection{Errors in numerical simulators}
\label{sec:data_generation_pipeline}

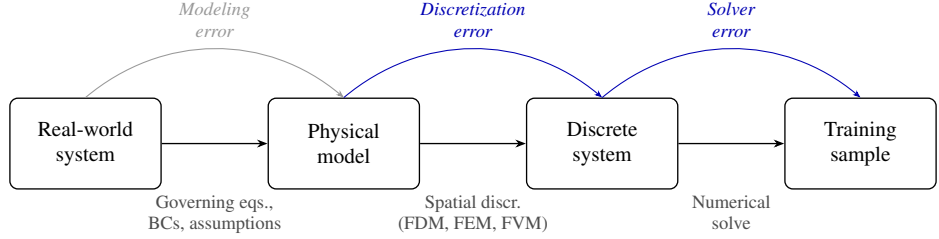
\begin{figure}[tbp]
    \centering
    \begin{tikzpicture}[
        box/.style={
            draw, rounded corners=3pt, minimum width=2.0cm, minimum height=1.2cm,
            align=center, font=\footnotesize, line width=0.5pt
        },
        process/.style={
            -{Stealth[length=4pt]}, line width=0.5pt, 
        },
        error/.style={
            -{Stealth[length=3pt]}, line width=0.4pt, bend left=40,
        },
        error_controlled/.style={
            error, color=blue!70!black,
        },
        error_fixed/.style={
            error, color=black!40,
        },
        errlabel/.style={
            font=\scriptsize\itshape, align=center,
        },
        errlabel_controlled/.style={
            errlabel, color=blue!70!black,
        },
        errlabel_fixed/.style={
            errlabel, color=black!40,
        },
        proclabel/.style={
            font=\scriptsize, align=center, text=black!70,
        },
        costlabel/.style={
            font=\tiny, color=blue!70!black, align=center,
        },
    ]
    
    \node[box] (real) {Real-world\\system};
    \node[box, right=1.4cm of real] (phys) {Physical\\model};
    \node[box, right=1.4cm of phys] (disc) {Discrete\\system};
    \node[box, right=1.4cm of disc] (sol) {Training\\sample};
    
    \draw[process] (real) -- (phys) 
        node[proclabel, midway, below=16pt] {Governing eqs.,\\BCs, assumptions};
    \draw[process] (phys) -- (disc) 
        node[proclabel, midway, below=16pt] {Spatial discr.\\(FDM, FEM, FVM)};
    \draw[process] (disc) -- (sol) 
        node[proclabel, midway, below=16pt] {Numerical\\solve};
    
    \draw[error_fixed] (real.north) to 
        node[errlabel_fixed, above=1pt] {Modeling\\error} (phys.north);
    \draw[error_controlled] (phys.north) to 
        node[errlabel_controlled, above=1pt] {Discretization\\error} (disc.north);
    \draw[error_controlled] (disc.north) to 
        node[errlabel_controlled, above=1pt] {Solver\\error} (sol.north);
    
    \end{tikzpicture}
    \caption{Data generation pipeline for neural surrogates, highlighting 
    the primary sources of numerical error. \emph{Discretization error} and \emph{solver error} (blue), which exhibit known convergence behaviors and can be reduced at the expense of additional compute.}
    \label{fig:data_generation_pipeline}
    \vspace{-12pt}
\end{figure}

For most differential equations governing physical systems of interest, there exists no analytical solution, necessitating the use of numerical methods.
We therefore assume access to a \emph{numerical simulator} $\mathcal{P}_h$ to generate training data for neural surrogates.
Given a (potentially vector-valued) discrete input field $\mathbf{f}_h$ at discretization $h$ sampled from distribution $\mathcal{F}_h$, $\mathcal{P}_h$ produces a target solution
\begin{equation}
    \mathbf{u}_h = \mathcal{P}_h(\mathbf{f}_h;\, \xi).
\end{equation}
Unavoidably, the output $\mathbf{u}_h$ is an imperfect approximation of the true solution.
We therefore introduce $\xi$ as a for now abstract scalar \emph{fidelity parameter} controlling the accuracy of the simulator.
By our convention, $\xi$ is ordered such that larger values correspond to higher accuracy.
We can then define the numerical error of the simulation as:
\begin{equation}
    e(\xi)
    =
    \mathbb{E}_{\mathbf{f}_h \sim \mathcal{F}_h} \left[
        \frac{
            \left\|
            \mathcal{P}_h(\mathbf{f}_h; \xi)
            -
            \lim_{\xi' \to \infty} \mathcal{P}_h(\mathbf{f}_h; \xi')
            \right\|_2
        }{
            \left\|
            \lim_{\xi' \to \infty} \mathcal{P}_h(\mathbf{f}_h; \xi')
            \right\|_2
        }
    \right].
\label{eq:numerical_error}
\end{equation}
\cref{fig:data_generation_pipeline} illustrates the simulation pipeline, highlighting the three primary sources of numerical error.

\paragraph{Modeling error.}
The first step is to formulate a \emph{physical model} of the underlying real-world system.
This involves selecting governing equations and boundary conditions, as well as simplifying assumptions (e.g. turbulence closure models).
This introduces \emph{modeling error}, which we do not address in this work, as to date there is limited evidence of meaningful transfer across modeling fidelities~\citep{zhang2024windfarm,setinek2025towards}.

\paragraph{Discretization error.}
The physical model must then be discretized in space to obtain a finite-dimensional system of equations.
Different methods exist for this purpose, such as \ac{FDM}, \ac{FEM}, or \ac{FVM}, each with specific requirements on the spatial grid or mesh.
This step introduces \emph{discretization error}.
Here, the fidelity parameter $\xi$ corresponds to a measure of resolution (e.g., grid points or total mesh nodes), and the error decays \emph{algebraically}, i.e., $e(\xi) = \mathcal{O}(\xi^{-\alpha})$, where $\alpha > 0$.

\paragraph{Solver error.}
The discretized system must then be solved numerically.
For linear problems, this yields a system of linear equations that can be tackled by direct methods (e.g., LU factorization) or iterative solvers (e.g., Conjugate Gradient, Gauss--Seidel, Multigrid).
For nonlinear problems, an outer nonlinear iteration (e.g., Newton--Raphson) is required, each step of which involves solving a linearized sub-problem.
When iterative methods are used, early termination introduces \emph{solver error}.
Here, $\xi$ corresponds to the number of iterations, and the error decays \emph{exponentially}, i.e. $e(\xi) = \mathcal{O}(\rho^{\xi})$, with $\rho \in (0,1)$.
The same axis has been exploited at training time by \citet{bhatia2025prdp} in the context of differentiable physics.

\paragraph{Cost--accuracy trade-off.}
In both regimes, increasing fidelity $\xi$ reduces the error but incurs additional computational cost $c(\xi)$.
Under a fixed compute budget $B$ to generate a dataset at a single fidelity $\xi_{\mathrm{fix}}$, one can afford at most $M = \lfloor B / c(\xi_{\mathrm{fix}}) \rfloor$  training samples.

\subsection{Multi-fidelity formalism}
\label{sec:formalism}
We now relax this requirement of a fixed fidelity and instead introduce a support of admissible fidelities $\Xi = \{\xi_1, \xi_2, \dots, \xi_{|\Xi|}\}$, ordered by accuracy, and with endpoints $\xi_{\min}$ and $\xi_{\max}$ (see \cref{app:fidelity_support} for desiderata of how to select $\Xi$).
These define a \emph{discrete spectrum} over which we will form fidelity mixtures parametrized by $\lambda \in \mathbb{R}$.
Because raw fidelity values have inconsistent units across error sources (e.g. $N$ grid points vs $K$ solver iterations) and span different orders of magnitude, we apply
a normalization and clamping $\hat{\xi} \leftarrow \xi$ involving $\xi_{\text{min}}$ and $\xi_{\text{max}}$ (\cref{app:fidelity-normalization}).
This ensures comparability of the mixture parameter $\lambda$ across experiments.

\begin{figure}[tbp]
    \centering
    \includegraphics[width=\linewidth]{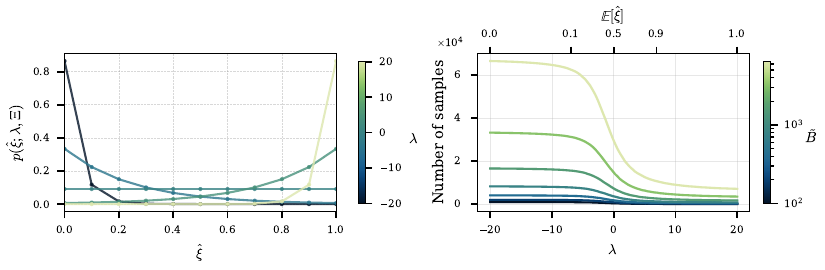}
    \caption{Overview of the multi-fidelity sampling approach. \textbf{Left:} Probability mass function with higher $\lambda$ biasing towards higher fidelities $\hat{\xi}$. \textbf{Right:} Example dataset compositions for the linear elasticity solver error experiment.}
    \label{fig:mf-sampling}
    \vspace{-10pt}
\end{figure}

Now, let $p(\hat{\xi};\lambda, \Xi)$ be a probability mass function for admissible fidelities parametrized by $\lambda$ and $\Xi$.
For this work, we employ a softmax distribution over the discrete spectrum of fidelities:
\begin{equation}
\label{eq:pmf}
    p(\hat{\xi}_i; \lambda, \Xi) = 
    \frac{\exp(\lambda \, \hat{\xi}_i)}{\sum_{j=1}^{|\Xi|} \exp(\lambda \, \hat{\xi}_j)}.
\end{equation}
Then, $\lambda >0$ biases towards higher fidelities, $\lambda <0$ towards lower fidelities, and $\lambda= 0$ equals a uniform distribution.
In the extreme cases, $\lambda \to \infty$ yields exclusively $\xi_{\max}$, while $\lambda \to -\infty$ results in only $\xi_{\min}$.
Ultimately, $\lambda$ is a hyperparameter controlling the trade-off between sample accuracy and dataset size.
We can now draw a dataset $D \sim \mathcal{D}(\lambda, B)$ of fidelity-tagged input-output pairs with $\mathbb{E}[M] = B / \mathbb{E}_{\hat{\xi} \sim p}[c(\hat{\xi})]$ expected samples.
Additionally, we
define the notion of a ``normalized budget'' as $\tilde{B} = B/c(\hat{\xi}_{\max})$.
We discuss the implementation of the sampling procedure in \cref{app:budgeted_sampler}.
For illustration, \cref{fig:mf-sampling} shows the probability mass function and example dataset compositions.

To fully leverage multi-fidelity data, our surrogate model $\mathcal{S}_{\theta}$ accepts $\hat{\xi}$ as an additional conditioning signal besides the input field $\mathbf{f}_h$.
The normalization in $\hat{\xi}$ gives the surrogate a conditioning input that is linear in the (theoretical) error, which we will see (\cref{sec:conditioning_experiments}) enables steering and interpolation across fidelities at inference.
We train it via empirical risk minimization
\begin{equation}
    \hat{\theta}(\lambda, B)
    = \arg\min_{\theta} \;
    \mathbb{E}_{
        (
            \mathbf{f}_h,\,
            \xi
        )
        \,
        \sim
        \,
        D(\lambda, B)
    } \left[
    \zeta\left(
        \mathcal{S}_{\theta}\left(\mathbf{f}_h; \hat{\xi}(\xi)\right),\;
        \mathcal{P}_h(\mathbf{f}_h; \xi)
    \right)
    \right]
\end{equation}
over the available dataset (for a definition of the loss function $\zeta$, see \cref{app:train_error}).
The learned parameters $\hat{\theta}(\lambda, B)$ depend on the dataset composition defined by the mixture hyperparameter $\lambda$ and the available budget $B$.
We evaluate the surrogate on the highest in-support fidelity $\xi_{\text{max}}$\footnote{Conditioning does not affect surrogate inference speed since the constant conditioning $\hat{\xi}(\xi_{\text{max}})$ can be baked into the weights of the network.}, via
\begin{equation}
    \mathcal{E}(\lambda, B)
    = \mathbb{E}_{\mathbf{f}_h\,\sim\,D_{\text{test}}}\!\left[
        \tilde{\zeta}\left(
            \mathcal{S}_{\hat{\theta}(\lambda, B)}(\mathbf{f}_h; \hat{\xi}(\xi_{\max})),\;
            \mathcal{P}_h(\mathbf{f}_h; \xi_{\max})
        \right)
    \right],
\end{equation}
over a hold-out dataset $D_{\text{test}}$$\sim \lim_{\lambda \to \infty} \mathcal{D}(\lambda, B_{\text{test}})$ (for a definition of the test metric $\tilde{\zeta}$, see \cref{app:test_error}).
As such, the test error is a function of the budget $B$ and the mixture
hyperparameter $\lambda$ besides the model architecture, training recipe and hyperparameters.
For each budget $B$, define the \emph{compute-optimal mixture}
\begin{equation}
    \lambda^{\star}(B) = \arg\min_{\lambda \in \mathbb{R}}
    \mathcal{E}(\lambda, B).
\end{equation}
The associated \emph{compute-optimal test-error Pareto frontier} is
$B \mapsto \mathcal{E}\bigl(\lambda^{\star}(B), B\bigr)$.

\section{Validating the multi-fidelity framework}
\label{sec:validating_mf_framework}

\subsection{Formulating fidelity as a fine-grained spectrum}
\label{sec:spectrum_vs_binary}

\begin{wrapfigure}{r}{1.8in}
    \centering
    \vspace{-20pt}
    \includegraphics[width=1.8in]{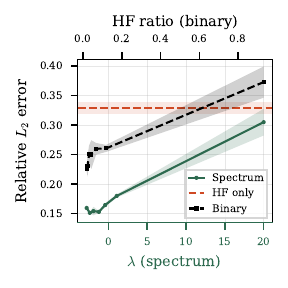}
    \caption{
    Spectrum vs.\ binary approach for a Transolver model on the hyperelasticity dataset (discretization error) at $\tilde{B}=200$.}
    \label{fig:mf_vs_bf_comparison}
    \vspace{-5pt}
\end{wrapfigure}

Treating fidelity as a fine-grained spectrum is natural given the underlying numerical simulation pipeline, where discretization and solver settings can be varied in small increments.
Nevertheless, most prior multi-fidelity work uses a binary formulation with two distinct fidelity levels (\cref{sec:related_work}).
We compare the two approaches at the same data generation budget.
The binary approach uses only the coarsest and finest resolution, whereas our spectrum approach samples 19 additional intermediate resolutions.
\cref{fig:mf_vs_bf_comparison} shows that the spectrum formulation achieves a 33\% lower error than the binary baseline, with a noticeably wider minimum in $\lambda$. Both mixtures outperform a surrogate trained on high-fidelity data alone. The spectrum formulation samples densely along the fidelity axis, forcing the surrogate to learn a coherent mapping that varies smoothly with $\xi$. This effect grows for wider endpoints and, consistent with this, the gap narrows when the fidelity range is smaller, as for solver error on a fixed mesh (\cref{app:binary_vs_spectrum}).

\subsection{Inference-time fidelity steering}
\label{sec:conditioning_experiments}
\begin{wrapfigure}{r}{1.8in}
    \centering
    \vspace{-20pt}
    \includegraphics[width=1.8in]{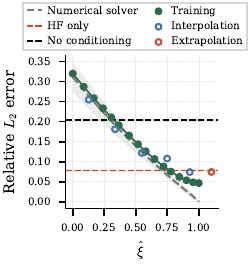}
    \caption{
    The multi-fidelity surrogate respects the conditioning signal. We show a Transolver model on the hyperelasticity dataset (solver error) trained at $\tilde{B}=1{,}600$ and $\lambda^{\star}=-3.59$.}
    \label{fig:conditioning_plot}
    \vspace{-10pt}
\end{wrapfigure}

Conditioning the surrogate on sample fidelity during training enables steering at inference. Feeding a chosen $\hat{\xi}$ produces an error matching the corresponding fidelity level.
\cref{fig:conditioning_plot} compares a surrogate trained with our multi-fidelity approach to one trained on the same training data composition but without the conditioning mechanism.
The unconditioned baseline exhibits a single error level regardless of $\hat{\xi}$, having no fidelity input.
The conditioned surrogate, by contrast, traces the simulator's error curve closely across the entire training support.
At interior $\hat{\xi}$ values unseen during training, the surrogate interpolates between adjacent training levels.
We found the smoothness level of the interpolation to be governed by the sample count of the specific fidelity region as well as the resolution of the conditioning encoding (\cref{app:interpolation_smoothness}).
Towards $\hat{\xi} \to 1$, the multi-fidelity surrogate plateaus above the simulator, reflecting its capacity floor relative to the chosen $\xi_{\text{max}}$ reference~\citep{koehler2025neural}, but reaches a lower error than the surrogate trained on high-fidelity samples only.
Beyond the training support ($\hat{\xi} > 1$), the error rises again.

\section{Pareto-optimal data scaling}
\label{sec:experiments}
Having validated our framing of multi-fidelity training in \cref{sec:validating_mf_framework}, we now present empirical evidence that multi-fidelity training enables Pareto-optimal data generation cost scaling across \acp{PDE}, discretization types, error sources, and surrogate architectures.
We evaluate the following claims:

\begin{enumerate}
    \item[\textbf{C1.}] \textbf{Nontrivial optimum under finite
    budget.} For $B < B_{\max}$ (a problem dependent
    crossover), the compute-optimal mixture is in the
    interior, i.e., $
        \mathcal{E}\bigl(\lambda^{\star}(B), B\bigr)
        <
        \mathcal{E}(+\infty, B)
    $.
    \item[\textbf{C2.}] \textbf{Asymptotic agreement with single
    fidelity.} As $B \to \infty$,
    $\lambda^{\star}(B) \to +\infty$ and
    $\mathcal{E}(\lambda^{\star}(B), B) \to \mathcal{E}(+\infty, B)$.
    The scaling curves converge at large budgets.
\end{enumerate}
\begin{table}[htbp]
\centering
\caption{
Overview of \acp{PDE}, data representations, sources of error and architectures.
}
\label{tab:experiments_overview}
\resizebox{\textwidth}{!}{
\begin{tabular}{@{} l l l c c l @{}}
\toprule
\multirow{2}{*}{\textbf{Application domain}} & \multirow{2}{*}{\textbf{Physics / Equation}} & \multirow{2}{*}{\textbf{Data format}} & \multicolumn{2}{c}{\textbf{Error source}} & \multirow{2}{*}{\textbf{Architectures}} \\
\cmidrule(lr){4-5}
& & & \textbf{Discretization} & \textbf{Solver} & \\
\midrule
\textbf{Mathematical / PDEs} & Poisson equation & 1D, 2D, 3D grids & \checkmark & \checkmark & FNO, U-Net \\
\midrule
\multirow{2}{*}{\textbf{Structural Mechanics}} & Linear elasticity & 2D meshes & \checkmark & \checkmark & AB-UPT, Transolver \\
                             & Hyperelasticity (nonlinear) & 2D meshes & \checkmark & \checkmark & AB-UPT, Transolver \\
\midrule
\textbf{Fluid Dynamics}      & RANS (airfoils) & 2D meshes & \checkmark & --$^{\star}$ & AB-UPT, Transolver \\
\bottomrule
\end{tabular}
}
\\[2pt]
{\footnotesize $^{\star}$ Solver error not studied for airfoils, for a discussion, see \cref{app:airfoil}.}
\end{table}

We verify these claims across the experiments summarized in \cref{tab:experiments_overview}.
For this, we test across two data representations, regular grids and irregular meshes, with architectures appropriate to each: \ac{FNO}~\citep{li2021fno} and U-Net~\citep{ronneberger2015unet} for grids, and Transolver~\citep{wu2024transolver} and AB-UPT~\citep{alkin2025abupt} for meshes.
\cref{sec:experiments_poisson} starts with the Poisson equation on regular grids, where moderate experimental cost allows us to span 1D to 3D cases.
\cref{sec:experiments_structural_mechanics} then moves to unstructured meshes for structural mechanics, covering both a linear and nonlinear case.
Finally, \cref{sec:experiments_airfoil} demonstrates a real-world example of data generation cost savings for external aerodynamics around airfoils.
All experiments are repeated across multiple random seeds for training (and dataset generation, if feasible).
We report seed means $\pm1$ std bands and bootstrap confidence intervals for the savings estimates.
Throughout \cref{sec:experiments_poisson,sec:experiments_structural_mechanics} we use budget in algorithmic units (nodes or iterations) rather than wall-clock time.
Under our otherwise fixed solver configurations, these
give hardware-independent, reproducible budgets.
\cref{sec:experiments_airfoil} instead uses CPU core-hours as a more interpretable unit.
In \cref{sec:lambda_heuristic}, we then test an approach for determining the optimal fidelity mixture at larger budgets from a few calibration runs.

\subsection{Poisson equation as a testbed for large parametric studies}
\label{sec:experiments_poisson}

\begin{figure}
    \centering
    \includegraphics[width=\linewidth]{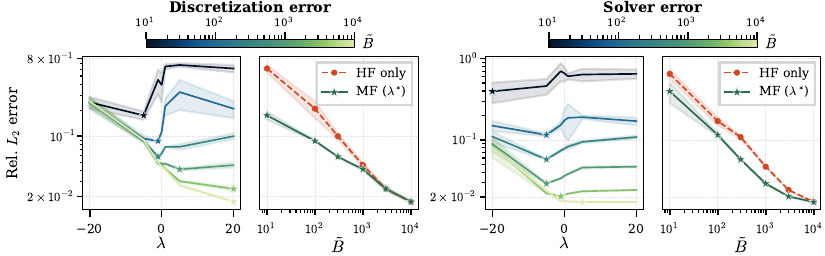}
    \caption{\textbf{Poisson 3D experiment} results for the \textbf{U-Net model}. Each group shows iso-budget curves over the mixture hyperparameter $\lambda$ (left) and the compute-optimal Pareto front using $\lambda^{\star}(B)$ at each budget (right). Multi-fidelity training enables Pareto-optimal scaling for both sources of error.
    Results are averaged across seeds with shaded bands indicating $\pm 1$ standard deviation.
    }
    \label{fig:poisson_2d_overview}
    \vspace{-10pt}
\end{figure}

The Poisson equation is fundamental to many scientific simulations, appearing both as a standalone problem and as a substep in larger pipelines (e.g., incompressibility projection~\citep{chorin1968numerical}).
Moreover, we see it as a suitable testbed for surrogates trying to approximate stationary problems since it ensures global spatial dependencies.
We study the discretization error, controlled by the total number of grid points $N$, and the solver error, controlled by the number of multigrid V-cycles $K$.
For a more detailed description see \cref{app:poisson}.

\cref{fig:poisson_2d_overview} reveals that the minima in iso-budget curves lie in the interior until $B_{\text{max}}$ and shift towards higher $\lambda^*$ with growing normalized budgets $\tilde{B}$, confirming claims \textbf{C1} and \textbf{C2}.
In 3D, the discretization menu spans a $216\times$ cost ratio $c(\xi_{\max}) / c(\xi_{\min})$, against only $4\times$ on the solver axis.
This drives a wider gap between \ac{HF} and \ac{MF} at any fixed error threshold on the discretization axis with $4.6\times$ over only $2.7\times$ in the solver error axis (see \cref{fig:savings_summary}).  At a moderate budget of $\tilde{B} = 300$, the 3D U-Net achieves $46.9\%$\,/\,$42.2\%$ error reduction on the solver\,/\,discretization axes via MF training by drawing roughly $3\times$\,/\,$6\times$ training samples more than the high-fidelity baseline (\cref{tab:poisson_truncation,tab:poisson_discretization}).

Across all 1D and 2D experiments, the U-Net reaches a $2.0$--$3.1\times$ lower terminal error (at the highest $\tilde{B}$ tested) than the FNO and the FNO recovers only $57$--$90\%$ of the U-Net's relative \ac{MF} gain.
This suggests that the magnitude of the \ac{MF} benefit depends on how far the surrogate sits from its capacity floor (full plots and numerical tables are in \cref{app:poisson_additional_results}).

\subsection{Structural mechanics on unstructured meshes}
\label{sec:experiments_structural_mechanics}
As our next experiment, we consider the equations of elasticity, central to many structural mechanics applications.
Compared to the Poisson setup, we introduce three sources of additional complexity to resemble real-world scenarios: complex geometries discretized with unstructured meshes rather than regular grids, the von Mises stress as the prediction target (a derived quantity with lower convergence rate than the displacement field it is computed from), and material nonlinearity via a Neo-Hookean constitutive model for the hyperelastic case.
As before, we vary the number of mesh nodes $N$ for the discretization axis and the number of solver iterations $K$ for the solver axis (see \cref{app:linear_elasticity,app:hyperelasticity} for governing equations, geometries, and error analyses).

\begin{figure}
    \centering
    \includegraphics[width=\linewidth]{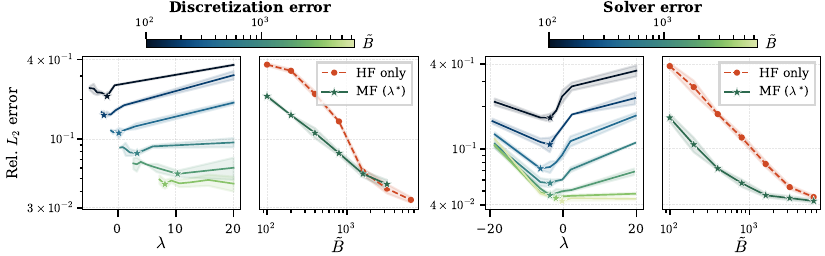}
    \caption{\textbf{Hyperelasticity experiment} results for the \textbf{Transolver model}. Multi-fidelity training also enables data savings in these cases on unstructured meshes with nonlinear \acp{PDE}. Solver error savings are higher (larger distance between the two scaling curves) due to operating at a consistent resolution.
    }
    \label{fig:hyperelasticity_overview}
    \vspace{-10pt}
\end{figure}

\cref{fig:hyperelasticity_overview} shows the multi-fidelity data scaling results for the hyperelasticity experiment with the Transolver model.
In general, we observe the same trends as presented above.
For both error sources, the optimal $\lambda^*$ lies in the interior at moderate budgets and shifts towards higher fidelities as $\tilde{B}$ increases, consistent with \textbf{C1} and \textbf{C2}.
At a moderate budget of $\tilde{B}\approx 400$, MF training reduces the test error by $59\%~/~49.4\%$ on the solver~/~discretization axes by drawing $8\times/~10.4\times$ more training samples than the \ac{HF} baseline (see \cref{tab:hyperelasticity_discretization,tab:hyperelasticity_solver}).

The cross-axis comparison, however, differs from the Poisson case.
Although the cost ratio is much larger for the discretization error than for the solver error ($138.7$ vs. $9.5$ due to polynomial versus exponential error scaling and linear cost scaling for both, see \cref{app:hyperelasticity,tab:hyperelasticity-fidelity-menus}), the surrogate cannot exploit this asymmetry.
The discretization axis yields only a $2.1\times$ data generation cost reduction at $1 \times 10^{-1}$
test error, while the solver axis yields a $4.8\times$ reduction at the same error level (see \cref{fig:savings_summary}).
We attribute this to the general difficulties of having neural surrogates generalize across a large range (here two orders of magnitude) of resolution, as reported in prior literature~\citep{sakarvadia2026false,gao2025discretization,hu2026transolverislineartransformer}. This suggests that multi-fidelity over solver error, which we are the first to investigate, is the practically more suitable axis.

The same trends as presented above also hold for the AB-UPT model on this setup and both models on the easier setup of linear elasticity, where MF training delivers up to $6.3\times$ cost savings on the solver axis at the same $10\%$ threshold (see \cref{fig:savings_summary})
The full plots and numerical results can be found in \cref{app:hyperelasticity_additional_results,app:linear_elasticity_additional_results}.

\subsection{Towards a real use-case in external aerodynamics}
\label{sec:experiments_airfoil}

\begin{figure}
    \centering
    \includegraphics[width=\linewidth]{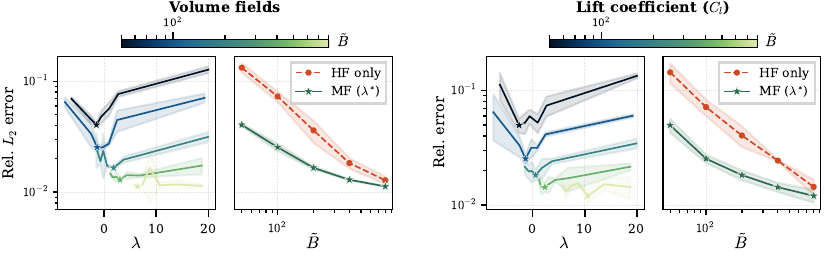}
    \caption{\textbf{Airfoil experiment} results on the \textbf{discretization error} for the \textbf{AB-UPT model}.
    Multi-fidelity training cuts simulation time by $1,820$ core-hours at a matched volumetric error of $2\%$. The savings also extend to derived engineering quantities like the lift coefficient (right plot).
    }
    \label{fig:airfoil_results_condensed}
    \vspace{-10pt}
\end{figure}

We now turn to an external aerodynamics setting, where neural surrogates have direct application in shape optimization for drag and lift behavior.
Based on the work of \citet{bonnet2022airfrans}, we simulate 2D flows past airfoils using the \ac{RANS} equations, varying the number of mesh nodes $N$ to construct a multi-fidelity dataset.
The surrogate predicts the pressure $p$ and velocity field $\mathbf{u} = [u_x, u_y]$ in the volume, as well as pressure $p$ and wall shear-stress $\boldsymbol{\tau} = [\tau_x, \tau_y]$ on the airfoil surface.
We measure dataset generation compute in wall-clock simulation time.
Generating all samples at the highest fidelity consumed approximately 13,580 CPU core-hours.

\cref{fig:airfoil_results_condensed} confirms \textbf{C1} and \textbf{C2} in this setting.
At an average $2\%$ relative $L_2$  error threshold (in the volumetric fields), multi-fidelity training reduces data generation cost by $2.4\times$/~$2.5\times$ for Transolver / 
AB-UPT (\cref{fig:savings_summary}).
This shrinks data generation time from the equivalent of two days (on our 64-core CPU) to under a day.
At a moderate budget of $\tilde{B} \approx 200$, AB-UPT cuts the relative $L_2$ error by $54.1\%$ on the volume fields (drawing $2.9\times$ more samples than the HF baseline) and by $60.3\%$ on the surface fields (\cref{tab:airfoil_volume,tab:airfoil_surface}).
The lift and drag coefficients, which are of direct engineering interest being the typical objectives in shape-optimization pipelines, also follow the multi-fidelity Pareto front (\cref{fig:airfoil_coefficients_overview}).
This indicates that the data generation savings transfer from the predicted fields to the downstream design metrics used in practice.

\subsection{Extrapolating optimal fidelity mixtures across budgets: a post-hoc analysis}
\label{sec:lambda_heuristic}

The previous sections clearly demonstrate the benefit of multi-fidelity training.
A remaining question is how one can, \emph{a priori}, determine the optimal fidelity mixture $\lambda^{\star}$ for a given training data generation compute budget $B$.
Previous work in neural scaling laws for language modeling~\citep{kaplan2020scaling,hoffman2022chinchilla} was able to give a functional form with universal parameters that extrapolated and transferred well due to strongly canonical settings (similar data, architectures and single-epoch training).
In scientific ML, however, every \ac{PDE}, geometry, approximation, discretization, solver, and surrogate architecture reshapes the optimal fidelity mixture landscape making them distinct for each specific case.
We therefore propose and test a calibration recipe rather than a universal law, using the simple Ansatz
\begin{equation}
\label{eq:lambda_star_functional_form}
    \lambda^{\star}(\tilde{B}) = a + b \log(\tilde{B}) \ .
\end{equation}
We outline the specifics of the fitting process and details on how to make it robust in \cref{app:fitting_details} as well as the overhead cost of the approach in \cref{app:lambda_fitting_overhead_cost}.

To test whether the coefficients calibrated at small budgets extrapolate to substantially larger ones, \cref{tab:regret_overview} gives a summary of the extrapolation capabilities for our mesh-based experiments (detailed results across all settings can be found in \cref{app:lambda_heuristic_detailed_results}).
We measure predictive power using regret $r(\tilde{B})= \mathcal{E}\bigl(\lambda_{\text{pred}}(\tilde{B}), \tilde{B}\bigr) / \mathcal{E}\bigl(\lambda^{\star}(\tilde{B}), \tilde{B}\bigr) - 1$, with $\lambda_{\text{pred}}$ being the predicted optimal mixing parameter.
\begin{table}[htbp]
\centering
\caption{Mean held-out regret (\%) of the extrapolation recipe per dataset and fidelity axis (mesh experiments) across the first three held-out budgets (ascending, means taken over architectures).}
\label{tab:regret_overview}
\begin{tabular}{lrrrrrr}
\toprule
\multirow{2}{*}{\textbf{Dataset}} & \multicolumn{3}{c}{\textbf{Solver error}} & \multicolumn{3}{c}{\textbf{Discretization error}} \\
\cmidrule(lr){2-4} \cmidrule(lr){5-7}
 & \textbf{1\textsuperscript{st}} & \textbf{2\textsuperscript{nd}} & \textbf{3\textsuperscript{rd}} & \textbf{1\textsuperscript{st}} & \textbf{2\textsuperscript{nd}} & \textbf{3\textsuperscript{rd}} \\
\midrule
Linear elasticity & 4.0 & 5.2 & 4.4 & 6.3 & 4.5 & 4.0 \\
Hyperelasticity & 2.3 & 3.4 & 3.9 & 9.0 & 10.0 & 4.3 \\
Airfoil & -- & -- & -- & 0.0 & 1.7 & -- \\
\bottomrule
\end{tabular}
\end{table}

The average regret across all extrapolation budgets is $<5\%$ on the solver error setups and $\leq 10\%$ on the mesh discretization settings, showing that the recipe yields good extrapolation at small calibration costs.
For almost all extrapolated mixtures $\lambda$, the error, albeit showing regret, is lower than the corresponding HF-only surrogate (except for some largest budget cases where the two scaling curves already converged (as per claim \textbf{C2})).
We report and discuss the full results and limitations of the recipe in \cref{app:lambda_heuristic_detailed_results}.

\section{Conclusion, limitations and future work}
\label{sec:conclusion}

\paragraph{Limitations and outlook.}
\label{sec:limitations-outlook}

Our experiments assume
\begin{enumerate*}[label=(\roman*)]
    \item that data generation cost dominates training cost, and 
    \item that the models are large enough and trained long enough to absorb the additional data.
\end{enumerate*}
To probe this, we provide ablations of increasing model size and longer training durations in \cref{app:model-size-ablation,app:train-duration-ablation} showing that our findings are robust across setups.
Our extrapolative recipe works well in most settings but degrades on some discretization error settings. 
A complete scaling law analysis would need to jointly vary model size, training compute, dataset generation compute and dataset fidelity mixture across all problems and sources of error, a combinatorial search space that is currently intractable.
This opens up two research directions that could improve mixture predictions avoiding this expensive scaling study:
\begin{enumerate*}[label=(\roman*)]
    \item a theoretical framework integrating both the approximation behavior of neural surrogates, so far studied in restricted settings and single fidelity data \citep{kovachki2021errorbounds}, with the numerical convergence behavior of the underlying simulator, and
    \item active learning \citep{musekamp2025activelearning} to discover optimal mixtures directly rather than extrapolating them.
\end{enumerate*} 
Moreover, our study covers (statistical-)steady-state problems at moderate scale.
Extending the framework to complex 3D geometries and industrial-scale meshes remains open.
Considering transient dynamics we see two natural extensions: autoregressive surrogates at a fixed time step can directly use our existing fidelity axes, while the time step itself could form an additional fidelity axis, provided the architecture handles varying step sizes, which may raise challenges similar to those we observed for discretization error on unstructured meshes.
As outlined in \cref{sec:data_generation_pipeline}, the numerical pipeline also introduces \emph{modeling error} (e.g., turbulence closures, boundary layer treatment), which offers additional fidelity axes that our framework could accommodate.
In the context of foundation models, this motivates extending our scalar fidelity conditioning to a conditioning vector that captures the full simulation context, and studying multi-fidelity scaling in this richer, but compute intensive setting.
Finally, while our approach already yields consistent improvements, more sophisticated distributions, for instance bimodal distributions, could be used for sampling.
Specialized fidelity fusion methods beyond our conditioning, as well as training strategies that go beyond na\"ive dataset mixing, such as fidelity scheduling or pretraining and finetuning, also represent promising directions for future work.

\paragraph{Conclusion.}

\begin{figure}
    \centering
    \includegraphics[width=\linewidth]{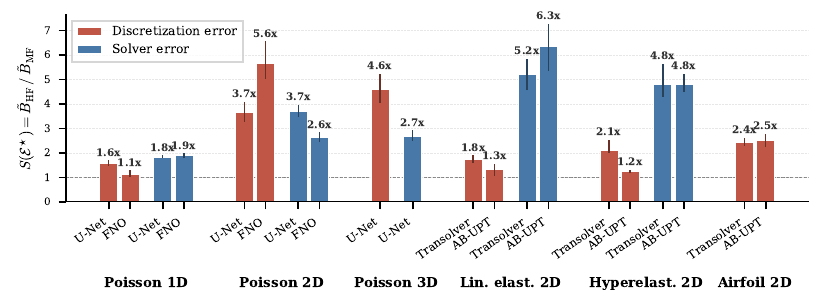}
    \caption{Cost-savings ratio $S(\mathcal{E}^\star)$ of multi-fidelity training relative to high-fidelity training. Relative $L_2$ error thresholds: $\mathcal{E}^\star=10\%$ for Poisson, linear elasticity, and hyperelasticity, and $\mathcal{E}^\star=2\%$ for the airfoil experiment (volume fields). Multi-fidelity training yields cost savings in every one of the 20 setups, up to $6.3\times$ with a mean of ${\sim}3.1\times$. Bar heights are bootstrap medians and whiskers denote 95\% confidence intervals, all excluding $1\times$. We describe the postprocessing and bootstrapping procedure in \cref{app:bootstrapping_details}.}
    \label{fig:savings_summary}
    \vspace{-10pt}
\end{figure}

Our empirical results consistently show that, at any finite data generation budget, training on optimal mixtures of simulation fidelities outperforms training on high-fidelity data alone, with the two approaches converging only at very large budgets.
This highlights that our multi-fidelity savings are not unique to one setup, but hold
across different neural surrogate 
architectures and error sources and, more importantly, across diverse 
physical domains encompassing different PDEs, spatial discretizations (finite differences, finite volumes, and finite elements) and 
numerical solvers.
At a matched test error, 
multi-fidelity training reduces the required data generation budget by an 
average of ${\sim}3.1\times$ across the $20$ setups summarized in 
\cref{fig:savings_summary}, with peaks up to $6.3\times$.
Concretely, on the airfoil case this turns ${\sim}2$ days of high-fidelity data generation on a 64-core CPU into ${\sim}19$ hours.

Beyond raw cost savings, our experiments reveal a practical asymmetry: multi-fidelity over solver error consistently outperforms multi-fidelity over discretization on problems \emph{requiring resolution invariance} and is also simpler to deploy since it changes only a single solver parameter rather than requiring a robust variable-resolution meshing pipeline.
We therefore recommend the solver axis as the natural starting point for practitioners.

The largest multi-fidelity gains for the presented datasets arise in the low-to-medium budget regime ($\tilde{B} \approx 100\text{--}500$ equivalent high-fidelity samples), exactly where high-fidelity data is scarce and the amortization point determines whether a surrogate is cost-effective at all.
At $\tilde{B} \approx 200$ on the airfoil case (AB-UPT), for example, multi-fidelity training reduces the relative lift coefficient error by $54.9\%$ relative to the high-fidelity baseline.
As the SciML community moves towards foundation-scale surrogates, simulation budgets increasingly dominate the total compute spend.
Established high-fidelity surrogate datasets such as DrivAerML~\citep{ashton2024drivearml}, containing $\sim 500$ high-fidelity samples, sit in the regime where our approach delivers its largest gains.
Treating sample fidelity as a first-class design variable rather than a single fixed setting offers a concrete and broadly applicable path to scaling these models at substantially lower data generation cost.

\newpage
\section*{Acknowledgements}
The authors thank Gianluca Galletti for the constructive discussions in the initial phase of the project.

The ELLIS Unit Linz, the LIT AI Lab, the Institute for Machine Learning, are supported by the Federal State Upper Austria. We thank the projects FWF AIRI FG 9-N (10.55776/FG9), AI4GreenHeatingGrids (FFG- 899943), Stars4Waters (HORIZON-CL6-2021-CLIMATE-01-01), FWF Bilateral Artificial Intelligence (10.55776/COE12). We thank NXAI GmbH, Silicon Austria Labs (SAL), Merck Healthcare KGaA, GLS (Univ. Waterloo), T\"{U}V Holding GmbH, Software Competence Center Hagenberg GmbH, dSPACE GmbH, TRUMPF SE + Co. KG.
We acknowledge EuroHPC Joint Undertaking for access to MareNostrum5 at BSC, Spain, Vega at IZUM, Slovenia, Karolina at IT4Innovations, Czech Republic, MeluXina at LuxProvide, Luxembourg, and LEONARDO, hosted by CINECA (Italy) and the LEONARDO consortium.
The authors gratefully acknowledge the scientific support and HPC resources provided by the Erlangen National High Performance Computing Center (NHR@FAU) of the Friedrich-Alexander-Universität Erlangen-Nürnberg (FAU) under the NHR project b278bb. NHR funding is provided by federal
and Bavarian state authorities. The authors also gratefully
acknowledge the computational and data resources as well
as the support provided by the Leibniz Supercomputing
Centre.
Felix Koehler and Nils Thuerey acknowledge funding from the Munich
Center for Machine Learning (MCML).

\bibliography{references}
\bibliographystyle{plainnat}


\newpage
\appendix
\section*{Appendix}

\section{Reproducibility Statement}
\label{app:reproducibility-statement}

In this appendix, we elaborate on the experimental details of our experiments. Specifically, we extend upon our used formalism for multi-fidelity training in \cref{app:extended-formalism}. In \cref{app:poisson,app:linear_elasticity,app:hyperelasticity,app:airfoil} we discuss problem formulations for the Poisson, linear elasticity, hyperelasticity, and airfoil experiments, respectively. This includes a description of the numerical methods used in data generation and precisely how the discretization and solver error is augmented in each case, including a discussion of their characteristic behaviors.
In \cref{app:grid-architectures-and-training,app:mesh-architectures-and-training}, we discuss architecture configurations and training details as well as details on the chosen fidelity menus.
In \cref{app:poisson_additional_results,app:linear_elasticity_additional_results,app:hyperelasticity_additional_results,app:airfoil_additional_results} we present all data from our extensive combinatorial studies including isobudget curves and derived scaling behavior.
We supplement this with numeric tables that allow for reproducing data generation compute savings or error improvement claims made in the main part.
Additionally, we provide ablations in \cref{app:ablations} supporting the design choices made throughout this work.

Our full set of experiments exclusively relies on open-source software.
Links to the source code are found on the first page.

\section{AI Disclosure Statement}

LLM systems have been involved at various stages of this research.
Coding agents (Claude Opus 4.8 to Fable 5.1) were used for workflow orchestration (SLURM job submissions, data aggregation, improvements to visualizations, etc.).
Core algorithmic code was written in collaboration with AI agents and completion systems but has always received full human oversight.
LLMs (mostly Claude Opus 4.8, 5, and Gemini Pro 3.1) have also been used for light editing of the manuscript.
The core text is human written.
Some additional citations in the literature review were found and analyzed using LLM tools.
The authors read and verified every cited work.
Finally, before submission, we gathered feedback on the draft from several AI systems (Claude Opus 4.8, Gemini 3.1 Pro, Grok 3, GPT-5.4, and Mistral Medium 3.5).
The authors take full responsibility for all content of this paper.

\section{Extended Formalism}
\label{app:extended-formalism}

\subsection{Designing the fidelity support}
\label{app:fidelity_support}
\paragraph{The accuracy oracle.}
As introduced in \cref{sec:data_generation_pipeline}, we define the relative $L_2$ error of our simulator at a certain fidelity via $e(\xi)$, defined in \cref{eq:numerical_error}.
To be precise, this is an \emph{accuracy oracle}.\footnote{In practice, we might realize via this sufficiently high $\xi$ that we consider a good reference. This can be the highest fidelity $\xi_{\text{max}}$. In our experiments, this varies based on feasibility. For more details, see \cref{app:grid-architectures-and-training,app:mesh-architectures-and-training}.}
Asymptotically, the behavior $e(\xi)$ will match the theoretical error convergence of the underlying problem.
In infinite precision, we expect 
\begin{equation}
    \lim_{\xi \to \infty}e(\xi) = 0.
\end{equation}
Realistically, though, we will have 
\begin{equation}
    \lim_{\xi \to \infty}e(\xi) \approx \epsilon_{\text{mach}},
\end{equation}
in which $\epsilon_{\text{mach}}$ is the noise floor of the considered floating point accuracy.
Denote by $\xi_\epsilon$ the fidelity level at which we first reach the machine precision.
This gives a first hard ceiling on $\Xi$, namely $\xi_{\max} \leq \xi_{\epsilon}$.
However we discuss practical rules for the floor and ceiling of $\Xi$ below, which we will see are much tighter.

\paragraph{Choosing the right support.}
The support $\Xi$ is itself a hyperparameter of the multi-fidelity procedure, and it cannot be optimized away by a subsequent choice of $\lambda$.
The attainable per-sample error and the minimum per-sample cost are bounded below by $\Xi$ alone,
\begin{equation}
    \min_{\xi \in \mathrm{supp}(p)} e(\xi)
    \;\geq\; e\bigl(\max(\Xi)\bigr),
    \qquad
    \min_{\xi \in \mathrm{supp}(p)} c(\xi)
    \;\geq\; c\bigl(\min(\Xi)\bigr),
\end{equation}
so $\Xi$ defines the feasibility envelope within which $\lambda$ navigates.
A mixture experiment trained over a poorly chosen $\Xi$ cannot be rescued by subsequent tuning of $\lambda$.
We therefore treat choosing the endpoints of $\Xi$ as an important design choice.

\begin{figure}[t]
    \centering
    \begin{tikzpicture}[
        >={Stealth[length=2.5mm]},
        brace/.style={decorate, decoration={brace, amplitude=5pt, raise=2pt}},
        menudot/.style={circle, fill=black, inner sep=1.3pt},
        descr/.style={font=\footnotesize, align=center},
    ]
        \def\Xstart{0}    \def\Xmin{3}     \def\Xmax{8}
        \def\Xeps{12}     \def\Y{0}
        \draw[gray!70, dashed, thick] (\Xstart,\Y) -- (\Xmin,\Y);
        \draw[thick]                  (\Xmin,\Y)   -- (\Xmax,\Y);
        \draw[gray!70, dashed, thick] (\Xmax,\Y)   -- (\Xeps,\Y);
        \draw[->, thick] (\Xeps,\Y) -- (\Xeps+0.8,\Y)
            node[right, font=\small] {$\xi$};
        \foreach \x in {3, 4.25, 5.5, 6.75, 8}
            \node[menudot] at (\x,\Y) {};
        \node[menudot] at (\Xeps,\Y) {};
        \draw[brace] (\Xstart,0.25) -- (\Xmin-0.05,0.25)
            node[midway, above=8pt, descr]
                {$\Xi_{\mathrm{bottom}}$\\sub-crude:\\tag is noise};
        \draw[brace] (\Xmin,0.25) -- (\Xmax,0.25)
            node[midway, above=8pt, descr]
                {$\Xi$\\admissible menu\\(discrete)};
        \draw[brace] (\Xmax+0.05,0.25) -- (\Xeps,0.25)
            node[midway, above=8pt, descr]
                {$\Xi_{\mathrm{top}}$\\over-solved:\\NN cannot distinguish};
        \node[below=6pt, align=center, font=\small] at (\Xmin,\Y)
            {$\xi_{\min}$\\[1pt]
             \footnotesize $e(\xi_{\min}) \approx 20\%$\\[-1pt]
             \footnotesize\itshape (typical)};
        \node[below=6pt, align=center, font=\small] at (\Xmax,\Y)
            {$\xi_{\max}$\\[1pt]
             \footnotesize $e(\xi_{\max}) \approx 0.5\%$\\[-1pt]
             \footnotesize\itshape (typical)};
        \node[below=6pt, align=center, font=\small] at (\Xeps,\Y)
            {$\xi_{\epsilon}$\\[1pt]
             \footnotesize $e(\xi_{\epsilon}) = \varepsilon_{\mathrm{mach}}$};
        \draw[<-, gray!70] (\Xstart,-1.7) -- (\Xmin+1,-1.7);
        \node[right, font=\footnotesize, gray!70] at (\Xmin+1,-1.7)
            {high error, cheap};
        \draw[->, gray!70] (\Xmax-1,-2.15) -- (\Xeps,-2.15);
        \node[left, font=\footnotesize, gray!70] at (\Xmax-1,-2.15)
            {low error, expensive};
    \end{tikzpicture}
    \caption{Schematic arrangement of the fidelity menu $\Xi$ on the
    fidelity axis. The admissible support (solid segment, five
    illustrative dots) sits between the two excluded tails
    $\Xi_{\mathrm{bottom}}$ (high-error, sub-crude: the fidelity tag
    carries no signal) and $\Xi_{\mathrm{top}}$ (low-error,
    over-solved relative to the surrogate floor). The
    machine-precision fidelity $\xi_{\epsilon}$ defines the far end
    of $\Xi_{\mathrm{top}}$. Concrete error values on the two
    endpoints are the \emph{typical} targets we use in practice
    ($e(\xi_{\min}) \approx 20\%$ and $e(\xi_{\max}) \approx 0.5\%$).}
    \label{fig:fidelity-menu}
\end{figure}
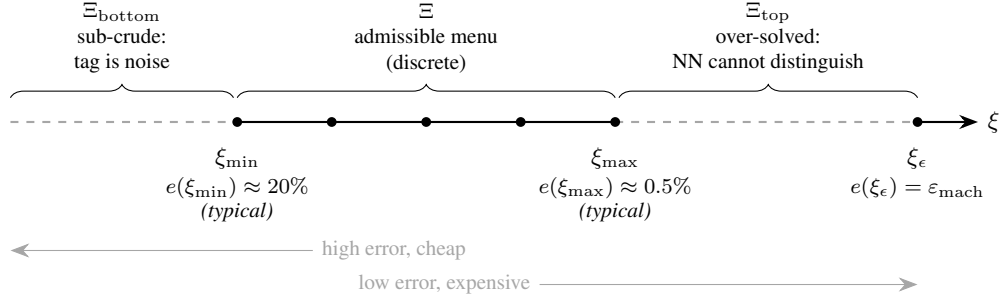

It is convenient to name the two regions the menu deliberately excludes.
On the high-error side lies
\begin{equation}
    \Xi_{\mathrm{bottom}}
    \;=\; \{\,\xi \,:\, \xi < \xi_{\min}\,\},
\end{equation}
where the solver is cruder than our lowest anchor and its outputs are noise-dominated and do not provide meaningful learning signal.
On the low-error side lies
\begin{equation}
    \Xi_{\mathrm{top}}
    \;=\; \{\,\xi \,:\, \xi_{\max} < \xi \le \xi_{\epsilon}\,\},
\end{equation}
where the solver is \emph{more} accurate than the surrogate's training floor.
Figure~\ref{fig:fidelity-menu} illustrates the arrangement schematically.
We now discuss how to position $\Xi$ between $\Xi_{\mathrm{top}}$ and $\Xi_{\mathrm{bottom}}$ in practice.
In \cref{app:endpoint-ablation}, we investigate the placement of the fidelity menu ablationally.

\paragraph{Top anchor $\xi_{\max}$: slightly below the surrogate floor.}
Choose $\xi_{\max} := \max(\Xi)$ so that the simulator's error $e(\xi_{\max})$ is well below the test error the surrogate is expected to achieve, but not so far below that additional fidelity yields no learnable signal.
If $e(\xi_{\max})$ is too low, this would \emph{not} lower the test error
because the surrogate cannot distinguish simulations at $\xi_{\max}$ from those at slightly cheaper fidelities.
The probability mass placed there is wasted computation~\citep{bhatia2025prdp}.
In practice, this is hard to define a priori without an extensive dataset generation and fidelity sweep.
As a proxy, we note that state-of-the-art neural surrogates achieve relative $L_2$ errors in the range of $10^{-1}-10^{-3}$ for settings similar to our experiments~\citep{li2021fno,bonnet2022airfrans,wu2024transolver,alkin2025abupt}.
Therefore, we target $e(\xi_{\max})=0.5\%$ across all our experiments.
This places $\xi_{\max}$ far below $\xi_{\epsilon}$, showing that the surrogate exhausts its capacity long before the solver exhausts its floating-point budget.

\paragraph{Bottom anchor $\xi_{\min}$: crude but informative.}
Choose $\xi_{\min} := \min(\Xi)$ at the lowest fidelity whose output
still carries a measurable signal about the input-to-output map, typically
\begin{equation}
    e(\xi_{\min}) \;\in\; [1\times10^{-1},\, 3\times10^{-1}],
\end{equation}
Two failure modes bracket the correct choice. If $e(\xi_{\min})$ is too high, the solver output is noise-dominated and provides no learnable signal. If $e(\xi_{\min})$ is too low, $\xi_{\min}$ is too close to $\xi_{\max}$ in error space. As such, the  $\lambda < 0$ regime then yields many almost-high-fidelity samples rather than genuinely cheap ones, eliminating the cost-quality tradeoff that motivates multi-fidelity training in the first place.
Empirically, $e(\xi_{\min}) \approx 20\%$ sits comfortably between these regimes across all our experiments.

\paragraph{Axis asymmetry.}
The two fidelity regimes introduced in \cref{sec:data_generation_pipeline} reach the top anchor very differently.
For axes with \emph{exponentially linear}
error
scaling (iterative solvers), over-solving is cheap and the risk is placing $\xi_{\max}$ too far above the surrogate floor, wasting a long tail of the menu on cells the optimizer cannot use.
For axes with \emph{algebraic} 
error
scaling (spatial discretization), over-solving is structurally expensive.
Here, the largest practically reachable $\xi$ is often the correct $\xi_{\max}$, even when $e(\xi_{\max})$ sits above the surrogate floor.

\subsection{Fidelity normalization and clamping}
\label{app:fidelity-normalization}

To ensure consistent behavior across different fidelity menus we use the following normalization:
\begin{equation}
    \hat{\xi}(\xi) = \frac{e(\xi_{\min}) - e(\xi)}{e(\xi_{\min}) - e(\xi_{\max})} \in [0,1],
\end{equation}
where $e(\xi)$ follows the asymptotic convergence of the respective error source (\emph{algebraic} or \emph{exponential}).\footnote{Note that to estimate $e(\xi)$ we do not need the $\xi_{\max}$ solution of that specific instance. Rather, we can estimate it based on the error convergence relation of the problem at hand.}

\subsection{Budgeted sampler}
\label{app:budgeted_sampler}

We draw fidelities i.i.d.\ from the categorical distribution induced by $p(\cdot; \lambda, \Xi)$ until the
cumulative cost crosses the budget:
\begin{equation}
    M(\lambda, B)
    = \min\left\{
        m \in \mathbb{N} \,:\, \sum_{j=1}^{m} c(\xi^{(j)}) \ge B
    \right\},
\end{equation}
\begin{equation}
    G(\lambda, B)
    = \bigl\{\xi^{(1)}, \xi^{(2)}, \dots, \xi^{(M)}\bigr\}.
\end{equation}
Both $M$ and $G$ are random (via the fidelity draws); the realized
sample count satisfies
$\mathbb{E}[M] \approx B \big/ \mathbb{E}_{\xi \sim p}[c(\xi)]$, which
grows as $\lambda$ decreases. In other words, $\lambda$ allows trading fidelity in the sample for sample counts.

In some scenarios, we use a \emph{greedy backfill strategy}. This means that we try to fill up leftover budget with cheap samples to ensure we approximately meet budget $B$ in each scenario.

Pairing each fidelity with an independently drawn input
$\mathbf{f}_h^{(j)} \sim \mathcal{F}_h$ gives the training dataset
\begin{equation}
    D(\lambda, B)
    = \Bigl\{
        \bigl(
            \mathbf{f}_h^{(j)},\;
            \mathbf{u}_h^{(j)},\;
            \xi^{(j)}
        \bigr)
    \Bigr\}_{j=1}^{M},
    \qquad
    \mathbf{u}_h^{(j)} = \mathcal{P}_h\bigl(\mathbf{f}_h^{(j)}; \xi^{(j)}\bigr).
\end{equation}

Crucially, each output function $\mathbf{u}_h^{(j)}$ in the dataset is produced at its corresponding fidelity $\xi^{(j)}$.

\subsection{Training objective}
\label{app:train_error}
For a problem with $F$ output fields (=vector components or channels), let $\mathbf{u}_h^{(f)}$ denote the $f$-th field of the simulator's ground truth at fidelity $\xi$ (generated as $\mathbf{u}_h = \mathcal{P}_h(\mathbf{f}_h; \xi)$), and  $\tilde{\mathbf{u}}_h^{(f)} = \mathcal{S}_{\theta}(\mathbf{f}_h; \hat{\xi})^{(f)}$  the corresponding surrogate prediction conditioned on the normalized fidelity $\hat{\xi}$.

For grid-based experiments (\cref{sec:experiments_poisson}), we use the standard mean squared error on normalized predictions and ground truth:
\begin{equation}
    \zeta\bigl(\tilde{\mathbf{u}}_h, \mathbf{u}_h\bigr) 
    \;=\;
    \frac{1}{N}
    \bigl\|
        \tilde{\mathbf{u}}_h - \mathbf{u}_h
    \bigr\|_2^2,
\end{equation}
where $N$ is the total number of grid points.

For mesh-based experiments (\cref{sec:experiments_structural_mechanics,sec:experiments_airfoil}) the training loss is the field-averaged relative $L_2$ error:
\begin{equation}
    \zeta\bigl(\tilde{\mathbf{u}}_h, \mathbf{u}_h\bigr) 
    \;=\; \frac{1}{F} \sum_{f=1}^{F} 
    \frac{
        \bigl\| \tilde{\mathbf{u}}_h^{(f)} - \mathbf{u}_h^{(f)} \bigr\|_2
    }{
        \bigl\| \mathbf{u}_h^{(f)} \bigr\|_2
    }.
\label{eq:rel_l2}
\end{equation}
Single-field problems (e.g. structural mechanics, where only the scalar von Mises stress is predicted) reduce to $F=1$ and recover the standard relative $L_2$ error.

\subsection{Test error}
\label{app:test_error}
We use the same error as defined in \cref{eq:rel_l2} in physical (denormalized) space.

\paragraph{Choice of test reference.}
We use $\mathcal{P}_h(\cdot; \xi_{\max})$ as the reference rather than $\lim_{\xi \to \infty} \mathcal{P}_h(\cdot; \xi)$, so that the residual $e(\xi_{\max})$ is absorbed into the definition of ``ground truth'' and does not contaminate the measurement.
The paper's central quantity is therefore \emph{the surrogate's approximation to the best-in-support label}, isolating the effect of the training-data mixture from the fixed fidelity-ceiling bias.
Replacing the reference with the $\xi \to \infty$ (practically $\xi_{\epsilon})$ limit yields a physically stricter metric but introduces a $\lambda$-independent floor of size $e(\xi_{\max})$; which would not change the ranking of mixtures.
We believe our choice to be practical, since the surrogate generation process naturally inherits some form of errors~\citep{koehler2025neural}. In \cref{app:test-fidelity-ablation} we ablate the influence of changing from $\xi_{\max}$ to $\xi \to \infty$.

\section{Poisson Experiments}
\label{app:poisson}

In this section, we explain the details of the Poisson experiment in 1D, 2D, and 3D. We introduce the continuous description with homogeneous Dirichlet boundaries, the discretization via finite difference stencils, and the solution of the arising linear system by a geometric multigrid scheme.

\subsection{Overview of the Two Types of Errors}

We consider the Poisson equation in $D \in \{1, 2, 3\}$ dimension on $\Omega = (0, L)^D$ with homogeneous Dirichlet boundary conditions

\begin{equation}
    \Delta u = -f \quad \text{on } \Omega, \qquad u|_{\partial\Omega} = 0.
\end{equation}

In a continuous setting, let $\Delta^{-1}$ be the exact solution to the \emph{continuous} problem. Then, we could abstractly write

\begin{equation}
    u = \Delta^{-1}(-f).
\end{equation}

Now we introduce a collection of interior points that make up an equidistant Cartesian grid.
Let $R_N$ be the restriction operator that maps a function $u$ into a discrete representation $\mathbf{u}_h$ evaluated on these grid points.
Now, let $\bm{\Delta}_h$ be the three-point, five-point, or seven-point stencil in 1D, 2D, or 3D, respectively~\citep{leveque2007fdm}.
It shall correctly respect the boundary condition.
Then, $\bm{\Delta}_h^{-1}$ is the \emph{exact} solution to the discrete approximation (i.e., the exact solution to the linear system of equations).
We know based on the consistency error that

\begin{equation}
    \| \bm{\Delta}_h^{-1}(R_N(f)) - R_N (\Delta^{-1}(f)) \|_\infty = \mathcal{O}(1/N^2).
\end{equation}

This is the \emph{spatial discretization error}. We can realize $\bm{\Delta}_h^{-1}$ via the discrete sine transform, a direct linear solver, or a converged iterative solver. Similarly, $R_N \circ \Delta^{-1}$ can be computed analytically, via a spectral solve, or given by a proxy with sufficiently high $N$.

Now, let $\Phi$ be the iterator associated with a reasonable multigrid solver (described in \cref{app:mg}) whose repeated application approximates the inverse \emph{discrete} Laplacian, i.e., 

\begin{equation}
    \underbrace{\left(
        \Phi_h \circ \Phi_h \circ \cdots \circ \Phi_h
    \right)}_{K \text{ times}}(\mathbf{f}_h)
    =
    \Phi_h^K(\mathbf{f}_h)
    \approx
    \bm{\Delta}_h^{-1}(\mathbf{f}_h).
\end{equation}

Its asymptotic behavior is

\begin{equation}
    \| \Phi^k_h(R_N(f)) - \bm{\Delta}_h^{-1}(R_N(f))\|_\infty = \mathcal{O}(\rho^k).
\end{equation}

In summary, when discretely resolving the Poisson equation on a computer, there are two sources of error. First, there is the spatial discretization error which scales polynomially quadratically with the number of points \emph{per dimension} $N$. The second source is the solver error, which in our case is based on the iterative truncation error of a multigrid scheme. This one scales exponentially linearly with convergence rate $\rho$ over iterations $k$.

\begin{figure}
    \centering
    \includegraphics[width=\linewidth]{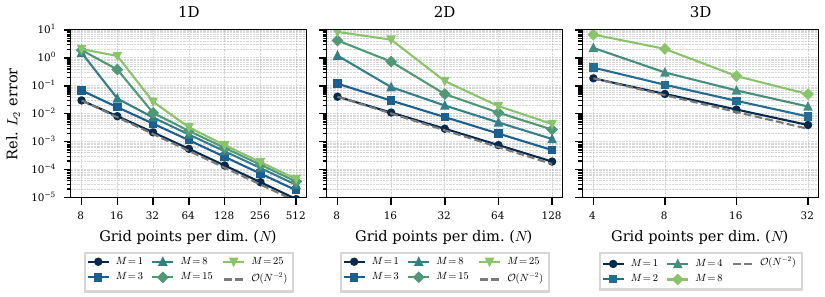}
    \caption{Discretization error convergence for the Poisson problem on a uniform Cartesian grid in 1D, 2D, and 3D. Asymptotically, each configuration reaches the polynomially quadratic decay in the optimality error.}
    \label{fig:poisson-discretization}
\end{figure}

\begin{figure}
    \centering
    \includegraphics[width=\linewidth]{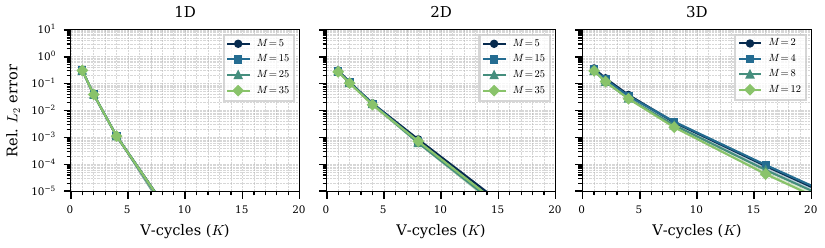}
l    \caption{Multigrid V-cycle iterative truncation error for 1D, 2D, and 3D. Each plot uses a fixed resolution of $64$ \emph{per dimension} (i.e., $64^D$ total degrees of freedom). Different colors indicate different numbers of active modes in the right-hand side distribution.}
    \label{fig:poisson-truncation}
\end{figure}

\subsection{Distribution of Right-Hand Sides}

While the asymptotic behavior of the spatial discretization error and the multigrid truncation error is polynomially quadratically and exponentially linear, respectively, the concrete convergence may be affected by the distribution of right-hand sides $f \sim \mathcal{F}$. Its spectral composition influences the pre-asymptotic regime (which oftentimes is slower than the asymptotic one) and the error offset.
For this experiment we look at distributions $\mathcal{F}_h$ that produce ``spectrally truncated white noise'' with the first $M$ modes being active. In higher dimensions, we use the hypercube of all modes such that their $\|\mathbf{k}\|_1 < M$.
As an example, we show the convergence behavior over the spatial discretization error in \cref{fig:poisson-discretization} over a range of discretization points per dimension $N$ in 1D, 2D, and 3D with different number of active modes. In \cref{fig:poisson-truncation} we display the iterative convergence of the multigrid scheme.

\subsection{Geometric Multigrid V-cycle}
\label{app:mg}

The truncation fidelity axis uses $\xi = K$, the number of V-cycles applied to
an initial zero guess:
\begin{equation}
    \Phi_h^{K}(\mathbf{f}_h)
    = \underbrace{(\Phi_h \circ \cdots \circ \Phi_h)}_{K \text{ times}}(\mathbf{0}).
\end{equation}
A single V-cycle $\Phi_h$ is a standard geometric multigrid scheme~\citep{briggs2000multigrid} that solves
$A_h \mathbf{u}_h = \mathbf{f}_h$ with $A_h = -\bm{\Delta}_h$ and homogeneous Dirichlet boundary conditions enforced by zero-padding throughout.
One dimension-generic implementation services $D \in \{1, 2, 3\}$, with $D$ inferred from the array shape.

\paragraph{Smoother and transfers.} On every level we use red--black
Gauss--Seidel with $\nu_1 = \nu_2 = 2$ pre- and post-smoothing sweeps, and
$\nu_c = 50$ sweeps at the coarsest level (in lieu of a direct solver).
Restriction uses separable full-weighting (the $[1,2,1]/4$ stencil applied
along each axis), and prolongation uses multilinear interpolation. Fine and
coarse interior grids are non-nested under the ``$N+1$ intervals''
convention, so prolongation evaluates the multilinear interpolant at
fractional coarse-padded coordinates rather than using straight injection.

\paragraph{Level hierarchy.} The hierarchy of $J$ levels is generated by full
coarsening, i.e., halving every axis while all current axis lengths are even
and $\geq 8$. At the $N = 64$ resolution used throughout, this yields $J = 5$
levels of shape $(64, 32, 16, 8, 4)$.

\paragraph{Initial guess and cost model.} Each truncation-fidelity sample
restarts from $\mathbf{u}_h = \mathbf{0}$, so $\Phi_h^K$ is a pure function of
$(\mathbf{f}_h, h, K)$ with no warm-start state between samples. Per-call
FLOPs depend only on the (fixed) grid shape and on the MG parameters
$(\nu_1, \nu_2, \nu_c, J)$, so we take $c(K) = K$ and report budgets in units
of V-cycles.

\section{Linear elasticity experiment}
\label{app:linear_elasticity}
This section describes the linear elasticity simulations, which model a perforated structural steel plate under mechanical loading, often found in various civil and mechanical engineering use cases.
We first describe the underlying governing equations and geometrical domain with boundary conditions in \cref{app:linear_elasticity_equations} and \cref{app:linear_elasticity_domain} respectively.
Finally, we provide a detailed discussion of the errors arising in the numerical data generation pipeline in \cref{app:linear_elasticity_errors}.

\subsection{Governing equations}
\label{app:linear_elasticity_equations}

The equations of static linear elasticity for the displacement field $\mathbf{u}$ in the absence of body forces read:
\begin{equation}
    \begin{aligned}
    -\nabla \cdot \boldsymbol{\sigma}(\mathbf{u}) &= \mathbf{0} && \quad \text{in } \Omega \\
    \boldsymbol{\sigma}(\mathbf{u}) &= \lambda \, \text{tr}(\boldsymbol{\varepsilon}(\mathbf{u})) \, \mathbf{I} + 2\mu \, \boldsymbol{\varepsilon}(\mathbf{u}) \\
    \boldsymbol{\varepsilon}(\mathbf{u}) &= \frac{1}{2} \left( \nabla \mathbf{u} + (\nabla \mathbf{u})^\top \right),
    \end{aligned}
\label{eq:linear_elasticity_strong}
\end{equation}
where $\boldsymbol{\sigma}$ is the Cauchy stress tensor, related to the displacement through Hooke's Law for isotropic materials, and $\boldsymbol{\varepsilon}$ is the linearized strain tensor.
Because we model a thin plate, we employ the plane-stress formulation.
The Lam\'{e} parameters, $\lambda$ and $\mu$ are related to Young's modulus $E$ and Poisson's ratio $\nu$ via
\begin{equation}
    \mu = \frac{E}{2(1+\nu)}, \qquad \lambda = \frac{2\mu\nu}{1-\nu}.
\end{equation}
The system is subject to the following boundary conditions:
\begin{equation}
    \begin{aligned}
    -\nabla \cdot \boldsymbol{\sigma}(\mathbf{u}) &= \mathbf{0} && \quad \text{in } \Omega \\
    \mathbf{u} &= \mathbf{0} && \quad \text{on } \Gamma_D \\
    \boldsymbol{\sigma}(\mathbf{u}) \cdot \mathbf{n} &= \mathbf{t} && \quad \text{on } \Gamma_N,
    \end{aligned}
\end{equation}
where $\Gamma_D$ and $\Gamma_N$ are the Dirichlet and Neumann boundaries respectively, $\mathbf{n}$ is the outward-pointing unit normal and $\mathbf{t}$ is the
prescribed surface traction.

This system is cast into a weak variational form and solved computationally using the the \ac{FEM} (for details, see \citep{hughes2012fem, zienkiewicz2013fem}).

\subsection{Domain and boundary conditions}
\label{app:linear_elasticity_domain}
\begin{figure}[htbp]
    \centering
    \begin{minipage}[c]{0.52\textwidth}
        \centering
        \includegraphics[width=\linewidth]{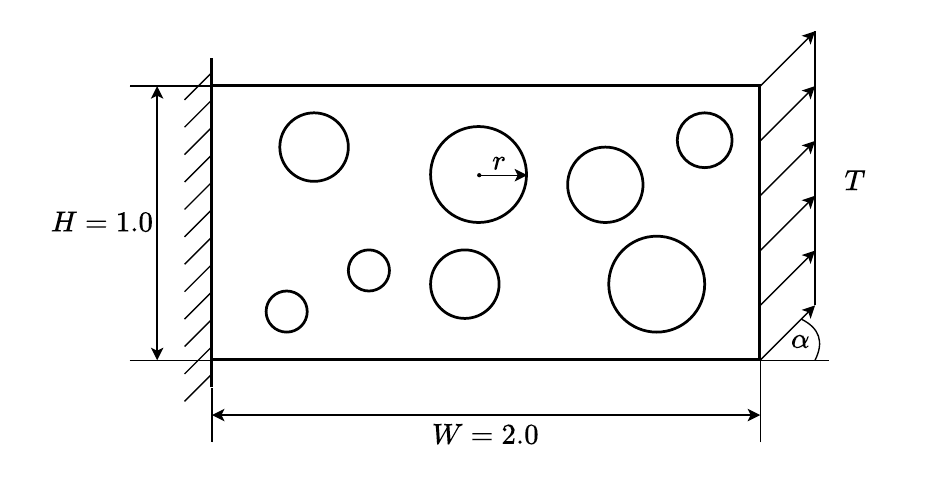}
        \captionof{figure}{Perforated steel plate, clamped on the left and traction on the right boundary.}
        \label{fig:plate_schematic}
    \end{minipage}
    \hfill
    \begin{minipage}[c]{0.44\textwidth}
        \centering
        \small
        \captionof{table}{Sampling ranges for boundary conditions and geometry parameters for the linear elasticity experiment.}
        \begin{tabular}{lll}
            \toprule
            \textbf{Parameter} & \textbf{Symbol} & \textbf{Range} \\
            \midrule
            Traction magnitude & $T$ & $10^5$--$10^7$\,Pa \\
            Traction angle     & $\alpha$ & $0$--$2\pi$ \\
            Number of holes    & --    & $5$--$20$ \\
            Hole radii         & $r$    & $0.02$--$0.15$ \\
            \bottomrule
        \end{tabular}
        \label{tab:plate_params}
    \end{minipage}
\end{figure}

In order to introduce geometric variability, we define $\Omega$ as a 2D plate with randomly placed circular holes.
The plate is clamped on the left edge, and a random traction vector is applied to the right edge.
\cref{fig:plate_schematic} shows a schematic overview of our domain and \cref{tab:plate_params} summarizes the sampling ranges for geometric variations and boundary conditions.
As material coefficients, we use Young's modulus $E = 210$\,GPa and Poisson's ratio $\nu = 0.3$, representative of standard structural steel~(e.g., ASTM A36 or S235).

\subsection{Simulation and postprocessing}
\label{app:linear_elasticity_simulation}
\paragraph{Simulation.}
We solve the weak form of \cref{eq:linear_elasticity_strong} using  linear (P1) Lagrange elements on triangular meshes, generated with \texttt{gmsh}~\citep{geuzaine2009gmsh} via a parametric meshing script with a single continuous parameter controlling mesh fidelity.
The simulation uses the FEniCSx finite-element framework~\citep{baratta2025dolfinx,alnaes2014ufl,scroggs2022basix2,scroggs2022basix1} with PETSc-based linear solvers.
For the discretization error experiments, we eliminate solver error by using a direct LU solver (MUMPS~\citep{amestoy2001mumps}).
For the solver-error experiments, we hold the mesh fixed and use a symmetric Gauss-Seidel iterative scheme to control solver fidelity (\cref{app:linear_elasticity_errors} below).

\paragraph{Postprocessing.}
With P1 elements, the displacement field $\mathbf{u}$ lives natively on mesh nodes, but the stress $\boldsymbol{\sigma}(\mathbf{u})$ and strain $\boldsymbol{\varepsilon}(\mathbf{u})$ fields, being derivatives of $\mathbf{u}$, are piecewise constant and discontinuous across elements.
For neural surrogate training we require nodal values, so we project these fields onto the continuous P1 space via $L_2$ projection.
This projection acts as a smoothing operation, closely related to superconvergent patch recovery techniques~\citep{zienkiewicz1992superconvergent_patch_recovery}.
The von Mises stress, derived from $\boldsymbol{\sigma}$, is similarly projected to P1.
Fields are then stored on the mesh nodes in HDF5 format, together with engineered geometric features (signed distance to the domain boundary, surface normals, nodal-equivalent surface tractions on the loaded edge) used as additional surrogate inputs.

\subsection{Sources of error}
\label{app:linear_elasticity_errors}
\begin{figure}[t]
    \centering
    \begin{subfigure}[b]{0.48\textwidth}
        \centering
        \includegraphics[width=\linewidth]{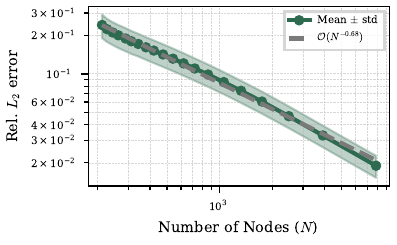}
        \caption{Empirical von Mises stress error convergence with number of mesh nodes $N$. We show the mean and $\pm1$ std across 10,000 simulations.}
        \label{fig:linear_elasticity_discretization_convergence}
    \end{subfigure}
    \hfill
    \begin{subfigure}[b]{0.48\textwidth}
        \centering
        \includegraphics[width=\linewidth]{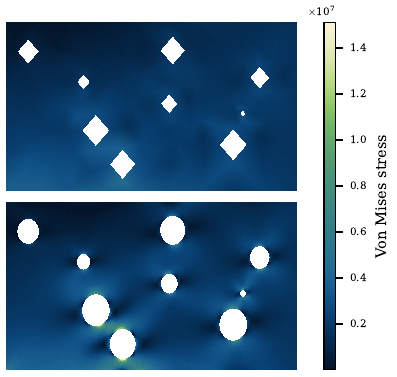}
        \caption{Visualization of the von Mises stress field at the lowest and highest mesh resolution for an example simulation.}
        \label{fig:linear_elasticity_discretization_lf_hf_comparison}
    \end{subfigure}
    \caption{Discretization convergence for the linear elasticity experiment.}
    \label{fig:linear_elasticity_errors}
\end{figure}

\paragraph{Discretization error.}
Let $h$ denote a measure of the spatial discretization, i.e. the average edge length and take the definition of error from \cref{eq:numerical_error}.
For P1 elements (piecewise linear basis) on triangular meshes, the displacement error scales as $e(h) = \mathcal{O}(h^2)$~\citep[Chapter~5]{brenner2008fem}.
Since the number of mesh nodes $N$ scales as $N \propto h^{-2}$ in 2D, this gives $e(N) = \mathcal{O}(N^{-1})$.
However, this optimal rate only applies to the displacement field $\mathbf{u}$.
Theoretically, for the strain $\boldsymbol{\varepsilon}(\mathbf{u})$ and stress $\boldsymbol{\sigma}(\mathbf{u})$ fields, being derivatives of the displacement, the error converges slower: $e(N) = \mathcal{O}(N^{-1/2})$.
The smoothing described in \cref{app:hyperelasticity_errors} partially recovers the lost order of convergence, causing the convergence of the projected von Mises stress field to lie between $\mathcal{O}(N^{-1/2})$ and $\mathcal{O}(N^{-1})$.
Our simulation and postprocessing setup leads to an empirical convergence rate of $e(N) = \mathcal{O}(N^{-0.68})$ as shown in \cref{fig:linear_elasticity_discretization_convergence}.
\cref{fig:linear_elasticity_discretization_lf_hf_comparison} shows von Mises stress fields at the lowest and highest discretization fidelities for an example simulation.

\begin{figure}[htbp]
    \centering
    \includegraphics[width=0.48\linewidth]{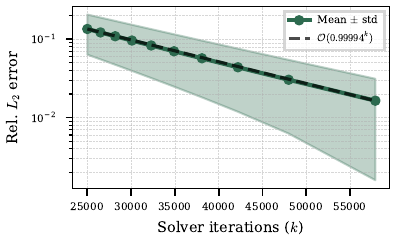}
    \caption{Empirical von Mises stress error convergence with iteration count $k$ for the linear elasticity solver error experiment. We show the mean and $\pm1$ std across 90,000 simulations.}
    \label{fig:linear_elasticity_solver_convergence}
\end{figure}

\begin{figure}[htbp]
    \centering
    \includegraphics[width=\linewidth]{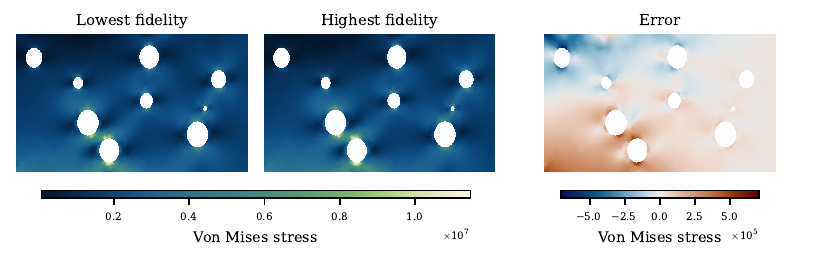}
    \caption{Visualization of the von Mises stress field at the lowest and highest solver fidelities for an example simulation.}
    \label{fig:linear_elasticity_solver_lf_hf_comparison}
\end{figure}

\paragraph{Solver error.}
After spatial discretization of the weak form, the problem reduces to a symmetric positive-definite linear system
\begin{equation}
    \label{eq:linear_system}
    \mathbf{K} \mathbf{u}_h = \mathbf{f}_h,
\end{equation}
where $\mathbf{K} \in \mathbb{R}^{n \times n}$ is the assembled stiffness matrix and $\mathbf{f}_h$ is the assembled load vector.
To study solver error, we hold the mesh fixed and replace the direct LU solver with a symmetric Gauss-Seidel iterative scheme.\footnote{Implemented via PETSc's Richardson iteration with symmetric successive over-relaxation (SSOR) at relaxation factor $\omega = 1.0$, mathematically equivalent to symmetric Gauss-Seidel.}
Starting from a zero initial guess  $\mathbf{u}_h = \mathbf{0}$, the relative $L_2$ error exhibits linear convergence: $e(k)=\mathcal{O}(\rho^k)$, with $k$ being the number of iterations and $\rho \in (0, 1)$ is the spectral radius of the Gauss-Seidel iteration matrix, which depends on the problem and solver settings.
Our setup results in $\rho \approx 0.99994$.
\cref{fig:linear_elasticity_solver_convergence} shows the empirical convergence and \cref{fig:linear_elasticity_solver_lf_hf_comparison} visualizes the von Mises stress field at the lowest and highest solver fidelities.

\section{Hyperelasticity experiment}
\label{app:hyperelasticity}
This section describes the hyperelasticity simulations, which model a perforated elastomeric membrane under tensile loading, representative of components found in e.g., soft-robotic actuators or biomedical seals.
We first describe the underlying governing equations and geometrical domain with boundary conditions in \cref{app:hyperelasticity_equations} and \cref{app:hyperelasticity_domain} respectively.
We then describe the simulation pipeline in \cref{app:hyperelasticity_simulation}.
Finally, we provide a detailed discussion of the errors arising in the numerical data generation pipeline in \cref{app:hyperelasticity_errors}.

\subsection{Governing equations}
\label{app:hyperelasticity_equations}
We consider a compressible Neo-Hookean hyperelastic material in the static regime, without body forces.
Let $\mathbf{u}$ denote the displacement field, $\mathbf{F} = \mathbf{I} + \nabla \mathbf{u}$ the deformation gradient, $\mathbf{C} = \mathbf{F}^\top \mathbf{F}$ the right Cauchy--Green tensor, and $J = \det(\mathbf{F})$ the volume ratio.
The strain energy density is given by
\begin{equation}
    \label{eq:neo_hookean}
    \psi(\mathbf{F}) = \frac{\mu}{2}\bigl(\mathrm{tr}(\mathbf{C}) - 3\bigr) - \mu \ln J + \frac{\lambda}{2}(\ln J)^2,
\end{equation}
where $\mu$ and $\lambda$ are the Lam\'{e} parameters, related to Young's modulus $E$ and Poisson's ratio $\nu$ via
\begin{equation}
    \mu = \frac{E}{2(1+\nu)}, \qquad \lambda = \frac{2\mu\nu}{1-\nu}.
\end{equation}
The first Piola--Kirchhoff stress tensor is obtained as $\mathbf{P} = \partial \psi / \partial \mathbf{F}$, and the Cauchy stress tensor follows from $\boldsymbol{\sigma} = J^{-1} \mathbf{P} \mathbf{F}^\top$.
The equilibrium equation in the reference configuration reads
\begin{equation}
    \label{eq:hyperelasticity_strong}
    -\nabla \cdot \mathbf{P}(\mathbf{u}) = \mathbf{0} \quad \text{in } \Omega,
\end{equation}
subject to boundary conditions
\begin{equation}
    \begin{aligned}
        \mathbf{u} &= \mathbf{0} && \text{on } \Gamma_D, \\
        \mathbf{P}(\mathbf{u}) \cdot \mathbf{n} &= \mathbf{t} && \text{on } \Gamma_N.
    \end{aligned}
\end{equation}
The problem is inherently nonlinear due to the dependence of $\mathbf{P}$ on $\nabla \mathbf{u}$.
A weak form can be derived and solved via the \ac{FEM}~\citep{hughes2012fem, zienkiewicz2013fem}.
The resulting nonlinear system is solved iteratively using Newton-type methods, as described in \cref{app:hyperelasticity_simulation}.

\subsection{Domain and boundary conditions}
\label{app:hyperelasticity_domain}

The domain $\Omega$ is a 2D rectangular plate with randomly placed circular holes, identical in geometry to the linear elasticity setup shown in \cref{fig:plate_schematic}.
Again, the left boundary is clamped and a randomized traction is applied on the right boundary.
Unlike in the linear elasticity case, in this setup compressive loading can cause element inversion due to the geometric nonlinearity.
We therefore restrict the traction direction to a $\pm 30^\circ$ cone around the horizontal axis, ensuring mainly tensile loading.
\cref{tab:hyperelasticity_params} summarizes the sampling ranges.

As material coefficients, we use $E = 2$\,MPa and $\nu = 0.3$, representative of a compressible soft polymer or foam elastomer (e.g., expanded silicone or cellular rubber).
With these parameters, the shear modulus is $\mu \approx 0.77$\,MPa and the traction-to-shear-modulus ratio ranges from $0.001$ to $0.065$, placing the problem in a mildly nonlinear regime.
The resulting Green--Lagrange strains typically remain below $30\%$, confirming that the nonlinear constitutive model is exercised beyond the linear regime.

\begin{table}[htbp]
    \centering
    \small
    \caption{Sampling ranges for boundary conditions and geometry parameters in the hyperelasticity experiment.}
    \begin{tabular}{lll}
        \toprule
        \textbf{Parameter} & \textbf{Symbol} & \textbf{Range} \\
        \midrule
        Traction magnitude & $T$       & $1$--$50$\,kPa \\
        Traction angle     & $\alpha$  & $-\pi/6$--$\pi/6$ \\
        Number of holes    & --        & $5$--$20$ \\
        Hole radii         & $r_i$     & $0.02$--$0.15$ \\
        \bottomrule
    \end{tabular}
    \label{tab:hyperelasticity_params}
\end{table}

\subsection{Simulation and postprocessing}
\label{app:hyperelasticity_simulation}

\paragraph{Simulation.}
We solve the weak form of \cref{eq:hyperelasticity_strong} using linear (P1) Lagrange elements on triangular meshes, generated with \texttt{gmsh}~\citep{geuzaine2009gmsh} via a parametric meshing script with a single continuous parameter controlling mesh fidelity.
The meshing setup mimics that of \cref{app:linear_elasticity_simulation}.
The simulation uses the FEniCSx finite-element framework~\citep{baratta2025dolfinx,alnaes2014ufl,scroggs2022basix2,scroggs2022basix1} with PETSc-based nonlinear solvers (SNES).
Each linearized step uses a direct LU solver (MUMPS~\citep{amestoy2001mumps}).
For the discretization-error experiments, we use a full Newton--Raphson scheme with relative tolerance $10^{-6}$, such that the solver error is negligible relative to the discretization error.
For the solver error experiments, we hold the mesh fixed and use a modified Newton (chord) scheme with frozen Jacobian to control solver fidelity (\cref{app:hyperelasticity_errors} below).

\paragraph{Postprocessing.}
The postprocessing pipeline mimics the one described in \cref{app:linear_elasticity_simulation}.

\subsection{Sources of error}
\label{app:hyperelasticity_errors}

\begin{figure}[t]
    \centering
    \begin{subfigure}[b]{0.48\textwidth}
        \centering
        \includegraphics[width=\linewidth]{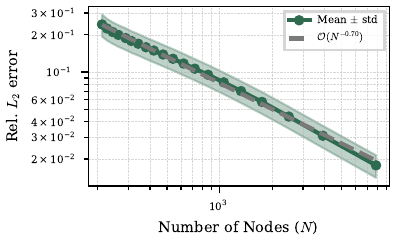}
        \caption{Empirical von Mises stress error convergence with number of mesh nodes $N$. We show the mean and $\pm 1$ std across 10,000 simulations.}
        \label{fig:hyperelasticity_discretization_convergence}
    \end{subfigure}
    \hfill
    \begin{subfigure}[b]{0.48\textwidth}
        \centering
        \includegraphics[width=\linewidth]{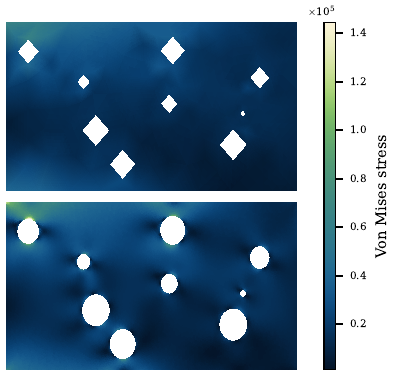}
        \caption{Visualization of the von Mises stress field at the lowest and highest mesh resolution for an example simulation.}
        \label{fig:hyperelasticity_discretization_lf_hf_comparison}
    \end{subfigure}
    \caption{Discretization convergence for the hyperelasticity experiment.}
    \label{fig:hyperelasticity_discretization_errors}
\end{figure}

\paragraph{Discretization error.}
As in the linear elasticity case, the spatial discretization uses P1 Lagrange elements on triangular meshes, controlled by a single continuous fidelity parameter that varies the average edge length.
The theoretical convergence rates carry over from the linear case: $\mathcal{O}(N^{-1})$ for displacement and $\mathcal{O}(N^{-1/2})$ for stress fields in the $L_2$ norm~\citep[Chapter~5]{brenner2008fem}.
The von Mises stress field is postprocessed via $L_2$ projection onto the continuous P1 space, and the empirical convergence rate lies between these two bounds.
Our simulation and postprocessing setup yields an empirical rate of $e(N) = \mathcal{O}(N^{-0.70})$ as shown in \cref{fig:hyperelasticity_discretization_convergence}.
\cref{fig:hyperelasticity_discretization_lf_hf_comparison} shows von Mises stress fields at the lowest and highest discretization fidelities for an example simulation.

\begin{figure}[htbp]
    \centering
    \includegraphics[width=0.48\textwidth]{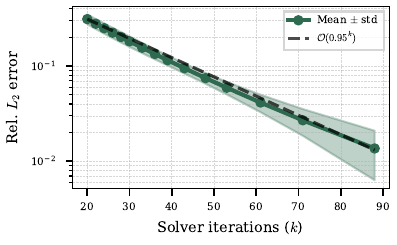}
    \caption{Empirical von Mises stress error convergence with iteration count $k$ for the hyperelasticity solver-error experiment. We show the mean and $\pm 1$ std across 90,000 simulations.}
    \label{fig:hyperelasticity_solver_convergence}
\end{figure}

\begin{figure}[htbp]
    \centering
    \includegraphics[width=\textwidth]{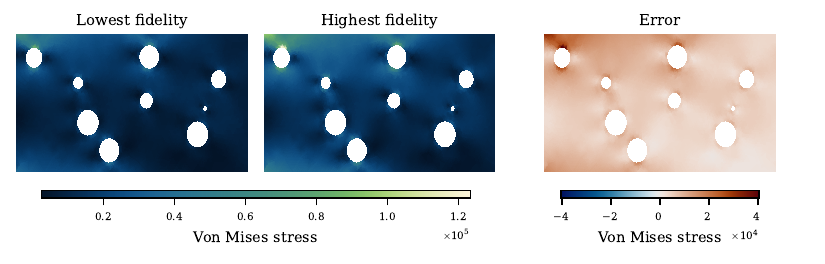}
    \caption{Visualization of the von Mises stress field at the lowest and highest solver fidelities for an example simulation.}
    \label{fig:hyperelasticity_solver_lf_hf_comparison}
\end{figure}

\paragraph{Solver error.}
The nonlinear system arising from the Neo-Hookean model is solved using a modified Newton (chord) method.
The Jacobian is computed once at the initial guess $\mathbf{u}_h^{(0)} = \mathbf{0}$ and frozen for all subsequent iterations.\footnote{Implemented via PETSc's SNES with \texttt{snes\_lag\_jacobian} and \texttt{snes\_lag\_preconditioner} both set to $-2$, with a basic line search and a damped chord method.}
This design ensures linear convergence ($\mathcal{O}(\rho^{k})$) rather than the quadratic convergence of standard Newton--Raphson.
The mesh is held fixed at $N\approx 1{,}600$ mesh nodes across all solver fidelity levels to isolate the solver error axis.
Our setup yields an empirical convergence rate of $\rho \approx 0.95$, as shown in \cref{fig:hyperelasticity_solver_convergence}.
\cref{fig:hyperelasticity_solver_lf_hf_comparison} shows von Mises stress fields at the lowest and highest solver fidelities for an example simulation.

\newpage
\section{Airfoil experiment}
\label{app:airfoil}
This case setup simulates the airflow around 2D wings, a common setting for aerodynamic shape optimization.
We follow the dataset construction pipeline of~\citet{bonnet2022airfrans} and extend it to vary the mesh resolution of the simulations.
We describe the governing equations in \cref{app:airfoil_equations}, the simulation and postprocessing methodology in \cref{app:airfoil_simulation}, the domain and boundary conditions in \cref{app:airfoil_domain}, and the tested error sources in \cref{app:airfoil_errors}.

\subsection{Governing equations}
\label{app:airfoil_equations}
We solve the steady-state, incompressible \ac{RANS} equations; for a
detailed derivation, we refer the reader to
\citet[Appendix F]{bonnet2022airfrans}. The incompressibility assumption
is justified by the subsonic regime (Mach number $< 0.3$) of the
simulations. Denoting ensemble-averaged quantities by $\bar{\,\cdot\,}$,
the equations read:
\begin{equation}
    \begin{aligned}
    \nabla \cdot \bar{\mathbf{u}} &= 0, \\
    \nabla \cdot (\bar{\mathbf{u}} \otimes \bar{\mathbf{u}}) &=
    -\nabla \bar{p} + (\nu + \nu_t)\, \nabla^2 \bar{\mathbf{u}},
    \end{aligned}
\end{equation}
where $\bar{\mathbf{u}}$ is the velocity field, $\bar{p}$ the reduced pressure (the effective pressure divided by the constant fluid density $\rho$), $\nu$ the kinematic viscosity, and $\nu_t$ the turbulent kinematic viscosity.
We use $\rho = 1.184\,\mathrm{kg/m^3}$ and $\nu = 1.56\times 10^{-5}\,\mathrm{m^2/s}$, representing air at sea level and $298.15\,\mathrm{K}$.
The dynamics of $\nu_t$ are governed by additional transport equations.
For closure, we use the $k\text{--}\omega$ SST turbulence model of~\citet{menter2003k_omega_sst}, standard practice in aerodynamics.

The wall shear stress $\boldsymbol{\tau}$ at the airfoil surface is computed from the wall-normal velocity gradient:
\begin{equation}
    \boldsymbol{\tau} = \rho (\nu + \nu_t) \, \nabla \bar{\mathbf{u}}
    \cdot \mathbf{n} \big|_{\partial\Omega_{\text{airfoil}}},
\end{equation}
where $\mathbf{n}$ is the outward-pointing wall-normal unit vector.
The total aerodynamic force on the airfoil is the integral of pressure and shear contributions over the airfoil surface
$\partial\Omega_{\text{airfoil}}$:
\begin{equation}
    \mathbf{F} = \oint_{\partial\Omega_{\text{airfoil}}}
    \left( -\rho\bar{p}\, \mathbf{n} + \boldsymbol{\tau} \right) \, dS.
\end{equation}
Decomposing $\mathbf{F}$ into components parallel and perpendicular to the free-stream velocity $\mathbf{U}_\infty$ yields the drag $D$ and lift $L$, respectively. The dimensionless drag and lift coefficients are then defined as
\begin{equation}
    C_d = \frac{D}{q_\infty A}, \qquad
    C_l = \frac{L}{q_\infty A},
\end{equation}
with dynamic pressure $q_\infty = \tfrac{1}{2} \rho U_\infty^2$ and
chord-based reference area $A = 1\,\mathrm{m^2}$.

\subsection{Domain and boundary conditions}
\label{app:airfoil_domain}
We model airfoil geometries from NASA's NACA 4- and 5-digit series~\citep{cummings2015applied}.
These airfoil families are parametrized by a small set of integer parameters that determine the maximum camber, its chordwise position, and maximum thickness, allowing systematic exploration of a broad design space.
The far-field boundary is placed at 200 chord lengths from the airfoil to minimize boundary-condition effects on the simulation.
\cref{tab:airfoil_params} summarizes the variability in geometry and flow conditions in our dataset.
\begin{table}[htbp]
    \centering
    \small
    \caption{Sampling ranges for flow conditions and airfoil geometry parameters
    in the airfoil experiment. Flow conditions (Reynolds, angle of attack)
    are sampled uniformly from continuous ranges; airfoil shape parameters
    are sampled uniformly and rounded to integers, following the NACA 4- and
    5-digit families. For 4-digit airfoils, the camber position is set to $0$ (uncambered) when the sampled value is below $1.5$, eliminating geometries with maximum camber too close to the leading edge, where NACA 4-digit parametrizations become numerically problematic.}
    \label{tab:airfoil_params}
    \begin{tabular}{lll}
        \toprule
        \textbf{Parameter} & \textbf{Family} & \textbf{Range} \\
        \midrule
        Reynolds number             & --        & $2$--$6\times 10^6$ \\
        Angle of attack             & --        & $-5^\circ$--$15^\circ$ \\
        \midrule
        Max camber ($M$)            & 4-digit   & $\{0, 1, \ldots, 7\}$ \\
        Camber position ($P$)       & 4-digit   & $\{0\} \cup \{2, 3, \ldots, 7\}$ \\
        Max thickness ($XX$)        & 4-digit   & $\{5, 6, \ldots, 20\}$ \\
        \midrule
        Design lift coeff. ($L$)    & 5-digit   & $\{0, 1, 2, 3, 4\}$ \\
        Camber position ($P$)       & 5-digit   & $\{3, 4, \ldots, 8\}$ \\
        Reflex flag ($Q$)           & 5-digit   & $\{0, 1\}$ \\
        Max thickness ($XX$)        & 5-digit   & $\{5, 6, \ldots, 20\}$ \\
        \bottomrule
    \end{tabular}
\end{table}

\subsection{Simulation and postprocessing}
\label{app:airfoil_simulation}
\paragraph{Simulation.}
We compute solutions using the steady-state incompressible \ac{RANS} solver \texttt{simpleFoam} from OpenFOAM~\citep{weller1998openfoam}.\footnote{We use OpenFOAM v2506 with Open MPI 4.1.1 for parallel execution.}
The pressure-velocity coupling uses the SIMPLEC algorithm~\citep{vandoormaal1984simplec}, with under-relaxation factors applied throughout (0.3 for pressure, 0.7 for momentum and turbulence).
Spatial discretization uses Gauss linear schemes for gradients (with cell-based limiters on turbulence quantities for robustness), bounded second-order linearUpwind for convective fluxes, and Gauss linear with limited surface-normal correction for Laplacian terms.
Three non-orthogonal correctors are applied per outer iteration to ensure accuracy on the unstructured mesh.
Linear systems are solved with PCG/DIC for pressure (absolute tolerance $10^{-6}$) and PBiCGStab/DILU for momentum and turbulence quantities (absolute tolerance $10^{-8}$). 
Each simulation is run for $2 \times 10^4$ or $4 \times 10^4$ outer iterations (depending on the angle of attack), providing a fixed iteration count across all cases and fidelities.
All simulations were run on a single AMD EPYC 9655P (96 cores, 192 threads, 4.5\,GHz max boost).
CPU core hours are reported as wall-clock time multiplied by core count.

\paragraph{Postprocessing.}
Following the standard finite-volume convention, OpenFOAM stores field values at cell centers.
We use \texttt{pyvista}~\citep{sullivan2019pyvista} to process the raw simulation data stored in VTK format.
To use these fields as targets for our neural surrogates, which operate on mesh nodes, we project the cell-centered values to mesh points by averaging the values of all cells incident to each point (\texttt{pyvista}'s \texttt{vtkCellDataToPointData} filter).
This is a standard post-processing step in CFD and introduces a small smoothing effect.
The projected fields are stored in HDF5 format.

\subsection{Sources of error}
\label{app:airfoil_errors}
\paragraph{Discretization error.}
We generate meshes ranging from 5{,}200 to 212{,}000 mesh nodes, varying the resolution of both the volume and the airfoil-surface mesh.

\begin{figure}[htbp]
    \centering
    \begin{subfigure}[b]{0.48\textwidth}
        \centering
        \includegraphics[width=\linewidth]{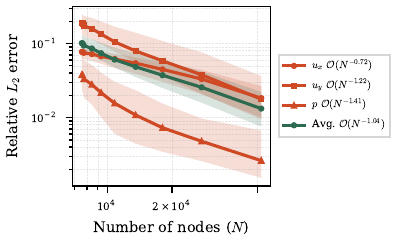}
        \caption{Volumetric fields error convergence.}
        \label{fig:airfoil_discretization_convergence_fields}
    \end{subfigure}
    \hfill
    \begin{subfigure}[b]{0.48\textwidth}
        \centering
        \includegraphics[width=\linewidth]{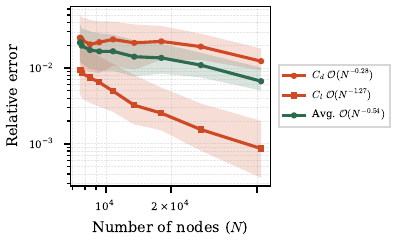}
        \caption{Aerodynamic coefficients error convergence.}
        \label{fig:airfoil_discretization_convergence_coeffs}
    \end{subfigure}
    \caption{Discretization error convergence for the airfoil experiment. We show the median (line) and 10th-90th percentiles (shaded band) across 1,627 simulations.}
    \label{fig:airfoil_discretization_errors}
\end{figure}

\cref{fig:airfoil_discretization_errors} displays the resulting empirical discretization error convergence.
\cref{fig:airfoil_discretization_convergence_fields} shows that each volumetric field exhibits a slightly different error convergence behavior.
However, when averaged across fields, the convergence rate is approximately $e(N) = \mathcal{O}(N^{-1.04})$, consistent with the second-order accuracy of the chosen discretization (again, with $N \propto h^{-2}$)~\citep[Chapter 4.4]{ferziger2019computational}.

\cref{fig:airfoil_discretization_convergence_coeffs} shows the convergence of the integrated quantities.
While the lift coefficient $C_l$ converges at a rate similar to the field errors, the drag coefficient $C_d$ converges substantially slower.

\paragraph{Solver error.}
Unlike the Poisson and structural mechanics setups, we do not study the solver error setting for the airfoil experiments.
The reason is methodological: the SIMPLEC iteration in steady-state \ac{RANS} does not admit a single, well-defined ``solver fidelity'' parameter that exhibits the clean exponential convergence as in the other experiments (\cref{app:poisson,app:linear_elasticity,app:hyperelasticity}).
Each outer SIMPLEC iteration combines several internal components: a pressure Poisson solve (PCG/DIC), velocity and turbulence transport solves (PBiCGStab/DILU), $k$--$\omega$ SST turbulence-model updates, and three non-orthogonal mesh correctors, each with its own iteration counts and tolerances.
The resulting convergence is non-monotonic in the outer iteration count  and depends on multiple knobs that interact non-trivially.
We therefore run all simulations to a fixed iteration count and study fidelity along the spatial discretization axis only.

\section{Architecture and training details (grid experiments)}
\label{app:grid-architectures-and-training}

This section specifies the architectures, the fidelity conditioning mechanism, the training objective and schedule, the data pipeline, and the evaluation protocol used for every grid-based Poisson experiment (\cref{app:poisson}).
The same code path services $D \in \{1,2,3\}$ and both fidelity axes (truncation, discretization). The dimension $D$ and the error axis (discretization or truncation) enter only through the array shapes and through the cost oracle $c(\xi)$. An overview of all hyperparameters is given in \cref{tab:grid-hyperparameters}.
We use a standard FiLM conditioning mechanism~\citep{perez2018film} applied to regular U-Net~\citep{ronneberger2015unet} and FNO~\citep{li2021fno} architectures.

\subsection{Training objective, optimizer, and schedule}
\label{app:grid-training}

Every training run uses the same loss, whitening, optimizer, and schedule; only the data source differs (streaming vs cached, truncation vs discretization axis).

\paragraph{Per-pixel whitening.}
The dataset of input and output pairs is whitened based on the sample mean and standard deviation. We use a warm-up subsample of $n_{w} = 8000$ training pairs drawn uniformly without replacement from the budgeted enumeration (or from the full enumeration if $M \le n_{w}$).

\paragraph{Loss and optimizer.}
The training loss is the per-pixel mean-squared error on the whitened output, over mini-batches of size $|\mathcal{B}| = 20$.
We optimize with Adam (default $\beta_1 = 0.9$, $\beta_2 = 0.999$) under a warm-up-cosine learning-rate schedule.
The learning rate ramps linearly from $0$ to $\eta_{\max} = 10^{-4}$ over the first $S_{\mathrm{warm}} = 2000$ steps and then decays as a cosine to $0$ over the remaining steps up to the total $S_{\mathrm{total}} = 20000$. 
We chose this scheduler as we found it to achieve consistent results across configuration cells.
For an ablation over the training duration, see \cref{app:train-duration-ablation}.

\subsection{Data pipeline}
\label{app:grid-data}
For simplicity we never write datasets to disk. Each new seed experiment also generates a new dataset.

\paragraph{Streaming vs cached generation.}
Two equivalent training paths are supported:
\begin{itemize}
    \item \textbf{Streaming}: each optimizer step regenerates its
          batch procedurally.
          No dataset is
          materialized, and memory is only $\mathcal{O}(|\mathcal{B}|)$. This is
          the default in 2D and is required in 3D for large $M$ (to avoid writing datasets to disk). Note that despite the dataset is procedurally ``re-generated'' each epoch, we ensure it is pseudo-randomly deterministic. As such, this streaming paradigm does not adversarially create more samples than the current experiment tries to investigate.
    \item \textbf{Cached}: the full $M$-sample enumeration is
          pre-materialized once, kept resident on the device
          (GPU or host RAM if it would otherwise exceed VRAM), and
          the training loop indexes into the cache. Cheaper per step
          when the cache fits in memory.
\end{itemize}
Because both paths call the same underlying generator with the same seed schedule, they produce identical training trajectories up to floating-point non-determinism on the device.

\paragraph{Truncation axis.}
A sample at fidelity $\xi = K$ is generated by drawing a random windowed Fourier RHS $\mathbf{f}_h$ (see \cref{app:poisson}) and applying $K$ V-cycles of the proper multigrid iterator described in \cref{app:mg} to a zero initial guess.
All samples live at the target resolution $N_{\star}$. The fidelity tag $\xi$ is the raw integer V-cycle count.

\paragraph{Discretization axis.}
A sample at fidelity $\xi = N$ is generated by sampling $\mathbf{f}_h$ at native resolution $N$, solving $-\Delta_h \mathbf{u}_h = \mathbf{f}_h$ exactly at that resolution (DST direct solver, \cref{app:poisson}), and then interpolating both the RHS and the solution from $(N)^D$ to the target grid $(N_{\star})^D$ (at which the networks are trained) via multilinear interpolation. Interpolating both sides simultaneously keeps the
training pair self-consistent under the target-resolution
$(2D+1)$-point Laplacian. Using this setup, all data samples live at $N_{\star}=64$ resolution but still contain varying degrees of spatial discretization error.
The fidelity tag is the
raw native integer $N$. In \cref{sec:interpolation-error}, we discuss the additional interpolation error this same-resolution approach incurs and how this affects the error improvements of the multi-fidelity approach.

\subsection{Fidelity menus, cost model, and swept budget range}
\label{app:grid-menus-budgets}

This subsection fixes the concrete numerical content of the per-$(D,\text{axis})$ fidelity menus, the cost oracle $c(\xi)$ used by the budgeted sampler, and the normalized-budget range $\tilde B = B / c(\xi_{\max})$ that the grid-Poisson sweeps traverse.
Menus are chosen following the endpoint-anchoring rule from \cref{app:fidelity_support}: $e(\xi_{\min})$ lands in $[1\times10^{-1},\, 3\times10^{-1}]$ (crude-but-informative) and $e(\xi_{\max})$ lands near 0.5\% (slightly below the surrogate floor).

\paragraph{Cost oracles.}
Both axes admit a simple closed-form cost per sample.
\begin{itemize}
    \item \textbf{Truncation.} A sample at fidelity $K$ costs $c(K) = K$
          V-cycles of the proper multigrid iterator (\cref{app:mg}).
          All truncation samples live at the common target resolution
          $N_\star = 64$, so the per-cycle constant $N_\star^D$ is the
          same across every menu entry and factors out of
          $\tilde B$; we report budgets in units of V-cycles.
    \item \textbf{Discretization.} A sample at fidelity $N$ costs
          $c(N) = N^{D}$, i.e.\ the volume of the native solver grid.
          This reflects the scaling of both the DST direct solve
          ($\mathcal{O}(N^{D}\log N)$ up to log factors) and of the
          dominant intermediate array computations. Reported budgets are therefore
          in units of native grid points.
\end{itemize}

\paragraph{Concrete menus.}
\Cref{tab:grid-fidelity-menus} lists the menu $\Xi$, the target per-endpoint errors used for menu anchoring, , and the normalized-budget range actually swept in our experiments.
The boundary hyperparameters $\lambda \to \pm \infty$ are implemented in practice as $\lambda = \pm 20$; on the min-max-clamped softmax this places a near-delta at the corresponding endpoint with a small residual mass leak on the interior cells.

\begin{table}[ht]
    \centering
    \caption{Fidelity menus used for the grid-based Poisson
    experiments.
    Cost $c(\xi)$ is expressed in V-cycles ($K$) for truncation or native
    grid points ($N^{D}$) for discretization. $\tilde B = B /
    c(\xi_{\max})$ is the budget expressed in units of one maximum-fidelity sample. All menus use $N_{\star} = 64$ as the target
    resolution at which the surrogate is trained and evaluated.}
    \label{tab:grid-fidelity-menus}
    \begin{tabular}{llllccr}
        \toprule
        \textbf{Axis} & $\boldsymbol{D}$ & \textbf{Menu} $\Xi$ &
        \textbf{Cost} $c(\xi)$ & $e(\xi_{\min})$ & $e(\xi_{\max})$ &
        $\tilde B$ \textbf{range} \\
        \midrule
        truncation      & 1 & $\{1,2,3\}$                   & $K$       & $\approx 20\%$ & $\approx 1\%$   & $10$–$10^{4}$ \\
        truncation      & 2 & $\{1,2,3,4,5,6,7\}$           & $K$       & $\approx 25\%$ & $\approx 0.5\%$ & $10$–$10^{4}$ \\
        truncation      & 3 & $\{2,3,4,5,6,7,8\}$           & $K$       & $\approx 13\%$ & $\approx 0.4\%$ & $10$–$10^{4}$ \\
        \midrule
        discretization  & 1 & $\{20,28,36,48\}$             & $N^{1}$   & $\approx 25\%$ & $\approx 0.5\%$ & $10$–$10^{4}$ \\
        discretization  & 2 & $\{20,32,48,64,96\}$          & $N^{2}$   & $\approx 30\%$ & $\approx 0.5\%$ & $10$–$10^{4}$ \\
        discretization  & 3 & $\{16,24,32,48,64,96\}$       & $N^{3}$   & $\approx 20\%$ & $\approx 0.5\%$ & $10$–$10^{4}$ \\
        \bottomrule
    \end{tabular}
\end{table}

\newpage
\subsection{Hyperparameter summary}

We summarize all hyperparameters in \cref{tab:grid-hyperparameters}.

\begin{table}[ht]
    \centering
    \small
    \caption{Architecture, optimization, and data hyperparameters
    shared across every grid-based Poisson run. Fidelity menus per
    $(D, \text{axis})$ are listed in \cref{tab:grid-fidelity-menus}.}
    \label{tab:grid-hyperparameters}
    \begin{tabular}{lll}
        \toprule
        \textbf{Group} & \textbf{Symbol / flag} & \textbf{Value} \\
        \midrule
        \multicolumn{3}{l}{\emph{Shared backbone / conditioning}} \\
        Target resolution          & $N_{\star}$        & $64$ (per axis, all $D$) \\
        Conditioning dim           & $d_{\mathrm{cond}}$ & $256$ \\
        Sinusoidal frequencies     & --                  & $16$, log-spaced in $[0.1, 10]$ \\
        Conditioning MLP           & --                  & 2 hidden layers, SiLU \\
        \midrule
        \multicolumn{3}{l}{\emph{FiLM-U-Net}} \\
        Levels                     & $L$                 & $4$ \\
        Base channels              & $C_0$               & $64$ \\
        Channel multiplier         & --                  & $2$ (so $(64, 128, 256, 512)$) \\
        Blocks per level           & --                  & $2$ \\
        Conv kernel                & --                  & $3$ (padding $1$) \\
        Norm                       & --                  & GroupNorm, $\min(8, C)$ groups \\
        Activation                 & --                  & SiLU \\
        Down / up transition       & --                  & strided-2 conv / NN-upsample + conv \\
        Output head                & --                  & $1{\times}1$ conv \\
        \midrule
        \multicolumn{3}{l}{\emph{FiLM-FNO}} \\
        Backbone blocks            & $B$                 & $4$ \\
        Hidden channels            & $H$                 & $64$ \\
        Modes per axis             & $K_{\mathrm{FNO}}$  & $12$ \\
        Dirichlet pad ratio        & $r_{\mathrm{pad}}$  & $1.5$ \\
        Activation                 & --                  & GELU \\
        \midrule
        \multicolumn{3}{l}{\emph{Optimization}} \\
        Optimizer                  & --                  & Adam ($\beta_1{=}0.9$, $\beta_2{=}0.999$) \\
        Peak learning rate         & $\eta_{\max}$       & $10^{-4}$ \\
        Warm-up steps              & $S_{\mathrm{warm}}$ & $2000$ \\
        Total steps                & $S_{\mathrm{total}}$ & $20000$ \\
        Schedule                   & --                  & warm-up + cosine decay to $0$ \\
        Batch size                 & $|\mathcal{B}|$     & $20$ \\
        Loss                       & --                  & MSE on whitened output \\
        Whitening warm-up          & $n_{w}$             & $8000$ samples \\
        \midrule
        \multicolumn{3}{l}{\emph{RHS distribution}} \\
        Modes per axis             & $M$                 & $15$ \\
        Domain extent              & $L$                 & $1.0$ \\
        Window                     & --                  & outer product of 1D Hann \\
        Normalization              & --                  & $\max|\mathbf{f}_h| = 1$ per sample \\
        \midrule
        \multicolumn{3}{l}{\emph{Test}} \\
        Test samples               & --                  & $100$ \\
        Test seed                  & --                  & $42$ \\
        Evaluation fidelity        & --                  & $\xi_{\max}$ \\
        \bottomrule
    \end{tabular}
\end{table}

\pagebreak
\newpage
\newpage
\clearpage
\newpage
\section{Architecture and training details (mesh experiments)}
\label{app:mesh-architectures-and-training}
This section specifies the architectures, fidelity conditioning mechanism, training objective and schedule, data pipeline, and evaluation protocol used for every mesh-based experiment (linear elasticity in \cref{app:linear_elasticity}, hyperelasticity in \cref{app:hyperelasticity}, and external aerodynamics in \cref{app:airfoil}).
The solid-mechanics experiments (linear elasticity, hyperelasticity) share a common set of hyperparameters listed in \cref{tab:mesh-hyperparameters-solid}, while the airfoil experiments use larger model variants and adapted training schedules listed in \cref{tab:mesh-hyperparameters-airfoil}.

We use two transformer based surrogate architectures designed for operator learning on irregular meshes: Transolver~\citep{wu2024transolver} and AB-UPT~\citep{alkin2025abupt}.
Both architectures consume the normalized fidelity $\hat{\xi}$ via a sinusoidal encoding. With $J$ frequency pairs, the encoding is constructed as:

\begin{equation}
\gamma(\hat\xi)=\big[\sin(\omega_0\hat\xi),\dots,\sin(\omega_{J-1}\hat\xi),
                       \cos(\omega_0\hat\xi),\dots,\cos(\omega_{J-1}\hat\xi)\big],
\qquad
\omega_j=\tfrac{2\pi}{\ell_{\min}}\,10^{-4j/J},
\label{eq:sinusoidal_encoding}
\end{equation}

i.e. wavelengths geometrically spaced over $[\ell_{\min},10^{4}\ell_{\min})$ with fixed dynamic range $\Lambda=10^4$.
We use $\ell_{\min}=2\pi/1000$ for our main experiments.
For an ablation on $\ell_{\min}$, see \cref{app:interpolation_smoothness}.
This is followed by an MLP, with the resulting conditioning vector then injected into the transformer blocks via adaLN modulation~\citep{peebles2023dit}.

\subsection{Data normalization}
\label{app:mesh-normalization}
All input and target fields are normalized with per-channel statistics (mean and standard deviation) computed on the training set, and applied as standard z-score normalization. 
Spatial coordinates are min-max scaled to $[-1000, 1000]$ along each axis, using the bounding box of the whole training set.
Statistics are computed once per training run from the budgeted dataset, stored alongside the model checkpoint, and reused at inference time.

\subsection{Training objective, optimizer, and schedule}
\label{app:mesh-training}

\paragraph{Loss.}
The training loss is the relative $L_2$ error (\cref{eq:rel_l2}), averaged over the $F$ output fields.
For single-field problems (von Mises stress in elasticity and hyperelasticity), $F = 1$ recovers the standard relative $L_2$ error.
For airfoil simulations, $F = 5$ covers velocity $\mathbf{u} = (u_x, u_y)$, pressure $p$, and wall shear stress $\boldsymbol{\tau} = (\tau_x, \tau_y)$.
Here, the loss is calculated on the volume and on the surface separately and then averaged with equal weighting as in~\citep{bonnet2022airfrans}.

\paragraph{Optimizer and schedule.}
We use AdamW under a one-cycle learning-rate schedule~\citep{smith2018onecycle} with cosine annealing.
Peak learning rate, total step count, weight decay, and warmup fraction are chosen per experiment family/architecture (see \cref{tab:mesh-hyperparameters-solid,tab:mesh-hyperparameters-airfoil}).

\subsection{Data pipeline}
\label{app:mesh-data}

Mesh datasets are pre-generated and stored on disk in HDF5 format 
(\cref{app:linear_elasticity_simulation,app:hyperelasticity_simulation,app:airfoil_simulation}),with each file containing nodal coordinates, mesh connectivity, field values at each node, the fidelity tag, and per-sample geometric features.
Unlike the grid experiments, mesh-based samples are not procedurally regenerated each epoch, since the high cost of the underlying FEM and CFD simulations would make this prohibitive.

\paragraph{Node subsampling.}
For the structural mechanics experiments, mesh sizes range from $212$ to $29{,}408$, and for the airfoil experiments from $5{,}200$ to $212{,}000$ nodes per sample, exceeding the available GPU memory when loaded at full resolution.
We therefore subsample to $16{,}000$ for the plate experiments and to $32{,}000$ for the airfoil experiments for Transolver training.
For training AB-UPT, this is handled by anchor point sampling (for detailed numbers, see \cref{tab:mesh-hyperparameters-solid,tab:mesh-hyperparameters-airfoil}).
Evaluation, is always performed at full resolution.

\subsection{Fidelity menus, cost model, and swept budget range (linear elasticity)}
\label{app:linear-elasticity-menus-budgets}

This subsection fixes the concrete numerical content of the linear elasticity fidelity menus, the cost oracle $c(\xi)$ used by the budgeted sampler, and the normalized-budget range $\tilde{B} = B / c(\xi_{\max})$ that the linear elasticity sweeps traverse.
Menus are chosen following the endpoint-anchoring rule from \cref{app:fidelity_support}.

\paragraph{Cost oracles.}
Both axes admit a simple closed-form cost per sample.
\begin{itemize}
    \item \textbf{Solver.} A sample at fidelity $K$ costs $c(K) = K$
          Gauss-Seidel iterations (\cref{app:linear_elasticity_errors}).
          The mesh is held fixed across all solver fidelity levels, 
          so the per-iteration constant factors out of $\tilde{B}$; 
          we report budgets in units of solver iterations.
    \item \textbf{Discretization.} A sample at fidelity $N$ costs
          $c(N) = N$, i.e.\ the number of mesh nodes. This reflects 
          the dominant scaling of the direct LU solve and of the 
          assembled stiffness matrix. Reported budgets are therefore 
          in units of mesh nodes.
\end{itemize}

\paragraph{Concrete menus.}
\cref{tab:linear-elasticity-fidelity-menus} lists the menu $\Xi$ (reduced form), the target per-endpoint errors used for menu anchoring, the cost of the most expensive sample, and the normalized-budget range actually swept in our experiments.
The boundary hyperparameters $\lambda \to \pm \infty$ are implemented in practice as $\lambda = \pm 20$.
Note that for the discretization axis, the menu values are continuous (mesh sizes from a parametric meshing script), and we report the rounded mean nodal counts at each fidelity level.

\begin{table}[ht]
    \centering
    \small
    \caption{Fidelity menus used for the linear elasticity 
    experiments. Per-endpoint errors $e(\xi_{\min}), e(\xi_{\max})$ are motivated by the target anchors from the menu-selection rule.
    Cost $c(\xi)$ is expressed in solver iterations ($K$) for the solver axis or mesh nodes ($N$) for the discretization axis.
    $\tilde{B} = B / c(\xi_{\max})$ is the budget expressed in units of one maximum-fidelity sample.}
    \label{tab:linear-elasticity-fidelity-menus}
    \begin{tabular}{lllccr}
        \toprule
        \textbf{Axis} & \textbf{Menu} $\Xi$ ($|\Xi|$) & 
        \textbf{Cost} $c(\xi)$ & $e(\xi_{\min})$ & 
        $e(\xi_{\max})$ & $\tilde{B}$ \textbf{range} \\
        \midrule
        solver         & $\{25{,}000,\, \ldots,\, 262{,}144\}$ ($11$) & $K$ & $\approx 15\%$ & $\approx 0.5\%$ & $100$–$6{,}400$ \\
        \midrule
        discretization & $\{212,\, \ldots,\, 29{,}408\}$ ($21$)       & $N$ & $\approx 25\%$ & $\approx 1\%$   & $100$–$3{,}200$ \\
        \bottomrule
    \end{tabular}
\end{table}

\subsection{Fidelity menus, cost model, and swept budget range (hyperelasticity)}
\label{app:hyperelasticity-menus-budgets}

The hyperelasticity setup uses the same cost oracles and discretization-axis menu as linear elasticity (\cref{app:linear-elasticity-menus-budgets}), with a different solver-axis menu reflecting the different convergence behavior of the modified Newton (chord) method.

\paragraph{Cost oracles.}
The cost oracles are identical to the linear elasticity setup: $c(K) = K$ for the solver axis (where $K$ is the modified Newton iteration count) and $c(N) = N$ for the discretization axis (where $N$ is the number of mesh nodes).

\paragraph{Concrete menus.}
\cref{tab:hyperelasticity-fidelity-menus} lists the menu $\Xi$ (reduced form), the target per-endpoint errors used for menu anchoring, the cost of the most expensive sample, and the normalized-budget range actually swept in our experiments.
The discretization menu is identical to that of linear elasticity (\cref{tab:linear-elasticity-fidelity-menus}); we repeat its endpoints here for completeness. The boundary hyperparameters $\lambda \to \pm \infty$ are implemented in practice as $\lambda = \pm 20$.

\begin{table}[ht]
    \centering
    \small
    \caption{Fidelity menus used for the hyperelasticity 
    experiments. The discretization menu is identical to that of 
    linear elasticity (\cref{tab:linear-elasticity-fidelity-menus}). 
    Cost $c(\xi)$ is expressed in modified Newton iterations 
    ($K$) for the solver axis or mesh nodes ($N$) for the 
    discretization axis.}
    \label{tab:hyperelasticity-fidelity-menus}
    \begin{tabular}{lllccr}
        \toprule
        \textbf{Axis} & \textbf{Menu} $\Xi$ ($|\Xi|$) & 
        \textbf{Cost} $c(\xi)$ & $e(\xi_{\min})$ & 
        $e(\xi_{\max})$ & $\tilde{B}$ \textbf{range} \\
        \midrule
        solver         & $\{20,\, \ldots,\, 190\}$        ($16$) & $K$ & $\approx 30\%$ & $\approx 0.5\%$ & $100$–$6{,}400$ \\
        \midrule
        discretization & $\{212,\, \ldots,\, 29{,}408\}$  ($21$) & $N$ & $\approx 25\%$ & $\approx 1\%$   & $100$–$3{,}200$ \\
        \bottomrule
    \end{tabular}
\end{table}

\subsection{Fidelity menus, cost model, and swept budget range (airfoil)}
\label{app:airfoil-menus-budgets}

The airfoil setup admits only a discretization error fidelity axis (\cref{app:airfoil_errors} discusses why the SIMPLEC iteration does not afford a clean solver error axis).
Unlike the synthetic unit-cost models used for the elasticity setups, here we can measure data generation cost directly in CPU core-hours via the wall-clock simulation time of each sample.

\paragraph{Cost oracle.}
A sample at fidelity $\xi = N$ (mesh node count) costs $c(N) = $ measured CPU core-hours.
Because cost depends on mesh anisotropy, boundary-layer treatment, and convergence behavior in non-trivial ways, we report the empirical mean cost at each fidelity level rather than a closed-form scaling, mimicking real-world simulation pipelines.
Reported budgets are therefore in real CPU core-hours.

\paragraph{Concrete menu.}
\cref{tab:airfoil-fidelity-menus} lists the menu $\Xi$ (reduced form), the target per-endpoint errors used for menu anchoring, and the normalized-budget range swept in our experiments. The menu spans $11$ continuous meshing-fidelity levels, ranging from $\sim$7,600 to $\sim$180,000 mesh nodes per sample, with corresponding mean simulation times from $\sim$0.9 to $\sim$8.3 CPU core-hours.
The boundary hyperparameters $\lambda \to \pm \infty$ are implemented in practice as $\lambda = \pm 20$.

\begin{table}[ht]
    \centering
    \small
    \caption{Fidelity menu used for the airfoil experiments. The 
    cost oracle is the empirically measured wall-clock simulation 
    time, in CPU core-hours; we report the mean across samples at 
    each fidelity level. $e(\xi_{\min})$ and $e(\xi_{\max})$ are 
    the volume-field-averaged relative $L_2$ errors at the menu 
    endpoints.}
    \label{tab:airfoil-fidelity-menus}
    \begin{tabular}{lllccr}
        \toprule
        \textbf{Axis} & \textbf{Menu} $\Xi$ ($|\Xi|$) & 
        \textbf{Cost} $c(\xi)$ & $e(\xi_{\min})$ & 
        $e(\xi_{\max})$ & $\tilde{B}$ \textbf{range} \\
        \midrule
        discretization & $\{7{,}600,\, \ldots,\, 180{,}000\}$ ($11$) & CPU core-hours & $\approx 10\%$ & $\approx 0.5\%$ & $50$–$800$ \\
        \bottomrule
    \end{tabular}
\end{table}

\newpage
\subsection{Hyperparameter summary (solid mechanics)}
\label{app:mesh-hyperparameters-solid}
\begin{table}[ht]
    \centering
    \small
    \caption{Architecture, optimization, and data hyperparameters 
    shared across the linear elasticity and hyperelasticity 
    experiments. The same configuration applies to both 
    discretization error and solver error experiments.}
    \label{tab:mesh-hyperparameters-solid}
    \begin{tabular}{ll}
        \toprule
        \textbf{Group} & \textbf{Value} \\
        \midrule
        \multicolumn{2}{l}{\emph{Transolver}} \\
        Layers                       & $8$ \\
        Hidden dim                   & $128$ \\
        Slice dim                    & $128$ \\
        Attention heads              & $4$ \\
        MLP expansion                & $4$ \\
        Conditioning dim             & $16$ \\
        Dropout                      & $0.1$ \\
        \midrule
        \multicolumn{2}{l}{\emph{AB-UPT}} \\
        Hidden dim                   & $96$ \\
        Attention heads              & $4$ \\
        Geometry depth               & $1$ \\
        Physics blocks               & [perceiver, self, cross, self] \\
        Domain anchor points         & $1024$ per domain \\
        Decoder blocks per domain    & $2$ \\
        Supernodes                   & $256$ \\
        Supernode neighbors          & $32$ \\
        Cross-attention chunk size   & $16{,}000$ \\
        Fidelity embedding dim       & $48$ \\
        Feature embedding dim        & $48$ \\
        \midrule
        \multicolumn{2}{l}{\emph{Optimization}} \\
        Optimizer                    & AdamW ($\beta_1{=}0.9$, $\beta_2{=}0.999$) \\
        Weight decay                 & $5 \times 10^{-4}$ \\
        Peak learning rate           & $2 \times 10^{-3}$ \\
        Schedule                     & one-cycle cosine annealing \\
        Warm-up fraction             & $0.2$ \\
        Total steps                  & $125{,}000$ \\
        Train batch size             & $16$ \\
        Loss                         & per-channel relative $L_2$ \\
        Gradient clipping            & enabled \\
        \midrule
        \multicolumn{2}{l}{\emph{Test}} \\
        Evaluation fidelity          & $\hat{\xi} = 1$ (corresponds to $\xi_{\max}$) \\
        Test samples                 & $1{,}000$ discretization, $4{,}500$ solver\\
        \bottomrule
    \end{tabular}
\end{table}

\newpage
\subsection{Hyperparameter summary (airfoil)}
\label{app:mesh-hyperparameters-airfoil}
\begin{table}[ht]
    \centering
    \small
    \caption{Architecture, optimization, and data hyperparameters 
    for the airfoil experiments.}
    \label{tab:mesh-hyperparameters-airfoil}
    \begin{tabular}{ll}
        \toprule
        \textbf{Group} & \textbf{Value} \\
        \midrule
        \multicolumn{2}{l}{\emph{Transolver (large)}} \\
        Layers                       & $12$ \\
        Hidden dim                   & $256$ \\
        Slice dim                    & $256$ \\
        Attention heads              & $8$ \\
        MLP expansion                & $4$ \\
        Conditioning dim             & $16$ \\
        Dropout                      & $0$ \\
        \midrule
        \multicolumn{2}{l}{\emph{AB-UPT (large)}} \\
        Hidden dim                   & $192$ \\
        Attention heads              & $3$ \\
        Geometry depth               & $1$ \\
        Physics blocks               & [perceiver, self, cross, self, cross, self, cross, self] \\
        Domain anchor points         & $4096$ volume, all surface points \\
        Decoder blocks per domain    & $2$ \\
        Supernodes                   & $256$ \\
        Supernode neighbors          & $32$ \\
        Cross-attention chunk size   & $16{,}384$ \\
        Fidelity embedding dim       & $96$ \\
        Feature embedding dim        & $96$ \\
        Dropout                      & $0$ \\
        \midrule
        \multicolumn{2}{l}{\emph{Optimization}} \\
        Optimizer                    & AdamW ($\beta_1{=}0.9$, $\beta_2{=}0.999$) \\
        Weight decay                 & $10^{-8}$ \\
        Peak learning rate           & $3 \times 10^{-4}$ \\
        Schedule                     & one-cycle cosine annealing \\
        Warm-up fraction             & $0.05$ \\
        Total steps                  & $100{,}000$ \\
        Train batch size             & $2$ \\
        Node subsampling             & $32{,}000$ per sample (training only, Transolver) \\
        Loss                         & per-channel relative $L_2$ (volume + surface) \\
        Gradient clipping            & enabled \\
        \midrule
        \multicolumn{2}{l}{\emph{Test}} \\
        Evaluation fidelity          & $\hat{\xi} = 1$ (corresponds to $\xi_{\max}$) \\
        Test samples                 & 163 \\
        \bottomrule
    \end{tabular}
\end{table}

\clearpage
\newpage
\section{Additional results}
\label{app:additional-results}
This section reports detailed results across all experiments.
We show scaling curves for multi-fidelity vs. high-fidelity training together with their respective numerical tables, as well as the needed computational resources required to produce them.

\subsection{Poisson experiments}
\label{app:poisson_additional_results}

In \cref{tab:compute-resources-poisson}, we list the compute resources used for all Poisson experiments in terms of GPU-hours. Note that since the Poisson experiment procedurally generates all training data during each cell run, we do not explicitly report data generation cost. In total, we consumed $\approx1500$ A100-equivalent GPU-hours ($\approx 62$ A100-equivalent GPU-days). This is dominated by the cost of the 3D experiments. We expect to have spent another equivalent (untracked) amount in exploratory experiments. Only negligible storage usage was required due to procedural data generation. Large-scale array studies allowed us to explore the experimental hypergrid virtually in parallel.

\begin{table}[ht]
\centering
\small
\caption{Compute resources used to produce all experiments related to the Poisson equation in 1D, 2D, and 3D (5{,}450 distinct training runs across the $(\text{axis}, \text{dim}, \text{arch}, \lambda, B, \text{seed})$ grid). The training recipe is fixed across all cells, so per-cell wallclock depends only on (dim, arch, cluster). A 1g.10gb MIG slice uses one seventh of a A100 resource. Cluster systems used : (i) FAU: Alex system at Friedrich-Alexander Universtität Nürnberg, (ii) LRZ: Leibniz Rechenzentrum AI resources MCML partition, (iii) Leonardo: Leonardo (CINECA) Boost partition}
\label{tab:compute-resources-poisson}
\begin{tabular}{lllrrrrr}
\toprule
\textbf{Cluster} & \textbf{Hardware} & \textbf{Dim} & \textbf{Cells} & \textbf{Seeds} & \textbf{Runs} & \textbf{h / run} & \textbf{GPU-h} \\
\midrule
FAU
& A100 80\,GB & 3D & 42 & 1 & 42 & 1--3 & 42--126 \\
LRZ
& A100 80\,GB (1g.10gb) & 1D & 648 & 6 & 3{,}888 & 0.04--0.1 & 156--389 \\
LRZ
& A100 80\,GB (1g.10gb) & 2D & $\approx$\,218 & 6 & 1{,}310 & 0.3--0.6 & 393--786 \\
Leonardo
& A100 64\,GB (slower PCIe) & 3D & $\approx$\,70 & 3 & 210 & 5.5--6 & 1{,}155--1{,}260 \\
\midrule
\textbf{Total} & & & $\approx$ 978 & & \textbf{5{,}450} & & $\approx$ 1{,}746--2{,}561 \\
\textbf{Full-A100} & & & & & & & $\approx$ 1{,}275--1{,}554 \\
\bottomrule
\end{tabular}
\end{table}

\begin{figure}[htbp]
    \centering
    \includegraphics[width=\linewidth]{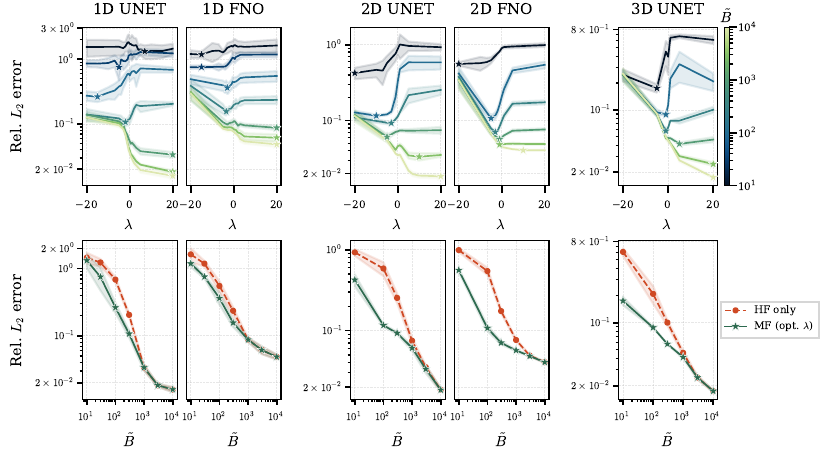}
    \caption{Poisson experiments of multi-fidelity training with discretization error. \textbf{Top:} Iso-budget curves over mixture hyper-parameter $\lambda$ in terms of test error. Minimum loss for each budget is highlighted with $\star$. \textbf{Bottom:} Compute-optimal Scaling behavior using $\lambda^*$ at each budget.}
    \label{fig:poisson_discretization_results_detailed}
\end{figure}

\begin{figure}[htbp]
    \centering
    \includegraphics[width=\linewidth]{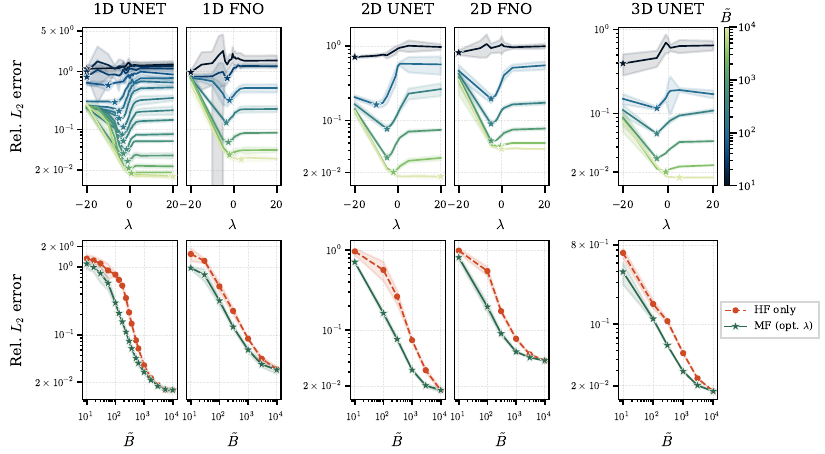}
    \caption{Poisson experiments of multi-fidelity training with truncation error. \textbf{Top:} Iso-budget curves over mixture hyper-parameter $\lambda$ in terms of test error. Minimum loss for each budget is highlighted with $\star$. \textbf{Bottom:} Compute-optimal Scaling behavior using $\lambda^*$ at each budget.}
    \label{fig:poisson_truncation_results_detailed}
\end{figure}

\begin{table}[htbp]
\centering
\caption{Summary of the Poisson experiment results (truncation/solver error). Multi-fidelity (MF) uses the optimal $\lambda^*$; High-fidelity (HF) is the high-fidelity-only baseline ($\lambda \to \infty$). $M_\mathrm{train}$ denotes the number of training samples; $\Delta (\%)$ denotes the relative change in number of training samples and error compared to the high-fidelity baseline.}
\label{tab:poisson_truncation}
\begin{tabular}{llrrrrrrrrrr}
\toprule
\multirow{2}{*}{\textbf{Dim}} & \multirow{2}{*}{\textbf{Arch}} & \multirow{2}{*}{$\bm{\tilde{B}}$} & \multirow{2}{*}{$\bm{\lambda^*}$} & \multicolumn{3}{c}{$\mathbf{M_{\textbf{train}}}$} & \multicolumn{3}{c}{\textbf{Rel.\ $\mathbf{L_2}$ error}} \\
\cmidrule(lr){5-7} \cmidrule(lr){8-10}
 & & & & \textbf{MF} & \textbf{HF} & $\bm{\Delta}$\textbf{(\%)} & \textbf{MF} & \textbf{HF} & $\bm{\Delta}$\textbf{(\%)} \\
\midrule
\multirow{14}{*}{1D} & \multirow{7}{*}{U-Net} & 10 & -20.00 & 30 & 10 & +200.0 & 1.113 & 1.325 & -16.0 \\
 &  & 30 & -20.00 & 90 & 30 & +200.0 & 0.803 & 1.123 & -28.5 \\
 &  & 100 & -7.00 & 292 & 100 & +192.0 & 0.295 & 0.770 & -61.6 \\
 &  & 300 & -3.50 & 751 & 300 & +150.3 & 0.081 & 0.214 & -62.2 \\
 &  & 1,000 & -1.00 & 1,781 & 1,000 & +78.1 & 0.028 & 0.036 & -20.4 \\
 &  & 3,000 & 0.75 & 3,999 & 3,000 & +33.3 & 0.018 & 0.018 & -3.4 \\
 &  & 10,000 & 20.00 & 10,000 & 10,000 & +0.0 & 0.015 & 0.015 & +0.0 \\
\cmidrule(lr){2-10}
 & \multirow{7}{*}{FNO} & 10 & -20.00 & 30 & 10 & +200.0 & 0.967 & 1.549 & -37.6 \\
 &  & 30 & -3.00 & 72 & 30 & +140.0 & 0.766 & 1.231 & -37.8 \\
 &  & 100 & -2.00 & 208 & 100 & +108.0 & 0.318 & 0.517 & -38.6 \\
 &  & 300 & -3.50 & 751 & 300 & +150.3 & 0.131 & 0.226 & -42.0 \\
 &  & 1,000 & -2.50 & 2,256 & 1,000 & +125.6 & 0.060 & 0.088 & -31.5 \\
 &  & 3,000 & -2.00 & 6,310 & 3,000 & +110.3 & 0.037 & 0.044 & -17.3 \\
 &  & 10,000 & 3.50 & 10,687 & 10,000 & +6.9 & 0.030 & 0.031 & -2.4 \\
\midrule
\multirow{12}{*}{2D} & \multirow{6}{*}{U-Net} & 10 & -20.00 & 67 & 10 & +570.0 & 0.711 & 0.968 & -26.5 \\
 &  & 100 & -10.00 & 567 & 101 & +461.4 & 0.163 & 0.566 & -71.2 \\
 &  & 300 & -5.00 & 1,195 & 302 & +295.7 & 0.077 & 0.263 & -70.8 \\
 &  & 1,000 & -5.00 & 3,988 & 1,005 & +296.8 & 0.032 & 0.075 & -57.6 \\
 &  & 3,000 & -2.00 & 7,551 & 3,016 & +150.4 & 0.020 & 0.031 & -35.7 \\
 &  & 10,000 & 20.00 & 10,053 & 10,053 & +0.0 & 0.018 & 0.018 & +0.0 \\
\cmidrule(lr){2-10}
 & \multirow{6}{*}{FNO} & 10 & -20.00 & 67 & 10 & +570.0 & 0.813 & 0.991 & -17.9 \\
 &  & 100 & -5.00 & 402 & 101 & +298.0 & 0.196 & 0.549 & -64.4 \\
 &  & 300 & -5.00 & 1,195 & 302 & +295.7 & 0.090 & 0.173 & -47.7 \\
 &  & 1,000 & -5.00 & 3,988 & 1,005 & +296.8 & 0.054 & 0.078 & -31.4 \\
 &  & 3,000 & -0.50 & 5,711 & 3,016 & +89.4 & 0.045 & 0.050 & -8.8 \\
 &  & 10,000 & 0.00 & 17,475 & 10,053 & +73.8 & 0.042 & 0.042 & -0.1 \\
\midrule
\multirow{6}{*}{3D} & \multirow{6}{*}{U-Net} & 10 & -20.00 & 40 & 10 & +300.0 & 0.396 & 0.649 & -39.1 \\
 &  & 100 & -5.00 & 293 & 101 & +190.1 & 0.115 & 0.170 & -32.3 \\
 &  & 300 & -5.00 & 860 & 302 & +184.8 & 0.058 & 0.109 & -46.9 \\
 &  & 1,000 & -5.00 & 2,898 & 1,005 & +188.4 & 0.029 & 0.047 & -38.1 \\
 &  & 3,000 & -1.00 & 5,495 & 3,014 & +82.3 & 0.020 & 0.024 & -16.8 \\
 &  & 10,000 & 5.00 & 11,031 & 10,047 & +9.8 & 0.017 & 0.017 & -0.3 \\
\bottomrule
\end{tabular}
\end{table}

\begin{table}[htbp]
\centering
\caption{Summary of the Poisson experiment results (discretization error). Multi-fidelity (MF) uses the optimal $\lambda^*$; High-fidelity (HF) is the high-fidelity-only baseline ($\lambda \to \infty$). $M_\mathrm{train}$ denotes the number of training samples; $\Delta (\%)$ denotes the relative change in number of training samples and error compared to the high-fidelity baseline.}
\label{tab:poisson_discretization}
\begin{tabular}{llrrrrrrrrrr}
\toprule
\multirow{2}{*}{\textbf{Dim}} & \multirow{2}{*}{\textbf{Arch}} & \multirow{2}{*}{$\bm{\tilde{B}}$} & \multirow{2}{*}{$\bm{\lambda^*}$} & \multicolumn{3}{c}{$\mathbf{M_{\textbf{train}}}$} & \multicolumn{3}{c}{\textbf{Rel.\ $\mathbf{L_2}$ error}} \\
\cmidrule(lr){5-7} \cmidrule(lr){8-10}
 & & & & \textbf{MF} & \textbf{HF} & $\bm{\Delta}$\textbf{(\%)} & \textbf{MF} & \textbf{HF} & $\bm{\Delta}$\textbf{(\%)} \\
\midrule
\multirow{14}{*}{1D} & \multirow{7}{*}{U-Net} & 10 & 7.00 & 10 & 10 & +0.0 & 1.321 & 1.454 & -9.2 \\
 &  & 30 & -5.00 & 65 & 30 & +116.7 & 0.752 & 1.232 & -39.0 \\
 &  & 100 & -15.00 & 239 & 100 & +139.0 & 0.263 & 0.685 & -61.5 \\
 &  & 300 & -2.00 & 539 & 300 & +79.7 & 0.107 & 0.204 & -47.7 \\
 &  & 1,000 & 20.00 & 1,000 & 1,000 & +0.0 & 0.034 & 0.034 & +0.0 \\
 &  & 3,000 & 20.00 & 3,000 & 3,000 & +0.0 & 0.018 & 0.018 & +0.0 \\
 &  & 10,000 & 20.00 & 10,001 & 10,001 & +0.0 & 0.016 & 0.016 & +0.0 \\
\cmidrule(lr){2-10}
 & \multirow{7}{*}{FNO} & 10 & -15.00 & 24 & 10 & +140.0 & 1.184 & 1.628 & -27.2 \\
 &  & 30 & -15.00 & 72 & 30 & +140.0 & 0.755 & 1.185 & -36.3 \\
 &  & 100 & -3.00 & 194 & 100 & +94.0 & 0.362 & 0.551 & -34.3 \\
 &  & 300 & -3.50 & 603 & 300 & +101.0 & 0.156 & 0.236 & -33.8 \\
 &  & 1,000 & 20.00 & 1,000 & 1,000 & +0.0 & 0.087 & 0.087 & +0.0 \\
 &  & 3,000 & 20.00 & 3,000 & 3,000 & +0.0 & 0.061 & 0.061 & +0.0 \\
 &  & 10,000 & 20.00 & 10,001 & 10,001 & +0.0 & 0.048 & 0.048 & +0.0 \\
\midrule
\multirow{12}{*}{2D} & \multirow{6}{*}{U-Net} & 10 & -20.00 & 214 & 10 & +2040.0 & 0.424 & 0.928 & -54.4 \\
 &  & 100 & -10.00 & 1,668 & 100 & +1568.0 & 0.116 & 0.586 & -80.2 \\
 &  & 300 & -3.00 & 2,020 & 300 & +573.3 & 0.093 & 0.254 & -63.3 \\
 &  & 1,000 & -5.00 & 10,100 & 1,000 & +910.0 & 0.061 & 0.075 & -18.5 \\
 &  & 3,000 & 10.00 & 3,029 & 3,000 & +1.0 & 0.033 & 0.035 & -5.2 \\
 &  & 10,000 & 20.00 & 10,002 & 10,002 & +0.0 & 0.018 & 0.018 & +0.0 \\
\cmidrule(lr){2-10}
 & \multirow{6}{*}{FNO} & 10 & -20.00 & 214 & 10 & +2040.0 & 0.559 & 0.991 & -43.6 \\
 &  & 100 & -5.00 & 1,004 & 100 & +904.0 & 0.107 & 0.544 & -80.3 \\
 &  & 300 & -3.00 & 2,020 & 300 & +573.3 & 0.071 & 0.174 & -59.5 \\
 &  & 1,000 & -1.00 & 3,727 & 1,000 & +272.7 & 0.057 & 0.076 & -25.6 \\
 &  & 3,000 & -1.00 & 11,176 & 3,000 & +272.5 & 0.048 & 0.049 & -1.3 \\
 &  & 10,000 & 10.00 & 10,100 & 10,002 & +1.0 & 0.041 & 0.041 & -0.2 \\
\midrule
\multirow{6}{*}{3D} & \multirow{6}{*}{U-Net} & 10 & -5.00 & 375 & 10 & +3650.0 & 0.175 & 0.611 & -71.4 \\
 &  & 100 & -1.00 & 627 & 100 & +527.0 & 0.088 & 0.209 & -57.7 \\
 &  & 300 & -1.00 & 1,899 & 300 & +533.0 & 0.058 & 0.100 & -42.2 \\
 &  & 1,000 & 5.00 & 1,151 & 1,000 & +15.1 & 0.041 & 0.046 & -10.6 \\
 &  & 3,000 & 20.00 & 3,001 & 3,001 & +0.0 & 0.025 & 0.025 & +0.0 \\
 &  & 10,000 & 20.00 & 10,004 & 10,004 & +0.0 & 0.017 & 0.017 & +0.0 \\
\bottomrule
\end{tabular}
\end{table}

\newpage

\subsection{Hyperelasticity experiments}

\begin{table}[ht]
\centering
\small
\caption{Compute resources used to produce all hyperelasticity experiments, reported as actual measured wall-clock GPU-hours per (architecture, cluster) with the mean per-run time. Partially, runs of one sweep were distributed across sites. The \emph{ablations} rows cover the figure ablations on the hyperelasticity solver axis (no-conditioning, $\ell_{\min}$, binary-vs-spectrum). Cluster systems used: (i) local: in-house GPU servers (NVIDIA H100 80\,GB, A100 80\,GB, A100 40\,GB), (ii) Karolina: IT4Innovations (A100 40\,GB), (iii) Vega: IZUM (A100 40\,GB), (iv) MeluXina: LuxProvide (A100 40\,GB).}
\label{tab:compute-resources-hyperelasticity}
\begin{tabular}{llllrrr}
\toprule
\textbf{Axis} & \textbf{Arch} & \textbf{Cluster} & \textbf{Hardware} & \textbf{Runs} & \textbf{h / run} & \textbf{GPU-h} \\
\midrule
solver & Transolver & local & A100 80 GB & 147 & $\sim$2.3 & $\sim$340 \\
 & Transolver & local & A100 40 GB & 21 & $\sim$2.7 & $\sim$58 \\
 & AB-UPT & local & H100 80 GB & 168 & $\sim$3.0 & $\sim$506 \\
\midrule
discretization & Transolver & Vega & A100 40 GB & 147 & $\sim$11.0 & $\sim$1{,}614 \\
 & AB-UPT & Vega & A100 40 GB & 115 & $\sim$6.6 & $\sim$760 \\
 & AB-UPT & MeluXina & A100 40 GB & 32 & $\sim$6.3 & $\sim$201 \\
\midrule
ablations & Transolver & local & A100 80 GB & 24 & $\sim$4.8 & $\sim$115 \\
 & Transolver & local & H100 80 GB & 57 & $\sim$1.8 & $\sim$96 \\
\midrule
\textbf{Total} &  &  &  & \textbf{711} & & $\sim$3{,}690 \\
\bottomrule
\end{tabular}
\end{table}

\label{app:hyperelasticity_additional_results}
\begin{figure}[htbp]
    \centering
    \includegraphics[width=\linewidth]{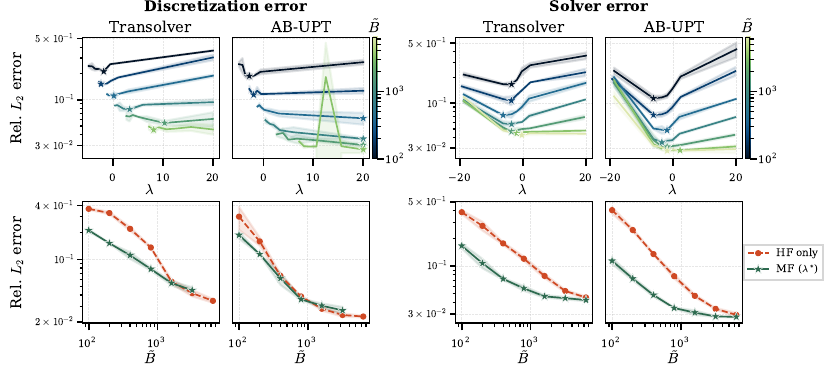}
    \caption{Hyperelasticity experiments. \textbf{Top:} Iso-budget curves over mixture hyper-parameter $\lambda$ in terms of test error. Minimum loss for each budget is highlighted with $\star$. \textbf{Bottom:} Compute-optimal Scaling behavior using $\lambda^*$ at each budget.}
    \label{fig:hyperelasticity_overview-appendix}
\end{figure}

\begin{table}[htbp]
\centering
\caption{Summary of the hyperelasticity experiment results (solver error). Multi-fidelity (MF) uses the optimal $\lambda^*$; High-fidelity (HF) is the high-fidelity-only baseline. $M_\mathrm{train}$ denotes the number of training samples; $\Delta (\%)$ denotes the relative change in number of training samples and error compared to the high-fidelity baseline.}
\label{tab:hyperelasticity_solver}
\begin{tabular}{lrrrrrrrr}
\toprule
\multirow{2}{*}{\textbf{Arch}} & \multirow{2}{*}{$\bm{\tilde{B}}$} & \multirow{2}{*}{$\bm{\lambda^*}$} & \multicolumn{3}{c}{$\mathbf{M_{\textbf{train}}}$} & \multicolumn{3}{c}{\textbf{Rel.\ $\mathbf{L_2}$ error}} \\
\cmidrule(lr){4-6} \cmidrule(lr){7-9}
 & & & \textbf{MF} & \textbf{HF} & $\bm{\Delta}$\textbf{(\%)} & \textbf{MF} & \textbf{HF} & $\bm{\Delta}$\textbf{(\%)} \\
\midrule
\multirow{7}{*}{AB-UPT} & 100 & -6.25 & 791 & 100 & +691.0 & 0.114 & 0.407 & -72.0 \\
 & 200 & -6.25 & 1,582 & 200 & +691.0 & 0.073 & 0.247 & -70.5 \\
 & 400 & -1.95 & 2,104 & 400 & +426.0 & 0.048 & 0.135 & -64.3 \\
 & 800 & -3.59 & 5,303 & 800 & +562.9 & 0.035 & 0.077 & -55.1 \\
 & 1,600 & -1.91 & 8,425 & 1,600 & +426.6 & 0.031 & 0.047 & -34.3 \\
 & 3,200 & 2.19 & 8,386 & 3,200 & +162.1 & 0.028 & 0.034 & -17.4 \\
 & 6,400 & -1.91 & 33,713 & 6,400 & +426.8 & 0.028 & 0.029 & -4.8 \\
\midrule
\multirow{7}{*}{Transolver} & 100 & -3.59 & 660 & 100 & +560.0 & 0.166 & 0.387 & -57.1 \\
 & 200 & -3.67 & 1,325 & 200 & +562.5 & 0.107 & 0.274 & -60.9 \\
 & 400 & -6.25 & 3,188 & 400 & +697.0 & 0.072 & 0.176 & -59.0 \\
 & 800 & -3.59 & 5,303 & 800 & +562.9 & 0.057 & 0.120 & -52.6 \\
 & 1,600 & -3.59 & 10,577 & 1,600 & +561.1 & 0.047 & 0.078 & -40.1 \\
 & 3,200 & -1.95 & 16,916 & 3,200 & +428.6 & 0.045 & 0.053 & -16.4 \\
 & 6,400 & -0.31 & 25,287 & 6,400 & +295.1 & 0.043 & 0.045 & -6.2 \\
\bottomrule
\end{tabular}
\end{table}

\begin{table}[htbp]
\centering
\caption{Summary of the hyperelasticity experiment results (discretization error). Multi-fidelity (MF) uses the optimal $\lambda^*$; High-fidelity (HF) is the high-fidelity-only baseline. $M_\mathrm{train}$ denotes the number of training samples; $\Delta (\%)$ denotes the relative change in number of training samples and error compared to the high-fidelity baseline. Optima in the saturated high-fidelity regime are reported as $\lambda^* \geq 15$ (HF-equivalent).}
\label{tab:hyperelasticity_discretization}
\begin{tabular}{lrrrrrrrr}
\toprule
\multirow{2}{*}{\textbf{Arch}} & \multirow{2}{*}{$\bm{\tilde{B}}$} & \multirow{2}{*}{$\bm{\lambda^*}$} & \multicolumn{3}{c}{$\mathbf{M_{\textbf{train}}}$} & \multicolumn{3}{c}{\textbf{Rel.\ $\mathbf{L_2}$ error}} \\
\cmidrule(lr){4-6} \cmidrule(lr){7-9}
 & & & \textbf{MF} & \textbf{HF} & $\bm{\Delta}$\textbf{(\%)} & \textbf{MF} & \textbf{HF} & $\bm{\Delta}$\textbf{(\%)} \\
\midrule
\multirow{6}{*}{AB-UPT} & 100 & -2.81 & 4,049 & 104 & +3793.3 & 0.187 & 0.300 & -37.8 \\
 & 200 & -1.95 & 5,416 & 206 & +2529.1 & 0.114 & 0.159 & -28.1 \\
 & 400 & $\geq 15$ & 583 & 411 & +41.8 & 0.061 & 0.065 & -6.2 \\
 & 800 & $\geq 15$ & 1,140 & 810 & +40.7 & 0.036 & 0.038 & -7.1 \\
 & 1,600 & $\geq 15$ & 2,279 & 1,605 & +42.0 & 0.030 & 0.028 & +8.2 \\
 & 3,200 & $\geq 15$ & 4,498 & 3,204 & +40.4 & 0.027 & 0.024 & +12.6 \\
\midrule
\multirow{6}{*}{Transolver} & 100 & -1.88 & 2,757 & 104 & +2551.0 & 0.211 & 0.367 & -42.5 \\
 & 200 & -2.42 & 6,710 & 206 & +3157.3 & 0.151 & 0.329 & -54.0 \\
 & 400 & 0.16 & 4,273 & 411 & +939.7 & 0.111 & 0.219 & -49.4 \\
 & 800 & 3.32 & 3,411 & 810 & +321.1 & 0.078 & 0.136 & -42.6 \\
 & 1,600 & 10.31 & 3,195 & 1,605 & +99.1 & 0.054 & 0.056 & -3.8 \\
 & 3,200 & 8.12 & 7,397 & 3,204 & +130.9 & 0.045 & 0.042 & +7.9 \\
\bottomrule
\end{tabular}
\end{table}

\newpage

\subsection{Linear elasticity experiments}
\label{app:linear_elasticity_additional_results}

\begin{table}[ht]
\centering
\small
\caption{Compute resources used to produce all linear elasticity experiments, reported as actual measured wall-clock GPU-hours per (architecture, cluster) with the mean per-run time. Partially, runs of one sweep were distributed across sites. Cluster systems used: (i) local: in-house GPU servers (NVIDIA H100 80\,GB, A100 80\,GB, A100 40\,GB), (ii) Karolina: IT4Innovations (A100 40\,GB), (iii) Vega: IZUM (A100 40\,GB), (iv) MeluXina: LuxProvide (A100 40\,GB).}
\label{tab:compute-resources-linear-elasticity}
\begin{tabular}{llllrrr}
\toprule
\textbf{Axis} & \textbf{Arch} & \textbf{Cluster} & \textbf{Hardware} & \textbf{Runs} & \textbf{h / run} & \textbf{GPU-h} \\
\midrule
solver & Transolver & local & A100 80 GB & 147 & $\sim$2.1 & $\sim$312 \\
 & Transolver & local & A100 40 GB & 21 & $\sim$2.8 & $\sim$59 \\
 & AB-UPT & local & H100 80 GB & 168 & $\sim$2.9 & $\sim$491 \\
\midrule
discretization & Transolver & local & A100 80 GB & 45 & $\sim$7.9 & $\sim$354 \\
 & Transolver & local & A100 40 GB & 56 & $\sim$14.7 & $\sim$822 \\
 & Transolver & Vega & A100 40 GB & 46 & $\sim$13.9 & $\sim$639 \\
 & AB-UPT & local & H100 80 GB & 42 & $\sim$4.3 & $\sim$179 \\
 & AB-UPT & Vega & A100 40 GB & 105 & $\sim$6.5 & $\sim$687 \\
\midrule
\textbf{Total} &  &  &  & \textbf{630} & & $\sim$3{,}544 \\
\bottomrule
\end{tabular}
\end{table}

\begin{figure}[htbp]
    \centering
    \includegraphics[width=\linewidth]{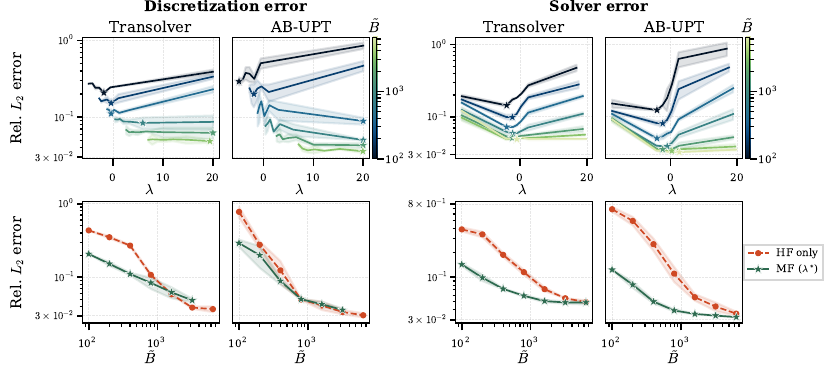}
    \caption{Linear elasticity experiments. \textbf{Top:} Iso-budget curves over mixture hyper-parameter $\lambda$ in terms of test error. Minimum loss for each budget is highlighted with $\star$. \textbf{Bottom:} Compute-optimal Scaling behavior using $\lambda^*$ at each budget.}
    \label{fig:linear_elasticity}
\end{figure}

\begin{table}[htbp]
\centering
\caption{Summary of the linear elasticity experiment results (solver error). Multi-fidelity (MF) uses the optimal $\lambda^*$; High-fidelity (HF) is the high-fidelity-only baseline. $M_\mathrm{train}$ denotes the number of training samples; $\Delta (\%)$ denotes the relative change in number of training samples and error compared to the high-fidelity baseline.}
\label{tab:elasticity_solver}
\begin{tabular}{lrrrrrrrr}
\toprule
\multirow{2}{*}{\textbf{Arch}} & \multirow{2}{*}{$\bm{\tilde{B}}$} & \multirow{2}{*}{$\bm{\lambda^*}$} & \multicolumn{3}{c}{$\mathbf{M_{\textbf{train}}}$} & \multicolumn{3}{c}{\textbf{Rel.\ $\mathbf{L_2}$ error}} \\
\cmidrule(lr){4-6} \cmidrule(lr){7-9}
 & & & \textbf{MF} & \textbf{HF} & $\bm{\Delta}$\textbf{(\%)} & \textbf{MF} & \textbf{HF} & $\bm{\Delta}$\textbf{(\%)} \\
\midrule
\multirow{7}{*}{AB-UPT} & 100 & -4.06 & 879 & 100 & +779.0 & 0.124 & 0.686 & -82.0 \\
 & 200 & -2.34 & 1,455 & 200 & +627.5 & 0.081 & 0.494 & -83.6 \\
 & 400 & -4.06 & 3,542 & 400 & +785.5 & 0.050 & 0.255 & -80.3 \\
 & 800 & -0.98 & 4,598 & 800 & +474.7 & 0.039 & 0.111 & -64.7 \\
 & 1,600 & -2.34 & 11,663 & 1,600 & +628.9 & 0.035 & 0.057 & -37.6 \\
 & 3,200 & 0.39 & 13,464 & 3,200 & +320.8 & 0.034 & 0.044 & -22.7 \\
 & 6,400 & 2.66 & 16,953 & 6,400 & +164.9 & 0.032 & 0.036 & -9.1 \\
\midrule
\multirow{7}{*}{Transolver} & 100 & -4.06 & 879 & 100 & +779.0 & 0.144 & 0.388 & -63.0 \\
 & 200 & -2.34 & 1,455 & 200 & +627.5 & 0.099 & 0.336 & -70.7 \\
 & 400 & -4.06 & 3,542 & 400 & +785.5 & 0.072 & 0.189 & -61.9 \\
 & 800 & -2.27 & 5,812 & 800 & +626.5 & 0.059 & 0.116 & -48.8 \\
 & 1,600 & -2.34 & 11,663 & 1,600 & +628.9 & 0.051 & 0.071 & -28.6 \\
 & 3,200 & -1.04 & 18,433 & 3,200 & +476.0 & 0.049 & 0.055 & -11.2 \\
 & 6,400 & -1.02 & 36,747 & 6,400 & +474.2 & 0.049 & 0.049 & -1.1 \\
\bottomrule
\end{tabular}
\end{table}

\begin{table}[htbp]
\centering
\caption{Summary of the linear elasticity experiment results (discretization error). Multi-fidelity (MF) uses the optimal $\lambda^*$; High-fidelity (HF) is the high-fidelity-only baseline. $M_\mathrm{train}$ denotes the number of training samples; $\Delta (\%)$ denotes the relative change in number of training samples and error compared to the high-fidelity baseline. Optima in the saturated high-fidelity regime are reported as $\lambda^* \geq 15$ (HF-equivalent).}
\label{tab:elasticity_discretization}
\begin{tabular}{lrrrrrrrr}
\toprule
\multirow{2}{*}{\textbf{Arch}} & \multirow{2}{*}{$\bm{\tilde{B}}$} & \multirow{2}{*}{$\bm{\lambda^*}$} & \multicolumn{3}{c}{$\mathbf{M_{\textbf{train}}}$} & \multicolumn{3}{c}{\textbf{Rel.\ $\mathbf{L_2}$ error}} \\
\cmidrule(lr){4-6} \cmidrule(lr){7-9}
 & & & \textbf{MF} & \textbf{HF} & $\bm{\Delta}$\textbf{(\%)} & \textbf{MF} & \textbf{HF} & $\bm{\Delta}$\textbf{(\%)} \\
\midrule
\multirow{6}{*}{AB-UPT} & 100 & -4.92 & 8,000 & 104 & +7592.3 & 0.291 & 0.776 & -62.4 \\
 & 200 & -1.93 & 5,399 & 206 & +2520.9 & 0.200 & 0.276 & -27.8 \\
 & 400 & $\geq 15$ & 565 & 411 & +37.5 & 0.089 & 0.124 & -28.4 \\
 & 800 & $\geq 15$ & 1,114 & 810 & +37.5 & 0.050 & 0.050 & +0.9 \\
 & 1,600 & $\geq 15$ & 2,229 & 1,605 & +38.9 & 0.043 & 0.041 & +4.1 \\
 & 3,200 & $\geq 15$ & 4,401 & 3,204 & +37.4 & 0.036 & 0.034 & +6.6 \\
\midrule
\multirow{6}{*}{Transolver} & 100 & -1.88 & 2,756 & 104 & +2550.0 & 0.208 & 0.434 & -52.2 \\
 & 200 & -0.43 & 2,847 & 206 & +1282.0 & 0.152 & 0.350 & -56.6 \\
 & 400 & -0.43 & 5,479 & 411 & +1233.1 & 0.111 & 0.270 & -58.9 \\
 & 800 & 5.94 & 2,238 & 810 & +176.3 & 0.084 & 0.108 & -22.0 \\
 & 1,600 & $\geq 15$ & 2,229 & 1,605 & +38.9 & 0.062 & 0.059 & +6.2 \\
 & 3,200 & $\geq 15$ & 4,461 & 3,204 & +39.2 & 0.049 & 0.039 & +25.6 \\
\bottomrule
\end{tabular}
\end{table}

\newpage

\subsection{Airfoil experiments}
\label{app:airfoil_additional_results}

\begin{table}[ht]
\centering
\small
\caption{Compute resources used to produce all airfoil experiments, reported as actual measured wall-clock GPU-hours per (architecture, cluster) with the mean per-run time. Partially, runs of one sweep were distributed across sites. Fifteen duplicated AB-UPT HF-only runs on Vega ($\sim$328 GPU-h), superseded by the Karolina re-runs, are excluded. Cluster systems used: (i) local: in-house GPU servers (NVIDIA H100 80\,GB, A100 80\,GB, A100 40\,GB), (ii) Karolina: IT4Innovations (A100 40\,GB), (iii) Vega: IZUM (A100 40\,GB), (iv) MeluXina: LuxProvide (A100 40\,GB).}
\label{tab:compute-resources-airfoil}
\begin{tabular}{lllrrr}
\toprule
\textbf{Arch} & \textbf{Cluster} & \textbf{Hardware} & \textbf{Runs} & \textbf{h / run} & \textbf{GPU-h} \\
\midrule
Transolver & local & A100 80 GB & 42 & $\sim$8.9 & $\sim$374 \\
Transolver & Karolina & A100 40 GB & 78 & $\sim$13.1 & $\sim$1{,}023 \\
AB-UPT & local & H100 80 GB & 21 & $\sim$12.8 & $\sim$269 \\
AB-UPT & Karolina & A100 40 GB & 99 & $\sim$19.9 & $\sim$1{,}974 \\
\midrule
\textbf{Total} &  &  & \textbf{240} & & $\sim$3{,}640 \\
\bottomrule
\end{tabular}
\end{table}

\begin{figure}[htbp]
    \centering
    \includegraphics[width=\linewidth]{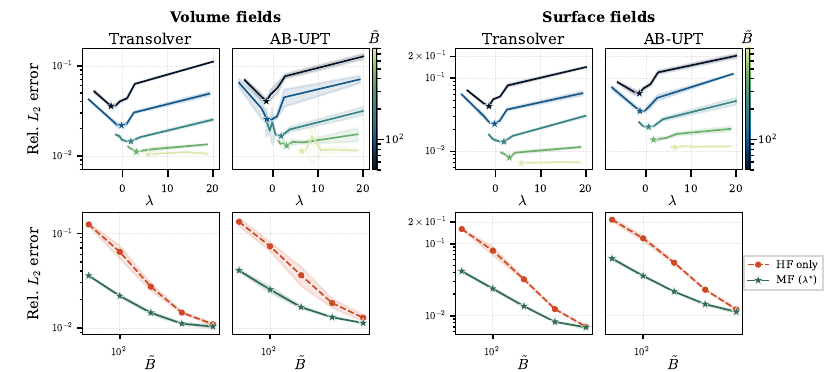}
    \caption{Airfoil experiments. \textbf{Top:} Iso-budget curves over mixture hyper-parameter $\lambda$ in terms of volume and surface test error. Minimum loss for each budget is highlighted with $\star$. \textbf{Bottom:} Compute-optimal Scaling behavior using $\lambda^*$ at each budget.}
    \label{fig:airfoil_results_overview}
\end{figure}

\begin{figure}[htbp]
    \centering
    \includegraphics[width=\linewidth]{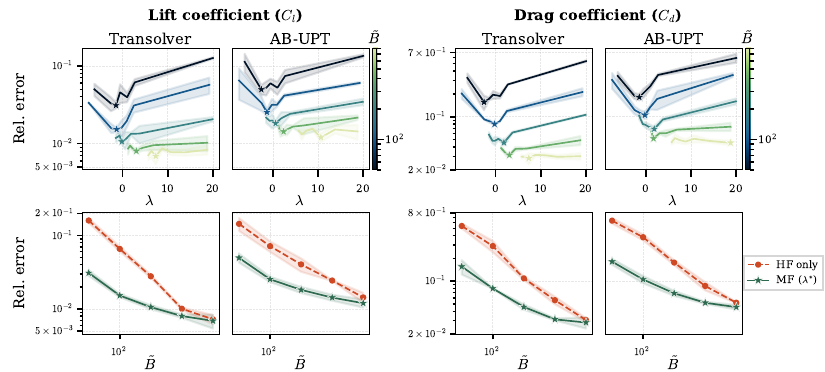}
    \caption{Airfoil experiments. \textbf{Top:} Iso-budget curves over mixture hyper-parameter $\lambda$ in terms of test error of aerodynamic coefficients. Minimum loss for each budget is highlighted with $\star$. \textbf{Bottom:} Compute-optimal Scaling behavior using $\lambda^*$ at each budget.}
    \label{fig:airfoil_coefficients_overview}
\end{figure}

\begin{table}[htbp]
\centering
\caption{Summary of the airfoil experiment results (volume field error). Multi-fidelity (MF) uses the optimal $\lambda^*$; High-fidelity (HF) is the high-fidelity-only baseline. $M_\mathrm{train}$ denotes the number of training samples; $\Delta (\%)$ denotes the relative change in number of training samples and error compared to the high-fidelity baseline.}
\label{tab:airfoil_volume}
\begin{tabular}{lrrrrrrrr}
\toprule
\multirow{2}{*}{\textbf{Arch}} & \multirow{2}{*}{$\bm{\tilde{B}}$} & \multirow{2}{*}{$\bm{\lambda^*}$} & \multicolumn{3}{c}{$\mathbf{M_{\textbf{train}}}$} & \multicolumn{3}{c}{\textbf{Rel.\ $\mathbf{L_2}$ error}} \\
\cmidrule(lr){4-6} \cmidrule(lr){7-9}
 & & & \textbf{MF} & \textbf{HF} & $\bm{\Delta}$\textbf{(\%)} & \textbf{MF} & \textbf{HF} & $\bm{\Delta}$\textbf{(\%)} \\
\midrule
\multirow{5}{*}{AB-UPT} & 50 & -1.46 & 316 & 52 & +507.7 & 0.040 & 0.133 & -69.5 \\
 & 100 & -1.33 & 634 & 101 & +527.7 & 0.025 & 0.073 & -65.2 \\
 & 200 & 1.84 & 584 & 200 & +192.0 & 0.017 & 0.036 & -54.1 \\
 & 400 & 3.05 & 872 & 397 & +119.6 & 0.013 & 0.018 & -29.3 \\
 & 800 & 6.41 & 1,236 & 798 & +54.9 & 0.011 & 0.013 & -12.0 \\
\midrule
\multirow{5}{*}{Transolver} & 50 & -2.56 & 385 & 52 & +640.4 & 0.036 & 0.125 & -71.3 \\
 & 100 & -0.23 & 503 & 101 & +398.0 & 0.022 & 0.064 & -65.9 \\
 & 200 & 1.84 & 584 & 200 & +192.0 & 0.014 & 0.027 & -47.1 \\
 & 400 & 3.05 & 872 & 397 & +119.6 & 0.011 & 0.015 & -23.6 \\
 & 800 & 5.70 & 1,301 & 798 & +63.0 & 0.010 & 0.011 & -5.6 \\
\bottomrule
\end{tabular}
\end{table}

\begin{table}[htbp]
\centering
\caption{Summary of the airfoil experiment results (surface field error). Multi-fidelity (MF) uses the optimal $\lambda^*$; High-fidelity (HF) is the high-fidelity-only baseline. $M_\mathrm{train}$ denotes the number of training samples; $\Delta (\%)$ denotes the relative change in number of training samples and error compared to the high-fidelity baseline. Optima in the saturated high-fidelity regime are reported as $\lambda^* \geq 15$ (HF-equivalent).}
\label{tab:airfoil_surface}
\begin{tabular}{lrrrrrrrr}
\toprule
\multirow{2}{*}{\textbf{Arch}} & \multirow{2}{*}{$\bm{\tilde{B}}$} & \multirow{2}{*}{$\bm{\lambda^*}$} & \multicolumn{3}{c}{$\mathbf{M_{\textbf{train}}}$} & \multicolumn{3}{c}{\textbf{Rel.\ $\mathbf{L_2}$ error}} \\
\cmidrule(lr){4-6} \cmidrule(lr){7-9}
 & & & \textbf{MF} & \textbf{HF} & $\bm{\Delta}$\textbf{(\%)} & \textbf{MF} & \textbf{HF} & $\bm{\Delta}$\textbf{(\%)} \\
\midrule
\multirow{5}{*}{AB-UPT} & 50 & -1.46 & 316 & 52 & +507.7 & 0.062 & 0.215 & -71.1 \\
 & 100 & -1.33 & 634 & 101 & +527.7 & 0.036 & 0.119 & -69.9 \\
 & 200 & 0.59 & 764 & 200 & +282.0 & 0.022 & 0.054 & -60.3 \\
 & 400 & 1.68 & 1,155 & 397 & +190.9 & 0.014 & 0.023 & -37.2 \\
 & 800 & 6.41 & 1,236 & 798 & +54.9 & 0.011 & 0.012 & -7.7 \\
\midrule
\multirow{5}{*}{Transolver} & 50 & -1.46 & 316 & 52 & +507.7 & 0.042 & 0.160 & -74.0 \\
 & 100 & -0.23 & 503 & 101 & +398.0 & 0.024 & 0.080 & -70.3 \\
 & 200 & 1.84 & 584 & 200 & +192.0 & 0.014 & 0.032 & -57.8 \\
 & 400 & 3.05 & 872 & 397 & +119.6 & 0.008 & 0.012 & -33.6 \\
 & 800 & 5.70 & 1,301 & 798 & +63.0 & 0.007 & 0.007 & -2.2 \\
\bottomrule
\end{tabular}
\end{table}

\begin{table}[htbp]
\centering
\caption{Summary of the airfoil experiment results (lift coefficient $C_l$ error). Multi-fidelity (MF) uses the optimal $\lambda^*$; High-fidelity (HF) is the high-fidelity-only baseline. $M_\mathrm{train}$ denotes the number of training samples; $\Delta (\%)$ denotes the relative change in number of training samples and error compared to the high-fidelity baseline. Optima in the saturated high-fidelity regime are reported as $\lambda^* \geq 15$ (HF-equivalent).}
\label{tab:airfoil_cl}
\begin{tabular}{lrrrrrrrr}
\toprule
\multirow{2}{*}{\textbf{Arch}} & \multirow{2}{*}{$\bm{\tilde{B}}$} & \multirow{2}{*}{$\bm{\lambda^*}$} & \multicolumn{3}{c}{$\mathbf{M_{\textbf{train}}}$} & \multicolumn{3}{c}{\textbf{Rel.\ error}} \\
\cmidrule(lr){4-6} \cmidrule(lr){7-9}
 & & & \textbf{MF} & \textbf{HF} & $\bm{\Delta}$\textbf{(\%)} & \textbf{MF} & \textbf{HF} & $\bm{\Delta}$\textbf{(\%)} \\
\midrule
\multirow{5}{*}{AB-UPT} & 50 & -2.56 & 385 & 52 & +640.4 & 0.050 & 0.144 & -65.6 \\
 & 100 & -1.33 & 634 & 101 & +527.7 & 0.025 & 0.072 & -64.6 \\
 & 200 & 0.59 & 764 & 200 & +282.0 & 0.018 & 0.040 & -54.9 \\
 & 400 & 2.36 & 1,013 & 397 & +155.2 & 0.014 & 0.024 & -41.4 \\
 & 800 & 10.62 & 1,024 & 798 & +28.3 & 0.012 & 0.014 & -16.6 \\
\midrule
\multirow{5}{*}{Transolver} & 50 & -1.46 & 316 & 52 & +507.7 & 0.031 & 0.160 & -80.7 \\
 & 100 & -1.33 & 634 & 101 & +527.7 & 0.015 & 0.066 & -76.8 \\
 & 200 & -0.16 & 940 & 200 & +370.0 & 0.011 & 0.028 & -62.0 \\
 & 400 & 3.05 & 872 & 397 & +119.6 & 0.008 & 0.010 & -20.4 \\
 & 800 & 7.34 & 1,159 & 798 & +45.2 & 0.007 & 0.007 & -4.8 \\
\bottomrule
\end{tabular}
\end{table}

\begin{table}[H]
\centering
\caption{Summary of the airfoil experiment results (drag coefficient $C_d$ error). Multi-fidelity (MF) uses the optimal $\lambda^*$; High-fidelity (HF) is the high-fidelity-only baseline. $M_\mathrm{train}$ denotes the number of training samples; $\Delta (\%)$ denotes the relative change in number of training samples and error compared to the high-fidelity baseline. Optima in the saturated high-fidelity regime are reported as $\lambda^* \geq 15$ (HF-equivalent).}
\label{tab:airfoil_cd}
\begin{tabular}{lrrrrrrrr}
\toprule
\multirow{2}{*}{\textbf{Arch}} & \multirow{2}{*}{$\bm{\tilde{B}}$} & \multirow{2}{*}{$\bm{\lambda^*}$} & \multicolumn{3}{c}{$\mathbf{M_{\textbf{train}}}$} & \multicolumn{3}{c}{\textbf{Rel.\ error}} \\
\cmidrule(lr){4-6} \cmidrule(lr){7-9}
 & & & \textbf{MF} & \textbf{HF} & $\bm{\Delta}$\textbf{(\%)} & \textbf{MF} & \textbf{HF} & $\bm{\Delta}$\textbf{(\%)} \\
\midrule
\multirow{5}{*}{AB-UPT} & 50 & -1.46 & 316 & 52 & +507.7 & 0.181 & 0.629 & -71.2 \\
 & 100 & -0.23 & 503 & 101 & +398.0 & 0.105 & 0.380 & -72.3 \\
 & 200 & 1.84 & 584 & 200 & +192.0 & 0.069 & 0.176 & -60.8 \\
 & 400 & 1.68 & 1,155 & 397 & +190.9 & 0.052 & 0.086 & -40.0 \\
 & 800 & $\geq 15$ & 880 & 798 & +10.3 & 0.045 & 0.052 & -12.6 \\
\midrule
\multirow{5}{*}{Transolver} & 50 & -2.56 & 385 & 52 & +640.4 & 0.156 & 0.536 & -70.9 \\
 & 100 & -0.23 & 503 & 101 & +398.0 & 0.080 & 0.291 & -72.5 \\
 & 200 & 1.84 & 584 & 200 & +192.0 & 0.046 & 0.109 & -58.2 \\
 & 400 & 3.05 & 872 & 397 & +119.6 & 0.031 & 0.056 & -44.2 \\
 & 800 & 7.34 & 1,159 & 798 & +45.2 & 0.028 & 0.031 & -7.4 \\
\bottomrule
\end{tabular}
\end{table}


\subsection{Bootstrapping approach}
\label{app:bootstrapping_details}
The savings ratio $S(\mathcal{E}^\star) = \tilde{B}_{\mathrm{HF}} / \tilde{B}_{\mathrm{MF}}$ in \cref{fig:savings_summary} compares the budgets at which high-fidelity-only training and multi-fidelity training at the optimal mixture reach the specified error threshold $\mathcal{E}^\star$.
To compute it, we average the test error of each $(\tilde{B}, \lambda)$ cell over its training seeds, take the minimum over $\lambda$ at each budget to obtain the respective multi-fidelity curve, and average the high-fidelity runs per budget for the baseline.
Both curves are interpolated linearly in log-log space between the two budgets surrounding $\mathcal{E}^\star$ to find the crossing points.

The confidence intervals come from a seed bootstrap.
In each of $2{,}000$ draws we resample the seeds of every cell with replacement, recompute the optimal multi-fidelity and high-fidelity curves, and recompute the ratio as described above.
Bar heights are the medians of the resulting distribution and whiskers span the 2.5th to 97.5th percentiles.
The intervals reflect training randomness only, since the sampled dataset of each cell is fixed.

\section{Extrapolating lambda values}
\label{app:lambda_heuristic}
This section outlines the detailed procedure used for fitting the coefficients from \cref{eq:lambda_star_functional_form} in \cref{app:fitting_details}, calculations on the associated overhead cost in \cref{app:lambda_fitting_overhead_cost}, as well as detailed results across all architectures and datasets in \cref{app:lambda_heuristic_detailed_results}.

\subsection{Fitting the functional form}
\label{app:fitting_details}

Following Approach 2 proposed by \citet{hoffman2022chinchilla}, we sweep $\lambda$ for a fixed set of budgets $\tilde{B}$.
We then fit a parabola around each iso budget's local minimum and record the optimal lambda value, which is in turn used to fit the variables $a$ and $b$ of \cref{eq:lambda_star_functional_form}.

Our approach aims to use a few low calibration budgets and extrapolate to optimal data mixtures at higher budgets.
Below, we provide a detailed recipe for the fitting procedure with safeguards important to achieve robust extrapolation:

\begin{enumerate}
    \item \textbf{Number and spacing of calibration budgets:} We use at least three calibration budgets, which are $\geq 2 \times$ apart from each other. Since a two-point linear fit is fully determined, we found it too sensitive to potential outliers in the calibration budget range and recommend three calibration points.
    
    \item \textbf{Local parabola fit:} To determine a smooth optimal $\lambda$ per calibration budget, we take the $\lambda$ at minimum error of the respective sweep together with its two neighboring values, and fit a 3 point parabola.
    Having trained three or more seeds, we bootstrap this process, compute the vertex per draw (300 draws with resampling of seeds within each $\lambda$ cell) and store the mean vertices and their standard deviation.

    \item \textbf{Avoidance of pre-asymptotic regime:} An optimum pinned at the low-fidelity extreme of the sweep (smallest $\lambda$), or even drifting toward lower optimal $\lambda$ with a larger budget, signals a pre-asymptotic regime.
    An example of this can be seen in \cref{fig:poisson_discretization_results_detailed} and \cref{fig:poisson_truncation_results_detailed} for the 1D U-Net, where the three smallest budgets show pre-asymptotic behavior.
    In this case, we slide the calibration budget window to higher budgets, essentially doubling all three calibration budgets.
    Repeat this if necessary.

    \item \textbf{Fitting the functional form:} To obtain a robust fit of the constants $a$ and $b$ from \cref{eq:lambda_star_functional_form}, use a \ac{WLS} procedure using the bootstrapped mean and $w = 1 / \max(\mathrm{std}, 0.25)^{2}$.

    \item \textbf{Saturation censoring:} Since $\lambda$ enters the sampling density exponentially (\cref{eq:pmf}), beyond a menu dependent threshold, a ``mixture'' focuses sampling almost exclusively on high-fidelity samples.
    We define $\lambda_{\text{sat}}=15$.
    This bound is used twice: (i) if the extrapolated prediction is larger than $\lambda_{\text{sat}}$, allocate the entire generation budget to high-fidelity samples, and (ii) if an empirical minimum of a calibration budget already lies above $\lambda_{\text{sat}}$, compute at larger budgets should be allocated to high-fidelity samples only.
\end{enumerate}

\subsection{Overhead cost}
\label{app:lambda_fitting_overhead_cost}
The calibration sweep incurs some overhead data generation cost.
Below, we quantify this overhead.

Na\"{i}vely, the calibration sweep of our setup with 7 mixtures at each of the 3 calibration budgets costs $7 \cdot (B_1+B_2+B_3)$.
Assuming an always doubling data cost across budgets ($B_1,B_2,B_3,B_4,...$), the overhead of the first deployment budget $B_4=2B_3$ would be $\sim613\%$.
However, this na\"{i}ve approach leads to a substantial overestimation, since samples can be reused: assuming all plans draw inputs from the same distribution, a sample generated at fidelity level $\xi$ can be part of any dataset containing that level.
The sweep therefore only needs, for each fidelity level, the maximum number of samples over all calibration plans, not their sum.
Since the calibration budgets grow geometrically, in most cases this maximum is set by the largest budget, so the smaller budgets add little.
By the same argument, the calibration samples can be folded into the deployment dataset, so only samples the deployment cannot absorb count as overhead.

As an example of this, on the linear-elasticity solver problem, this overhead amounts to $\sim51\%$ of the deployment budget when deploying once at $2B_3$, shrinking to $\sim9\%$ at our largest tested budget of $16B_3$. For comparison, sweeping $\lambda$ directly at the deployment budget, with the same sample reuse, costs an additional $170\%$.

\subsection{Detailed results}
\label{app:lambda_heuristic_detailed_results}

\cref{tab:regret_solver} displays the results across all solver error setups, split by dataset and architecture.
Since training at arbitrary $\lambda$ is infeasible, predictions are evaluated at the nearest swept grid cell.
Reported regrets therefore include a quantization component that depends on the $\lambda$ grid spacing and local smoothness of the error landscape.
The table shows that our functional fit extrapolates reliably.
Regrets stay below $14\%$ in every cell and do not grow with extrapolation depth, tested at budgets up to $16\times$ the calibration ceiling.

Results on the discretization axis (\cref{tab:regret_discretization}) exhibit two distinct behaviors.
First, for the linear elasticity and hyperelasticity AB-UPT settings the calibration sweeps already place their minima at the high-fidelity boundary of the grid, above $\lambda_{\text{sat}}$.
The saturation censoring (step 5. above) is therefore triggered and HF-only extrapolation is adopted.
Second, all Poisson discretization settings undergo an abrupt transition of optimal $\lambda$ value above the calibration window (\cref{tab:poisson_discretization}).
The empirical minimum drifts slowly within the calibration range and then abruptly jumps to the high-fidelity regions of the sweeps.
Our log-linear Ansatz cannot capture this behavior and hence the largest regrets concentrate at the transition budgets (e.g.\ $48.9\%$ for Poisson-1D FNO, $70.3\%/99.2\%$ for Poisson-2D U-Net).

Notably, the recipe partially recovers beyond the transition.
Once the extrapolated prediction crosses $\lambda_{\text{sat}}$, the saturation clause causes HF-only deployment, which is optimal in the post transition regime (as indicated by zero regret at the deepest holdout budgets of Poisson 1D).

\begin{table}[htbp]
\centering
\caption{Per setting recipe results on the \textbf{solver error} axis: regret at each holdout budget (ascending).}
\label{tab:regret_solver}
\begin{tabular}{llcrrrr}
\toprule
\multirow{2}{*}{\textbf{Dataset}} & \multirow{2}{*}{\textbf{Arch}} & \multirow{2}{*}{\textbf{Calibration} $\bm{\tilde{B}}$} & \multicolumn{4}{c}{\textbf{Regret at holdout} $\bm{\tilde{B}}$ \textbf{(\%)}} \\
\cmidrule(lr){4-7}
 & & & \textbf{1\textsuperscript{st}} & \textbf{2\textsuperscript{nd}} & \textbf{3\textsuperscript{rd}} & \textbf{4\textsuperscript{th}} \\
\midrule
\multirow{2}{*}{Poisson 1D} & FNO & 10, 30, 100 & 13.6 & 10.7 & 8.1 & 5.9 \\
 & U-Net & 100, 233, 520 & 8.7 & 4.1 & 2.4 & 3.3 \\
\midrule
\multirow{2}{*}{Poisson 2D} & FNO & 100, 300, 1000 & 0.1 & 0.3 & -- & -- \\
 & U-Net & 100, 300, 1000 & 12.9 & 3.1 & -- & -- \\
\midrule
\multirow{1}{*}{Poisson 3D} & U-Net & 100, 300, 1000 & 0.0 & 3.1 & -- & -- \\
\midrule
\multirow{2}{*}{Linear elasticity} & AB-UPT & 100, 200, 400 & 0.0 & 0.6 & 5.2 & 0.0 \\
 & Transolver & 100, 200, 400 & 8.0 & 9.9 & 3.6 & 0.9 \\
\midrule
\multirow{2}{*}{Hyperelasticity} & AB-UPT & 100, 200, 400 & 4.7 & 1.7 & 0.0 & 0.3 \\
 & Transolver & 200, 400, 800 & 0.0 & 5.1 & 7.8 & -- \\
\bottomrule
\end{tabular}
\end{table}

\begin{table}[htbp]
\centering
\caption{Per-setting recipe results on the \textbf{discretization-error} axis: regret at each held-out budget (ladder rungs above the calibration window, in ascending order).}
\label{tab:regret_discretization}
\begin{tabular}{llcrrrr}
\toprule
\multirow{2}{*}{\textbf{Dataset}} & \multirow{2}{*}{\textbf{Arch}} & \multirow{2}{*}{\textbf{Calibration} $\bm{\tilde{B}}$} & \multicolumn{4}{c}{\textbf{Regret at held-out} $\bm{\tilde{B}}$ \textbf{(\%)}} \\
\cmidrule(lr){4-7}
 & & & \textbf{1\textsuperscript{st}} & \textbf{2\textsuperscript{nd}} & \textbf{3\textsuperscript{rd}} & \textbf{4\textsuperscript{th}} \\
\midrule
\multirow{2}{*}{Poisson 1D} & FNO & 10, 30, 100 & 48.9 & 9.2 & 0.0 & 0.0 \\
 & U-Net & 30, 100, 300 & 27.7 & 28.9 & 0.0 & -- \\
\midrule
\multirow{2}{*}{Poisson 2D} & FNO & 100, 300, 1000 & 1.2 & 9.1 & -- & -- \\
 & U-Net & 100, 300, 1000 & 70.3 & 99.2 & -- & -- \\
\midrule
\multirow{1}{*}{Poisson 3D} & U-Net & 10, 100, 300 & 0.0 & 22.8 & 55.6 & -- \\
\midrule
\multirow{2}{*}{Linear elasticity} & AB-UPT & 200, 400, 800 & 0.0 & 0.0 & -- & -- \\
 & Transolver & 100, 200, 400 & 12.6 & 9.0 & 4.0 & -- \\
\midrule
\multirow{2}{*}{Hyperelasticity} & AB-UPT & 100, 200, 400 & 7.7 & 0.0 & 0.0 & -- \\
 & Transolver & 100, 200, 400 & 10.3 & 20.0 & 8.6 & -- \\
\midrule
\multirow{2}{*}{Airfoil} & AB-UPT & 50, 100, 200 & 0.0 & 3.4 & -- & -- \\
 & Transolver & 50, 100, 200 & 0.0 & 0.0 & -- & -- \\
\bottomrule
\end{tabular}
\end{table}

\section{Ablations}
\label{app:ablations}

\subsection{Binary vs. spectrum formulation of fidelity}
\label{app:binary_vs_spectrum}

\cref{sec:spectrum_vs_binary} shows that the spectrum formulation  outperforms the binary baseline across all mixtures on the hyperelasticity problem with discretization error, where mesh sizes span two orders of magnitude (\cref{app:hyperelasticity}).
To test the limits of this finding, we repeat the experiment but hold the mesh fixed and instead vary the number of solver iterations.
The fidelities are therefore much closer in the sense that they all share the same discretization.

\begin{figure}[htbp]
    \centering
    \begin{subfigure}[b]{0.48\textwidth}
        \centering
        \includegraphics[width=\linewidth]{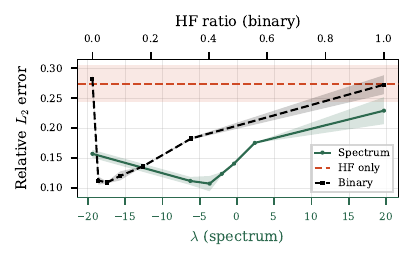}
        \caption{$\tilde{B}=200.$}
        \label{fig:mf_vs_bf_solver_b200}
    \end{subfigure}
    \hfill
    \begin{subfigure}[b]{0.48\textwidth}
        \centering
        \includegraphics[width=\linewidth]{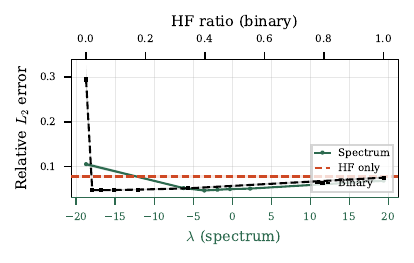}
        \caption{$\tilde{B}=1{,}600$.}
        \label{fig:mf_vs_bf_solver_b1600}
    \end{subfigure}
    \caption{Spectrum vs.\ binary fidelity formulation at fixed data generation budgets $\tilde{B}$ for a Transolver model on the the hyperelasticity dataset (solver error).}
    \label{fig:mf_vs_bf_comparison_solver}
\end{figure}

\cref{fig:mf_vs_bf_comparison_solver} shows the result at two data generation budgets ($\tilde{B}=200$ and $\tilde{B}=1{,}600$).
Here, the picture is less clear-cut than in \cref{fig:mf_vs_bf_comparison}.
At both budgets, neither curve dominates the other across the entire mixing range.
The spectrum formulation (green line) never peaks as high as the binary formulation (black dashed line), making it more robust to the choice of mixing ratio.
At the optima, however, the binary formulation achieves marginally lower errors.
We attribute this trade-off to the narrower fidelity range in this setting: because all samples share the same discretization, the two endpoints of the binary formulation are much closer.

This raises a natural question: \emph{what is the right family of mixing distributions for multi-fidelity training?}
We suspect that no single distribution dominates across all settings.
Our framework allows for arbitrary parametrizable distributions over the discrete fidelity menu $\Xi$, including bimodal distributions that recover the binary formulation as an edge case.
However, adding parameters to the distribution enlarges the search space of mixing hyperparameters.
We therefore opted for the expressive but simple softmax parametrization and leave a systematic study of alternative mixture distributions for future work (\cref{sec:conclusion}).

\subsection{On interpolation smoothness}
\label{app:interpolation_smoothness}
We found that the surrogates' error smoothness at interpolation fidelities (unseen during training but in support $\hat{\xi}$) is mainly dependent on two factors, namely the number of samples in adjacent fidelity regions as well as the conditioning resolution of the sinusoidal fidelity encoding.
The ablation below demonstrates these findings for a Transolver model trained on the hyperelasticity dataset using different training configurations.

\begin{figure}[htbp]
    \centering
    \begin{subfigure}[b]{0.32\textwidth}
        \centering
        \includegraphics[width=\linewidth]{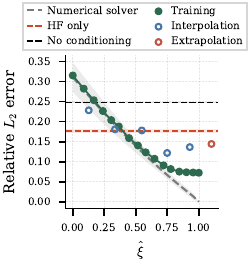}
        \caption{$\tilde{B}=400$, $\lambda^{\star} = -6.25$.}
        \label{fig:conditioning_hyperelasticity_solver_b400}
    \end{subfigure}
    \hfill
    \begin{subfigure}[b]{0.32\textwidth}
        \centering
        \includegraphics[width=\linewidth]{figs/hyperelasticity/conditioning_hyperelasticity_solver_b1600_main.pdf}
        \caption{$\tilde{B}=1{,}600$, $\lambda^{\star} = -3.59$.}
        \label{fig:conditioning_hyperelasticity_solver_b1600}
    \end{subfigure}
    \hfill
    \begin{subfigure}[b]{0.32\textwidth}
        \centering
        \includegraphics[width=\linewidth]{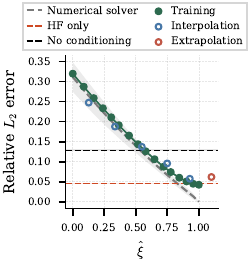}
        \caption{$\tilde{B}=6{,}400$, $\lambda^{\star}= -0.31$.}
        \label{fig:conditioning_hyperelasticity_solver_b6400}
    \end{subfigure}
    \caption{Influence of training budget and fidelity mixture on interpolation smoothness. Showing Transolver models trained on the optimal fidelity mixtures $\lambda^{\star}$ at different budgets $\tilde{B}$ on the hyperelasticity dataset (solver error axis). (b) repeats \cref{fig:conditioning_plot} for comparison.}
    \label{fig:conditioning_sample_dependence}
\end{figure}

\cref{fig:conditioning_sample_dependence} visualizes the error smoothness at interpolation fidelities for models trained with different budgets $\tilde{B}$ and optimal fidelity mixtures $\lambda^{\star}$.
Models trained at $\tilde{B}=400$ show prediction errors at interpolation fidelities that deviate significantly from the respective errors of the numerical solver, especially for higher fidelity levels $\hat{\xi}>0.5$ (\cref{fig:conditioning_hyperelasticity_solver_b400}).
This can be attributed to the fact that the overall sample count at this budget is low and the optimal fidelity mix is skewed to lower fidelities.
\cref{fig:conditioning_hyperelasticity_solver_b1600} already displays a smoother interpolation for models trained at $\tilde{B}=1600$, while still exhibiting some deviations at higher fidelities.
In contrast, \cref{fig:conditioning_hyperelasticity_solver_b6400} clearly shows the benefits of a balanced mix ($\lambda^{\star} \approx 0$, uniform distribution over fidelities) with extensive training samples at $\tilde{B}=6400$, leading to a smooth interpolation throughout the entire fidelity support.

\begin{figure}[htbp]
    \centering
    \begin{subfigure}[b]{0.32\textwidth}
        \centering
        \includegraphics[width=\linewidth]{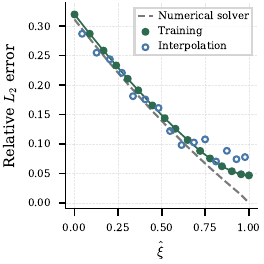}
        \caption{$\ell_{\min} = 2\pi/1000$.}
        \label{fig:condscale_1000_ab_b1600}
    \end{subfigure}
    \hfill
    \begin{subfigure}[b]{0.32\textwidth}
        \centering
        \includegraphics[width=\linewidth]{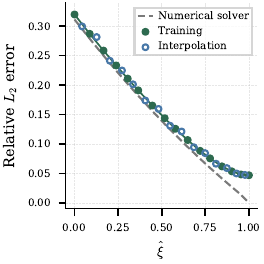}
        \caption{$\ell_{\min} = 2\pi/50$.}
        \label{fig:condscale_50_ab_b1600}
    \end{subfigure}
    \hfill
    \begin{subfigure}[b]{0.32\textwidth}
        \centering
        \includegraphics[width=\linewidth]{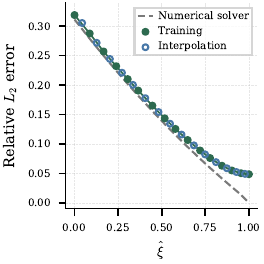}
        \caption{$\ell_{\min} = 2\pi$.}
        \label{fig:condscale_1_ab_b1600}
    \end{subfigure}
    \caption{Influence of conditioning resolution on interpolation smoothness. Showing Transolver models trained with $\tilde{B} = 1{,}600$ and $\lambda^{\star} = -3.59$ across different encoding resolutions $\ell_{\min}$.}
    \label{fig:conditioning_resolution_dependence}
\end{figure}

\cref{fig:conditioning_resolution_dependence} shows the error smoothness with respect to the sinusoidal encoding resolution of the conditioning signal.
We ablate $\ell_{\min}\in\{2\pi,\,2\pi/50,\,2\pi/1000\}$ (\cref{eq:sinusoidal_encoding}) at a budget of $\tilde{B}=1600$ and the corresponding optimal fidelity mixture.
The resulting errors at interpolation fidelities become increasingly smooth as the encoding resolution decreases (larger $\ell_{\min}$).

The finest resolution $\ell_{\min}=2\pi/1000$ achieves the lowest test error at $\hat\xi=1$ with an increasing performance difference at lower budgets.
Since the primary goal of our multi-fidelity training is to approximate the highest-fidelity simulator, our main experiments use this setting, trading interpolation smoothness for test performance.
If smooth interpolation across fidelities is desired, a resolution calibrated to the spacing of the training fidelities ($\ell_{\min}=2\pi/50$ here) retains it at a  small increase in test error at moderate budgets.

\subsection{On interpolation errors for the grid experiments}
\label{sec:interpolation-error}

In the Poisson experiments, as described in \cref{app:poisson}, the surrogates operate on one fixed resolution per dimension of $N^*=64$.
This design choice was taken in order to benchmark a standard U-Net architecture, which is by default not resolution invariant.
Additionally, we saw that models not being fully resolution invariant likely is the issue of reduced MF advantages in mesh experiments.
However, to have samples at a consistent resolution carry \emph{discretization error}, we had to interpolate both input $\mathbf{f}_h$ and output $\mathbf{u}_h$ to the operating resolution.
For this, we employed a multi-linear interpolation scheme.
This incurs a non-negligible error, which we analyze with this ablation.

Multi-linear interpolation is second-order accurate, the same order as the discretization itself (\cref{app:poisson}).
It therefore does not change the asymptotic convergence
of the total error, which retains the expected second-order rate
(as visible in \cref{fig:poisson-discretization}). It can, however, change the error constant,
and with it the value of low-fidelity samples for training.

To quantify this effect, we reran the 1D U-Net discretization experiment using error-free interpolation with the discrete sine transform. This applies here because the states are bandlimited. In \cref{tab:interpolation_ablation} we report errors using both approaches and the corresponding error improvements of using the optimal mixture $\lambda ^*$ versus a high-fidelity only run.

\begin{table}[htbp]
\centering
\caption{Interpolation ablation on the 1D U-Net Poisson experiment (discretization error). We compare the multi-linear prolongation of the main experiments with prolongation via the discrete sine transform (DST). Both branches use the same mixtures $\lambda \in \{-20, -5, -1, 0, 1, 5, 20\}$ and 6 seeds, so the multi-linear values can differ slightly from Table~\ref{tab:poisson_discretization}, which uses a finer $\lambda$ grid. Multi-fidelity (MF) uses the optimal $\lambda^*$. High-fidelity (HF) is the high-fidelity-only baseline. $M_\mathrm{train}$ denotes the number of training samples. $\Delta (\%)$ denotes the relative change in number of training samples and error compared to the high-fidelity baseline. Optima in the saturated high-fidelity regime are reported as $\lambda^* \geq 15$ (HF-equivalent).}
\label{tab:interpolation_ablation}
\begin{tabular}{lrrrrrrrr}
\toprule
\multirow{2}{*}{\textbf{Prolongation}} & \multirow{2}{*}{$\bm{\tilde{B}}$} & \multirow{2}{*}{$\bm{\lambda^*}$} & \multicolumn{3}{c}{$\mathbf{M_{\textbf{train}}}$} & \multicolumn{3}{c}{\textbf{Rel.\ $\mathbf{L_2}$ error}} \\
\cmidrule(lr){4-6} \cmidrule(lr){7-9}
 & & & \textbf{MF} & \textbf{HF} & $\bm{\Delta}$\textbf{(\%)} & \textbf{MF} & \textbf{HF} & $\bm{\Delta}$\textbf{(\%)} \\
\midrule
\multirow{6}{*}{Multi-linear} & 10 & 5.00 & 11 & 10 & +10.0 & 1.422 & 1.454 & -2.2 \\
 & 100 & -20.00 & 240 & 100 & +140.0 & 0.273 & 0.685 & -60.1 \\
 & 300 & -1.00 & 488 & 300 & +62.7 & 0.114 & 0.204 & -44.3 \\
 & 1,000 & $\geq 15$ & 1,000 & 1,000 & +0.0 & 0.034 & 0.034 & +0.0 \\
 & 3,000 & $\geq 15$ & 3,000 & 3,000 & +0.0 & 0.018 & 0.018 & +0.0 \\
 & 10,000 & $\geq 15$ & 10,001 & 10,001 & +0.0 & 0.016 & 0.016 & +0.0 \\
\midrule
\multirow{6}{*}{DST} & 10 & $\geq 15$ & 10 & 10 & +0.0 & 1.378 & 1.378 & +0.0 \\
 & 100 & -20.00 & 240 & 100 & +140.0 & 0.255 & 0.754 & -66.1 \\
 & 300 & -5.00 & 644 & 300 & +114.7 & 0.070 & 0.200 & -64.9 \\
 & 1,000 & -1.00 & 1,623 & 1,000 & +62.3 & 0.027 & 0.036 & -23.9 \\
 & 3,000 & $\geq 15$ & 3,000 & 3,000 & +0.0 & 0.018 & 0.018 & +0.0 \\
 & 10,000 & $\geq 15$ & 10,001 & 10,001 & +0.0 & 0.016 & 0.016 & +0.0 \\
\bottomrule
\end{tabular}
\end{table}

In regimes of low budget ($10 < \tilde{B} \leq 1{,}000$), optimal MF-errors are lower on the error-free interpolation branch. This seems aligned with different $\lambda^*$ being found at many budgets.
The HF-only error does not significantly change between the two branches.
Correspondingly, the MF advantage is higher under DST interpolation.

Overall, removing the interpolation error increases rather than decreases the MF gain.
The multi-linear interpolation used in the main experiments therefore biases the results against MF training, and the MF gains we report for the grid experiments can be seen as conservative.

\subsection{Model size ablation}
\label{app:model-size-ablation}

We conduct a small ablation to investigate the effect of model parameter size on multi-fidelity gains. In \cref{tab:model-size-ablation}, we show variations to the U-Net architecture for the 2D Poisson task over the solver error axis.

\begin{table}[htbp]
\centering
\caption{Varying the parameter size of the surrogate architecture via the base channels of the U-Net (2D Poisson, solver error). NN floor is the HF-only error at the highest budget. $\tilde{B}_{10\%}$ is the budget at which the MF ($\lambda^*$) and the HF-only scaling curves reach a relative $L_2$ error of 10\%, and the saving is their ratio, as in Fig.~\ref{fig:savings_summary} (95\% seed-bootstrap confidence interval in brackets). All widths use the mixtures $\lambda \in \{-20, -5, -1, 0, 1, 5, 20\}$ and 6 seeds. With the finer $\lambda$ grid of the main experiment, the saving of the 64-channel model is 3.7$\times$.}
\label{tab:model-size-ablation}
\begin{tabular}{rrcccc}
\toprule
\textbf{Base ch.} & \textbf{Params} & \textbf{NN floor} & $\bm{\tilde{B}_{10\%}}$\textbf{, MF}$\bm{(\lambda^*)}$ & $\bm{\tilde{B}_{10\%}}$\textbf{, HF-only} & \textbf{Saving @ 10\%} \\
\midrule
16 & 0.69M & 4.0\% & 260 & 715 & 2.8$\times$ [2.5, 3.0] \\
32 & 2.74M & 2.5\% & 236 & 745 & 3.2$\times$ [2.8, 3.5] \\
64 (main) & 10.92M & 1.8\% & 224 & 758 & 3.4$\times$ [3.2, 3.6] \\
128 & 43.65M & 1.4\% & 207 & 654 & 3.2$\times$ [2.7, 3.5] \\
\bottomrule
\end{tabular}
\end{table}

The error floor drops monotonically from 
4\%
to 1.4\% as capacity grows.
While the larger models seem to reach the 10\% error threshold training on multiple fidelities with a smaller budget (likely due to 10\% being further from their error floor), the achieved multi-fidelity savings remain at about 3$\times$.
Note that training duration was kept constant in update steps (wall-clock and training FLOPs grow with model size).
We also refer to \cref{app:train-duration-ablation} for an ablation of the number of training steps.

\subsection{Training duration ablation}
\label{app:train-duration-ablation}

For comparability across $\lambda$ sweeps and thus different dataset sizes, we hold the number of update steps within one study fixed.
To ablate whether multi-fidelity gains only occur due to under- or over-training, we repeat the experiments from the 1D U-Net Poisson study over the solver error axis at budget $\tilde{B} = 300$, varying the update steps (the learning rate scheduler was modified accordingly).
In \cref{tab:training-duration-ablation}, we present the achieved test errors of the optimal MF network, the corresponding HF-only network, and the relative change.

\begin{table}[htbp]
\centering
\caption{Training duration ablation for the 1D U-Net Poisson experiment (solver error) at $\tilde{B} = 300$. All step counts use the mixtures $\lambda \in \{-20, -5, -1, 0, 1, 5, 20\}$ and 6 seeds, so the 20,000-step row can differ from Table~\ref{tab:poisson_truncation}, which uses a finer $\lambda$ grid. Multi-fidelity (MF) uses the optimal $\lambda^*$. High-fidelity (HF) is the high-fidelity-only baseline. $M_\mathrm{train}$ denotes the number of training samples. $\Delta (\%)$ denotes the relative change in number of training samples and error compared to the high-fidelity baseline.}
\label{tab:training-duration-ablation}
\begin{tabular}{rrrrrrrr}
\toprule
\multirow{2}{*}{\textbf{Steps}} & \multirow{2}{*}{$\bm{\lambda^*}$} & \multicolumn{3}{c}{$\mathbf{M_{\textbf{train}}}$} & \multicolumn{3}{c}{\textbf{Rel.\ $\mathbf{L_2}$ error}} \\
\cmidrule(lr){3-5} \cmidrule(lr){6-8}
 & & \textbf{MF} & \textbf{HF} & $\bm{\Delta}$\textbf{(\%)} & \textbf{MF} & \textbf{HF} & $\bm{\Delta}$\textbf{(\%)} \\
\midrule
1,000 & -1.00 & 534 & 300 & +78.0 & 0.181 & 0.246 & -26.6 \\
2,500 & -1.00 & 534 & 300 & +78.0 & 0.153 & 0.239 & -36.0 \\
5,000 & -1.00 & 534 & 300 & +78.0 & 0.139 & 0.227 & -38.7 \\
10,000 & -5.00 & 826 & 300 & +175.3 & 0.124 & 0.222 & -44.2 \\
20,000 (main) & -5.00 & 826 & 300 & +175.3 & 0.094 & 0.214 & -56.2 \\
40,000 & -5.00 & 826 & 300 & +175.3 & 0.069 & 0.212 & -67.2 \\
80,000 & -5.00 & 826 & 300 & +175.3 & 0.059 & 0.227 & -74.1 \\
\bottomrule
\end{tabular}
\end{table}

Under longer training durations, the MF network consistently improves on the test error, reducing the error from 18.1\% at 1,000 steps to 5.9\% at 80,000 (a reduction of 3.07$\times$).
In contrast, the HF-only surrogate reduces the error by only 1.09$\times$,
indicating that the corresponding HF-only dataset is too small.

Given these findings and insights from neural scaling laws~\citep{hoffman2022chinchilla}, this would hint that compute-optimal (in the sense of training compute and not our data generation compute) MF models should be smaller but train for more epochs.
This, together with the fact that $\lambda^\star$ decreases with increasing training steps, underscores the importance of a full scaling study across all four axes that we identified as future work.
Nevertheless, the savings we report in our main study at 20,000 steps are therefore likely underestimates of the potential of the method.

\subsection{Ablating fidelity menu endpoints}
\label{app:endpoint-ablation}

In \cref{app:fidelity_support}, we presented the heuristic we use to select $\xi_{\text{min}}$ and $\xi_{\text{max}}$.
To ablate this, we run the 2D solver-error axis U-Net experiment with a modified high-fidelity endpoint $\xi_{\text{max}}$.
Since the numerical fidelity is induced by the number of V-cycles $K$ (\cref{app:mg}), we varied $K_{\text{max}} \in \{3, 5, 6, 7, 9\}$ (7 was used in the main experiments) across the full $(\tilde{B},\lambda)$ sweep.
To make menus with different tops comparable, we
\begin{enumerate*}[label=(\roman*)]
    \item evaluate every menu against the same menu-independent reference (the DST-exact discrete solution, the $K \to \infty$ limit (\cref{app:test-fidelity-ablation})), and
    \item normalize all budgets by a single fixed cost anchor ($K=7$), so equal budget means equal total solver compute.
\end{enumerate*}
\cref{tab:endpoint-ablation} displays errors at $\lambda^*$ and the relative change $\Delta(\%)$ compared to an HF-only run within the menu of the respective $K_{\max}$.

\begin{table}[htbp]
\centering
\caption{Endpoint ablation for the 2D Poisson U-Net experiment (solver error). Each entry reports the mean relative $L_2$ error (\%) at $\lambda^*$ and, in parentheses, the relative change $\Delta (\%)$ in error compared to HF-only training within the same menu. Budgets $\tilde{B}_c$ are normalized by a fixed cost anchor ($K=7$), and all menus are evaluated against the same DST-exact reference. All menus use the mixtures $\lambda \in \{-20, -5, -1, 0, 1, 5, 20\}$ and 6 seeds, so the $K_{\max}=7$ column differs from \cref{tab:poisson_truncation}, which uses a finer $\lambda$ grid and the menu top as reference.}
\label{tab:endpoint-ablation}
\begin{tabular}{rccccc}
\toprule
$\bm{\tilde{B}_c}$ & $\bm{K_{\max}=3}$ & $\bm{K_{\max}=5}$ & $\bm{K_{\max}=6}$ & $\bm{K_{\max}=7}$ \textbf{(main)} & $\bm{K_{\max}=9}$ \\
\midrule
100 & 14.94 (-62.4) & 15.97 (-68.7) & 18.00 (-68.3) & 20.64 (-62.5) & 18.98 (-72.5) \\
300 & 7.78 (-22.8) & 6.90 (-53.7) & 7.36 (-64.2) & 7.75 (-70.9) & 8.65 (-78.2) \\
1,000 & 4.67 (-15.0) & 3.21 (-46.3) & 3.11 (-53.8) & 3.19 (-57.5) & 3.43 (-60.1) \\
3,000 & 4.14 (-0.4) & 2.12 (-14.6) & 2.07 (-25.7) & 2.12 (-32.5) & 2.11 (-45.3) \\
10,000 & 4.06 (-0.8) & 1.84 (+0.0) & 1.76 (-1.6) & 1.77 (+0.0) & 1.77 (-3.9) \\
\bottomrule
\end{tabular}
\end{table}

First, our choice is near-optimal and insensitive to over-specification: The compute-optimal error saturates for $K_{\max} \ge 6$ at $\approx1.8$\%, matching the network's capacity floor (\cref{app:model-size-ablation}), and moving the high-fidelity endpoint from $K=7$ to $K=9$ changes it by $<$1\% relative.
Our heuristic ($K=7$, placed just below the network floor) therefore lies on the compute-optimal envelope, and drawing the endpoint finer buys essentially nothing.

Second, a finer endpoint matters only in the data-limited regime.
At the smallest budget $\tilde{B}_c=100$, a cruder top is preferable, since more samples per unit compute outweigh their lower individual accuracy.
The compute-optimal top then rises with budget ($K=3 \to K=5 \to K=6$ as $\tilde{B}_c$ grows), and from $\tilde{B}_c=3{,}000$ on $K=6$, $7$ and $9$ lie within $\approx$3\% of each other.
An over-specified endpoint thus costs accuracy only at low budget, and moderately so (e.g., 8.65\% vs.\ 6.90\% at $\tilde{B}_c=300$ for $K=9$), whereas an under-specified one caps the achievable error at its own accuracy ($\approx$4.1\% for $K=3$ vs.\ $\approx$1.8\% at $\tilde{B}_c=10{,}000$).

Third, our choice is conservative for the central claim, and mixing helps at every endpoint wherever data is the binding constraint.
For $\tilde{B}_c = 300$ to $3{,}000$, the error reduction of multi-fidelity training grows monotonically with endpoint fineness (e.g., $-78.2$\% at $K=9$ vs.\ $-70.9$\% at $K=7$ at $\tilde{B}_c = 300$).
Notably, the reduction grows with fineness even where the achievable error is nearly identical, because a high-fidelity sample costs $K$: a finer endpoint makes the single-fidelity baseline more expensive and hence more data-starved at fixed budget (at $\tilde{B}_c=300$ the pure-HF dataset holds $\approx$3$\times$ fewer samples at $K=9$ than at $K=3$, giving 39.7\% vs.\ 10.1\% single-fidelity error), which the mixture recovers.
Conversely, even at the cheapest endpoint ($K=3$) multi-fidelity training still reduces the error wherever data is the binding constraint ($-62.4$\% at $\tilde{B}_c=100$ and $-22.8$\% at $\tilde{B}_c=300$).
Multi-fidelity mixing thus pays off regardless of where the high-fidelity line is drawn, and a finer line only raises the stakes of going single-fidelity.

In summary, we find that our heuristic endpoint choice lies within the compute-optimal regime and is robust to over-specification. A further refinement, which we leave to future work, is to move beyond the monotone softmax mixture (which for any $\lambda$ concentrates toward an endpoint of the menu) to a family whose mode can lie in the interior (e.g., a truncated Gaussian over log-fidelity with a directly tunable mode).
Such a family could emphasize intermediate fidelities directly, decoupling the compute-optimal allocation from the placement of the endpoints.

\subsection{On the test reference fidelity}
\label{app:test-fidelity-ablation}

In \cref{app:test_error}, we argued that we use $\mathcal{P}_h(\mathbf{f}_h, \xi_{\max})$ and not $\lim_{\xi \to \infty}\mathcal{P}_h(\mathbf{f}_h, \xi)$ as the reference data for testing the surrogates.
To put it differently, we measure the networks against the \emph{best in-support label}.
This was chosen pragmatically to maintain consistency, since some of our mesh-based experiments do not allow us to converge the discretization error all the way to machine precision.
Nevertheless, it may raise the question of how this could affect multi-fidelity gains.

\Cref{tab:test-fidelity-ablation} showcases the result of the 2D U-Net Poisson experiment on the solver error axis, where we additionally evaluate all trained models against the fully converged reference.
Across all budgets $\tilde{B}$, the errors against the converged reference are higher, but the relative difference remains at most $\approx$0.5\%.
As expected, this small difference between the two settings aligns with our desiderata of choosing the highest point of the fidelity menu $\Xi$ to have $e(\xi_{\text{max}}) \approx 0.5\%$, which was outlined in \cref{app:fidelity_support} and lies below the network's error floor (\cref{app:model-size-ablation}).
Since this residual error equally applies to the HF-only models, the relative change $\Delta(\%)$ shifts by at most 0.2 percentage points.

\begin{table}[H]
\centering
\caption{Results for evaluating the surrogates not only against their best in-support label at $\xi_{\max}$, but also against the converged reference $\xi \to \infty$ for the 2D U-Net Poisson experiment on the solver error axis ($K_{\max}=7$ runs of \cref{tab:endpoint-ablation}). Errors are mean relative $L_2$ errors at $\lambda^*$ over 6 seeds; rel.\ diff.\ is computed from unrounded values. $\Delta (\%)$ denotes the relative change in error at $\lambda^*$ compared to the high-fidelity baseline within the menu.}
\label{tab:test-fidelity-ablation}
\begin{tabular}{rccccc}
\toprule
 & \multicolumn{3}{c}{\textbf{Error at $\bm{\lambda^*}$ (\%)}} & \multicolumn{2}{c}{$\bm{\Delta}$\textbf{(\%)}} \\
\cmidrule(lr){2-4} \cmidrule(lr){5-6}
$\bm{\tilde{B}}$ & \textbf{menu top} & \textbf{conv.\ ref} & \textbf{rel.\ diff.} & \textbf{menu top} & \textbf{conv.\ ref} \\
\midrule
100 & 20.56 & 20.64 & +0.42\% & -62.7 & -62.5 \\
300 & 7.73 & 7.75 & +0.25\% & -71.0 & -70.9 \\
1,000 & 3.18 & 3.19 & +0.25\% & -57.6 & -57.5 \\
3,000 & 2.12 & 2.12 & +0.08\% & -32.4 & -32.5 \\
10,000 & 1.77 & 1.77 & +0.51\% & +0.0 & +0.0 \\
\bottomrule
\end{tabular}
\end{table}

\end{document}